\documentclass[10pt,twocolumn,letterpaper]{article}

\usepackage[review]{cvpr}      %
\usepackage{graphicx}
\usepackage{amsmath}
\usepackage{amssymb}
\usepackage{booktabs}
\usepackage{multirow}
\usepackage{pifont}
\usepackage{adjustbox}
\usepackage{enumitem}
\usepackage{array}
\usepackage{tabulary}
\usepackage{mathrsfs}
\usepackage{amsfonts}

\usepackage{colortbl}
\usepackage[accsupp]{axessibility}

\toggletrue{cvprfinal}
\usepackage[dvipsnames]{xcolor}

\definecolor{cvprblue}{rgb}{0.21,0.49,0.74}
\usepackage[pagebackref,breaklinks,colorlinks,citecolor=cvprblue]{hyperref}

\def\paperID{} %
\def\confName{CVPR}
\def\confYear{2024}

\begin{document}

\title{ALGM: \underline{A}daptive \underline{L}ocal-then-\underline{G}lobal Token \underline{M}erging for Efficient Semantic Segmentation with Plain Vision Transformers}

\author{Narges Norouzi, Svetlana Orlova,  Daan de Geus, Gijs Dubbelman\\
Eindhoven University of Technology\\
{\tt\small \{n.norouzi, s.orlova, d.c.d.geus, g.dubbelman\}@tue.nl}
}

\maketitle
\def\thefootnote{\arabic{footnote}}

\begin{abstract}
This work presents Adaptive Local-then-Global Merging (ALGM), a token reduction method for semantic segmentation networks that use plain Vision Transformers.
ALGM merges tokens in two stages: (1) In the first network layer, it merges similar tokens within a small local window and (2) halfway through the network, it merges similar tokens across the entire image. This is motivated by an analysis in which we found that, in those situations, tokens with a  high cosine similarity can likely be merged without a drop in segmentation quality. With extensive experiments across multiple datasets and network configurations, we show that ALGM not only significantly improves the throughput by up to 100\%, but can also enhance the mean IoU by up to +1.1, thereby achieving a better trade-off between segmentation quality and efficiency than existing methods. Moreover, our approach is adaptive during inference, meaning that the same model can be used for optimal efficiency or accuracy, depending on the application. Code is available at \url{https://tue-mps.github.io/ALGM}.

\end{abstract}

\section{Introduction}
\label{sec:intro}

Vision Transformers (ViTs) have shown to be very effective for image segmentation tasks~\cite{kirillov2023segment,zhang2022segvit,chen2022vitadapter,xie2021segformer,strudel2021segmenter,cheng2021mask2former,xu2022groupvit,hengcan2023scalegate,zheng2021setr}. However, the computational complexity of the multi-head self-attention operation scales quadratically with the number of input pixels. This harms the computational efficiency, especially on the high-resolution images that are typically used for image segmentation. To alleviate this burden and improve the efficiency, several works have proposed methods to reduce the number of tokens that the ViT has to process. Most token reduction methods have been introduced for image classification~\cite{heo2023local_global,bolya2023tome,zong2022self-slimmed,ryoo2021tokenlearner,bonnaerens2023learned_threshold,fayyaz2022ats,yin2022avit,wang2021not_all_worth,liang2022evit,meng2022adavit,kong2022spvit,rao2021dynamicvit,wang2023zerotprune,huang2022super_token_sampling,chen2023diffrate,Haurum_2023_which_token_to_use,yu2023xpruner,wei2023joint,long2022beyond,li2022sait,chen2023cfvit,zeng2022TCFormer,wu2023ppt}, but there is also an increasing amount of work that focuses on tasks like semantic segmentation~\cite{tang2023dtop,liu2023dovit,courdier2022paumer,lu2023cts,liang2022expediting,liu2023revisiting,li2023ailurus}. In this work, we also focus on semantic segmentation, and aim to design a token reduction method that achieves a better balance between efficiency and segmentation quality than existing works. 

\begin{figure}[!t]
    \centering
    \includegraphics[width=\linewidth]{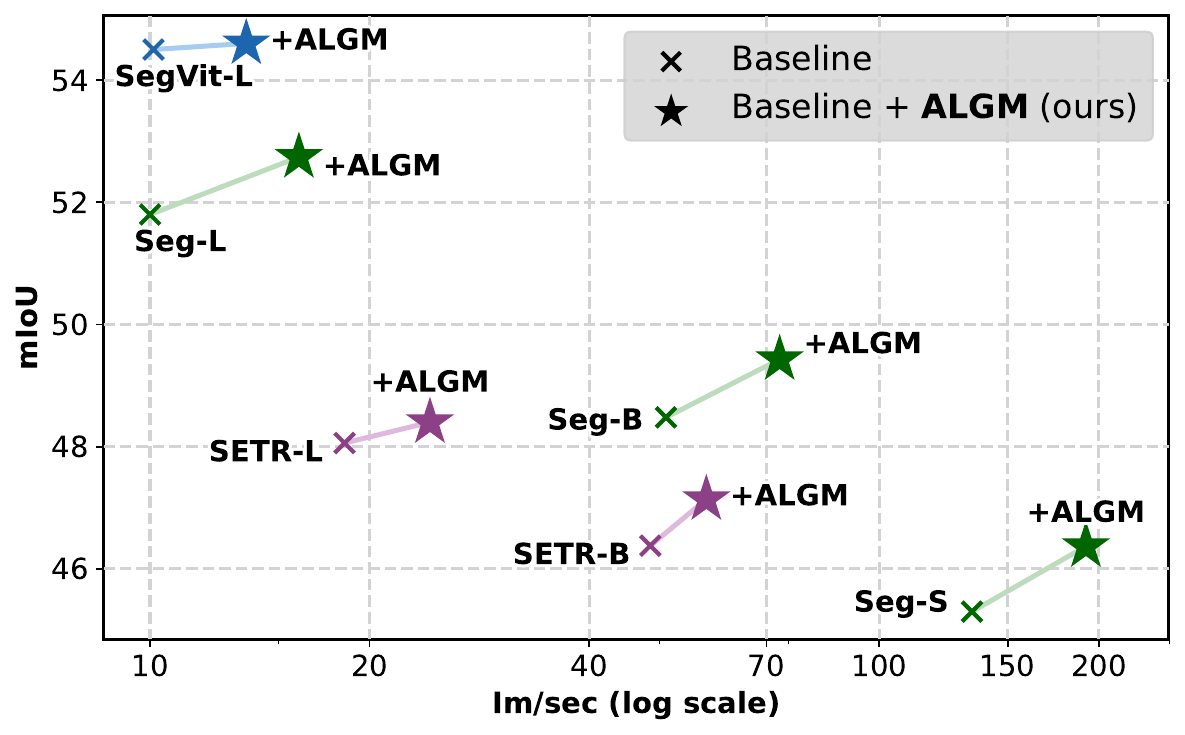}
    \vspace{-15pt}
     \caption{\textbf{Efficiency and segmentation quality for ALGM,} applied to 
 Segmenter~\cite{strudel2021segmenter}, SegViT~\cite{zhang2022segvit}, and SETR~\cite{zheng2021setr} on ADE20K. On average, ALGM improves the throughput of these baselines by 39\%, while improving the mIoU by +0.7.}
    \label{fig:eye_catcher}
    \vspace{-15
pt}
\end{figure}

This objective is motivated by the limitations of existing works. First, \textit{token pruning} methods~\cite{fayyaz2022ats,rao2021dynamicvit,meng2022adavit,yin2022avit}, which are popular for image classification, discard uninformative tokens. They are not directly applicable to semantic segmentation, as each token requires a prediction. To overcome this, \textit{token pausing} or \textit{halting} methods~\cite{courdier2022paumer,tang2023dtop,liu2023dovit} retain discarded tokens and aggregate them with the other tokens at the end of the ViT. However, these methods observe a drop in segmentation quality, possibly because useful information contained in the halted tokens is not available in later network layers. Alternatively, \textit{token sharing} and \textit{merging} methods avoid discarding tokens, and represent multiple image patches or tokens with a smaller set of tokens~\cite{lu2023cts,liang2022expediting,li2023ailurus}. This approach allows these methods to maintain the segmentation quality, but requires them to introduce additional computational overhead to identify tokens for sharing or merging, and they apply token reduction only once, limiting the efficiency gain. Furthermore, token merging methods that have been designed for image classification yield a notable decline in segmentation quality when applied to semantic segmentation~\cite{liang2022expediting,bolya2023tome,liang2022evit}.

Based on these existing works, we make two observations:
\textbf{(a)} CTS~\cite{lu2023cts} demonstrates that local token sharing in the early network stages enhances efficiency without compromising segmentation quality, but it inefficiently requires a pre-processing network. Therefore, our first objective is to \textit{merge redundant tokens early in the network without requiring a pre-processing, and still maintain the segmentation quality}.
\textbf{(b)} Token merging approaches like ToMe~\cite{bolya2023tome} show that gradually merging redundant tokens across the entire image (\ie, globally) can greatly boost the efficiency, but at the cost of segmentation quality. Thus, our second objective is to also \textit{apply global token merging to further improve efficiency, but without harming the segmentation quality.}

To achieve these objectives, we need to find an efficient method to identify tokens that can be merged without causing a drop in segmentation quality. Inspired by existing token merging methods~\cite{bolya2023tome,long2022beyond,wei2023joint}, in \cref{sec:method:similarity_analysis}, we explore if the cosine similarity is a suitable measure to identify mergeable tokens.
Concretely, we compare the cosine similarities between tokens representing the same category -- \ie, \textit{intra-class tokens}, which are potentially redundant and can be merged -- and tokens representing different categories  -- \ie, \textit{inter-class tokens}, which should not be merged. 
We find that \textbf{(a)} already in the 1\textsuperscript{st} network layer, the similarities between intra-class tokens in small local windows are much higher than for inter-class tokens, and \textbf{(b)} comparing tokens globally, intra-class token similarities become increasingly higher than inter-class similarities in later layers.

Based on these findings, we present our Adaptive Local-then-Global Merging (ALGM) module that integrates two token merging phases. In the first network layer, ALGM adopts a local merging strategy. This is followed by a global merging mechanism in an intermediate layer, to also reduce global token redundancies. 
Moreover, rather than using a predetermined number of merged tokens, our approach dynamically decides the number of merged tokens based on the semantic complexity of the image content. Finally, we restore the original token resolution to make segmentation predictions. For details, see \cref{sec:method:algm}.

ALGM offers multiple advantages. \textbf{(a)} Unlike methods that use token pausing, redundant tokens remain active in the network, and continue to contribute in subsequent network layers via their merged representation.
\textbf{(b)} ALGM avoids the need for preprocessing layers and the significant overhead associated with token sharing or merging methods.
\textbf{(c)} Global merging is only applied when token similarity is sufficiently reflective of category similarity, reducing the chance of merging tokens that should remain separate.
\textbf{(d)} Being a parameter-free approach, the ALGM module is naturally compatible with  all plain ViT backbones, as well as any segmentation decoder, with or without re-training.

Through experiments outlined in \cref{sec:exp}, we demonstrate that ALGM consistently enhances the throughput by considerable margins when applied to a wide range of different segmentation methods (see \cref{fig:eye_catcher}). Moreover, we observe that, on top of this improved efficiency, ALGM also enhances the segmentation quality. From an investigation into the cause of this improvement, we find that it can be attributed to two factors: a better balance between frequent and infrequent categories in the self-attention operation, and the denoising of tokens. For more detailed results, we refer to \cref{sec:results}.

 We summarize our main contributions as follows:
\begin{itemize}[noitemsep, topsep=0pt, parsep=0pt, partopsep=0pt]
    \item A generally applicable token merging framework that integrates local and global merging, enhancing both the efficiency and segmentation quality of ViT-based semantic segmentation networks.
    \item An analysis of similarities between intra- and inter-class tokens, within local windows and across network layers.
    \item An exploration of the cause of the segmentation quality improvement obtained by ALGM.

\end{itemize}

\section{Related work}
\label{sec:related_work}

Since the introduction of the Vision Transformer (ViT), a substantial amount of work has been dedicated to improving the efficiency of these ViTs. In this work, we focus on token reduction, which aims to decrease the number of tokens that are processed by the ViT, to improve efficiency.

\paragraph{Token reduction in general.}
The majority of token reduction methods have been introduced for ViTs that conduct image classification. Some methods use a token pruning strategy, where uninformative tokens are identified and simply discarded \cite{rao2021dynamicvit, kong2022spvit, meng2022adavit, fayyaz2022ats, liang2022evit, wang2023zerotprune, yin2022avit}. Uninformative tokens are identified by making intermediate predictions with auxiliary heads \cite{rao2021dynamicvit, kong2022spvit, meng2022adavit}, or obtaining importance scores from attention weights \cite{fayyaz2022ats, liang2022evit, wang2023zerotprune}. Pruned tokens can be discarded completely \cite{yin2022avit, fayyaz2022ats, meng2022adavit, rao2021dynamicvit} or fused into one token to preserve information flow \cite{liang2022evit, kong2022spvit}. While token pruning can notably enhance the throughput of transformers, discarding tokens is not possible for semantic segmentation as each token requires a prediction, and fused tokens representing multiple regions or categories cannot be trivially reconstructed to make a semantic segmentation prediction. Alternatively, token merging methods combine groups of tokens into a smaller set of representative tokens. Some works introduce a learned layer to map the original token set to a smaller one \cite{zong2022self-slimmed, renggli2022tokenmereger, ryoo2021tokenlearner, chang2023making}, while most methods merge tokens if they have a high similarity score~\cite{bolya2023tome, heo2023local_global, bonnaerens2023learned_threshold, chen2023diffrate, cao2023pumer, long2022beyond, wei2023joint}. 
Other methods combine different merging, pruning or fusing approaches~\cite{bonnaerens2023learned_threshold, chen2023diffrate, cao2023pumer, long2022beyond, wu2023ppt}. 

\paragraph{Token reduction for semantic segmentation.}
Some token merging methods for image classification can also be applied to semantic segmentation \cite{chang2023making, bolya2023tome,liang2022evit, heo2023local_global, long2022beyond} by reconstructing merged tokens to their original positions to make a prediction. However, while these approaches improve efficiency, they consistently cause a drop in segmentation quality, as also shown by Liang \etal~\cite{liang2022expediting}. 

Other token reduction methods have been proposed specifically for the semantic segmentation task. Token pausing or halting approaches \cite{courdier2022paumer, tang2023dtop, liu2023dovit} identify tokens that produce high-confidence predictions in early network layers, and do not process them any further. Instead of discarding tokens like pruning methods, they retain tokens and reconstruct them later to make a final prediction. However, in subsequent layers, these `paused' tokens do not participate in the self-attention operation anymore, meaning that potentially useful information is no longer available, which negatively affects the segmentation quality. 
Alternatively, ELViT \cite{liang2022expediting} and AiluRus~\cite{li2023ailurus} introduce a non-parametric token clustering layer that merges redundant neighboring tokens in one network layer. These methods are able to reduce tokens while maintaining the segmentation quality, but their efficiency gain is limited. It is likely that this is because their clustering layer introduces computational overhead, and because they reduce tokens only once, not using the token redundancies potentially present in other layers.

CTS \cite{lu2023cts} uses a CNN-based policy network to identify image patches that can share a token before the first transformer layer. This approach can also reduce tokens while maintaining segmentation quality, but its policy network introduces computational overhead, and it only merges tokens in local windows, ignoring global redundancies.

Inspired by the advantages and limitations of existing work, this work proposes a parameter-free token merging method for semantic segmentation that applies both local and global merging based on cosine similarities between tokens. Importantly, with minimal computational overhead, we identify where cosine similarities can be used to select tokens for merging while maintaining or even improving the segmentation quality.

\section{Method}
\label{sec:method}
\subsection{Preliminaries}
\label{sec:method:prelims}

A ViT-based segmentation architecture typically consists of two subnetworks: \textbf{(a)} An encoder \(\mathcal{E} : I \xrightarrow{\mathcal{L}_1, \ldots, \mathcal{L}_L} \mathit{T}_{L}\), which uses \( L \) distinct transformer layers to map the input image \( I \) to \( \mathit{T}_{L} \), the set of tokens representing the image content at the final layer \( \mathcal{L}_L \). The encoder \( \mathcal{E} \) first splits the input image \( I \in \mathbb{R}^{3 \times H \times W} \) into \( N = \frac{H \times W}{p^2} \) non-overlapping patches, determined by a patch size \( p \). Following patch embedding and positional encoding steps, an initial set of token embeddings \( \mathit{T}_0 = \{t_1, \ldots, t_N\} \) is obtained. Each token \( t_i \) belongs to \( \mathbb{R}^{d} \), with \( d \) denoting the feature dimension. These token embeddings are subsequently processed by transformer layers \( \mathcal{L} = \{\mathcal{L}_1, . . . , \mathcal{L}_L\} \), resulting in the output \( \mathit{T_L} \). Each layer \( \mathcal{L}_{l} \) in \(\mathcal{L}\) integrates a multi-head self-attention (MHSA) block followed by a multi-layer perceptron (MLP) block which output \( \mathit{T}'_l = \text{MHSA}(\mathit{T}_{l-1}) + \mathit{T}_{l-1} \) and \( \mathit{T}_l = \text{MLP}(\mathit{T}'_l) + \mathit{T}'_l \) respectively. 
\textbf{(b)} A decoder \(\mathcal{D} : \mathit{T}_{L} \xrightarrow{} P\), which utilizes transformer or convolutional layers to process \( \mathit{T}_{L} \) and generate the per-pixel segmentation prediction \( P \), where \( P \in \mathbb{R}^{C \times H \times W} \) and \( C \) represents the number of classes. 
Our primary objective is to reduce the number of tokens in \(\mathit{T} \), the total set of tokens, by identifying those tokens that can be merged without adversely affecting the segmentation quality of \( P \).

\begin{figure}[!t]
    \centering
    
    \begin{subfigure}[b]{0.49\linewidth}
        \includegraphics[width=\linewidth]{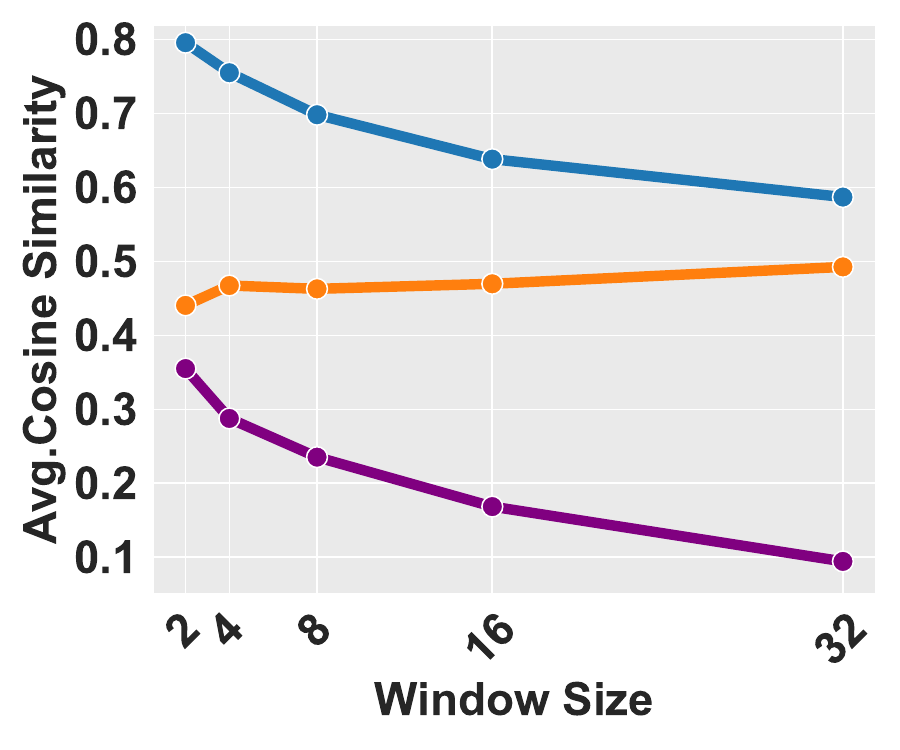}
        \caption{Local token similarities}
        \label{fig:local_sim}
    \end{subfigure}
    \hfill
    \begin{subfigure}[b]{0.49\linewidth}
        \includegraphics[width=\linewidth]{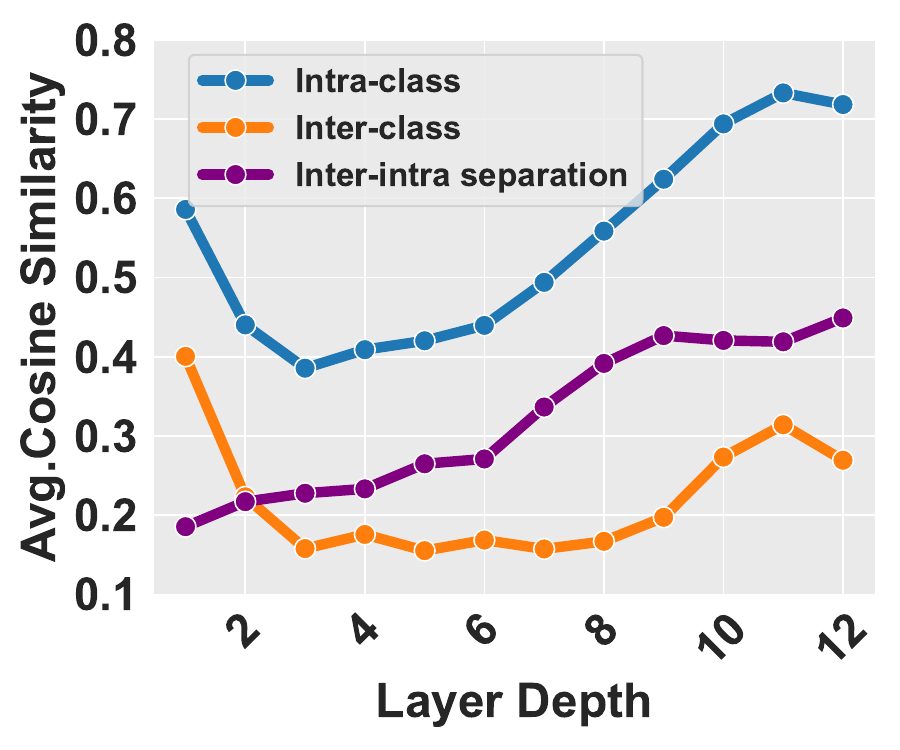}
        \caption{Global token similarities}
        \label{fig:global_sim}
    \end{subfigure}
    
    \caption{\textbf{Comparison of cosine similarity between intra-class and inter-class tokens.} On ADE20K training set using Segmenter + ViT-S~\cite{strudel2021segmenter,dosovitskiy2021an}. (a) Local similarities across 5 window sizes in the first layer. (b) Layer-wise analysis of global similarities.}

    \label{fig:local_global_merging_diff_sim}
    \vspace{-10pt}
\end{figure}

\begin{figure*}[!t]
    \centering
    \includegraphics[width=1\linewidth]{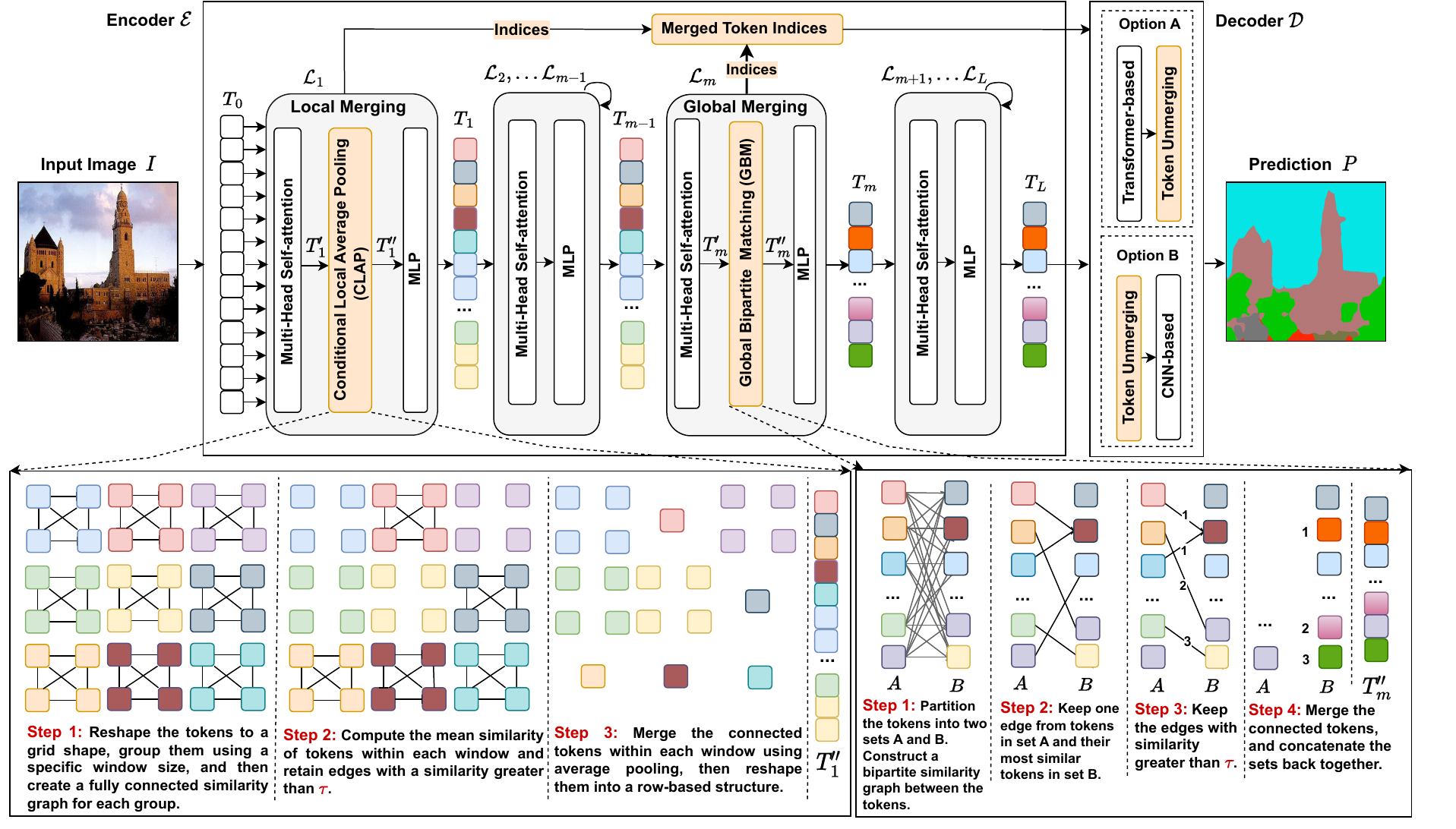}
    \vspace{-19pt}
    \caption{\textbf{ALGM} comprises two primary modules: \textbf{(1) Conditional Local Average Pooling (CLAP)} for local merging and \textbf{(2) Global Bipartite Matching (GBM)} for global merging. The top section illustrates the placement of these modules in the first and middle layers, while the bottom provides a detailed visualization of the individual modules.}
    \label{fig:algm}
    \vspace{-10pt}
\end{figure*}

\subsection{Token similarity analysis}

\label{sec:method:similarity_analysis}
As highlighted in \cref{sec:intro}, considering the advantages and limitations of existing methods, our goal is to find a method to (a) apply early local token merging without requiring a pre-processing network and (b) also apply global token merging to further improve efficiency, without harming the segmentation quality. To achieve this, we need to find a method that can efficiently identify tokens suitable for merging while preserving the segmentation quality. CTS~\cite{lu2023cts} is based on the hypothesis that tokens that represent the same semantic class can be merged without compromising segmentation quality, since they carry redundant information. On the other hand, several token merging methods for image classification~\cite{bolya2023tome,long2022beyond,wei2023joint} merge tokens with a high cosine similarity. They get promising efficiency gains, but sometimes at the cost of accuracy. This motivates us to examine if and when cosine similarity can be an effective metric to identify tokens that represent the same category and are thus suitable for both local and global merging.

To analyze this, we extract and compare the similarities between tokens from Segmenter with ViT-S~\cite{strudel2021segmenter,dosovitskiy2021an} trained on the ADE20K~\cite{zhou2017ade20k} training set. (1) We first analyze the local similarities within ${k}$$\times$${k}$ windows in the first transformer layer. We calculate the cosine similarity between tokens representing different categories (\ie, \textit{inter-class tokens}), which should not be merged, and tokens representing the same category (\ie, \textit{intra-class tokens}), which can be merged. As illustrated in \cref{fig:local_sim}, the smaller the window size $k$, the more accurately cosine similarity reflects that tokens depict the same category. Consequently, tokens with a high cosine similarity within small local windows in the first layer can likely be merged without a drop in segmentation quality. (2) We analyze the global similarities by calculating the cosine similarities for inter-class and intra-class tokens across the entire image for all transformer layers. As seen in \cref{fig:global_sim}, the global similarities in early layers do not accurately represent category correspondence, so they should not be employed to identify tokens for merging. However, deeper in the network, cosine similarity becomes a better measure to identify tokens that can be merged globally without affecting segmentation quality.

\subsection{Adaptive Local-then-Global Merging} 

\label{sec:method:algm}
From the analysis we know that (a) local token similarities in early layers and (b) global token similarities in intermediate layers are likely good indicators of the mergeability of tokens. To exploit this, we propose the Adaptive Local-then-Global Merging (ALGM) approach. As demonstrated in \cref{fig:algm}, the process begins with local merging in the first layer using the Conditional Local Average Pooling (CLAP) module. In an intermediate layer, we adopt the Global Bipartite Merging (GBM) module which is based on the BSM~\cite{bolya2023tome} algorithm for global merging. The procedure concludes with a token unmerging module to restore the original token resolution. 

\paragraph{Local token merging.}
\label{sec:ltm}
Given the insights from \cref{sec:method:similarity_analysis}, we aim to merge tokens in the first layer if they have a high similarity with neighboring tokens in a small window. To implement this, we introduce the CLAP module, which is positioned in layer \(\mathcal{L}_1\), between the MHSA and MLP blocks, as illustrated in \cref{fig:algm}. The CLAP module follows these steps: (1) It receives the token embeddings \(\mathit{T}'_1 \in \mathbb{R}^{N \times d}\) from layer \(\mathcal{L}_1\) and reshapes them into a grid \(\mathit{T}'_{G1} \in \mathbb{R}^{ \frac{H}{p} \times \frac{W}{p} \times d}\). It then defines a window of size \(k \times k\) and groups the tokens within each window into separate sets \( W = \{w_1, \ldots, w_s\} \), where \(s = \frac{N}{k^2}\). Each ${w} \in  \mathbb{R}^{k \times k \times d}$ is a set of token embeddings, represented as \( w = \{ t_1, t_2, \ldots, t_{k^2} \} \). (2) Subsequently, for each $w$ in $W$, it computes the cosine similarity between all pairs of tokens $t_i$ in $w$, and calculates the mean of these similarities to get $\mu_{w}$. Then, as we hypothesize that the similarity between tokens represents the mergeability, CLAP merges only the tokens in windows $w$ for which $\mu_{w} > \tau$, where $\tau$ is an automatically determined similarity threshold, which is elaborated later in this section.
(3) Finally, the tokens inside selected windows $w$ are merged into a single token by taking the average of these tokens. The indices of these tokens are also stored for later unmerging. 
Once completed, merged and non-merged tokens are concatenated to produce the output token embeddings \( T''_1 = \{t_1, \dots, t_{N'}\} \) where \( N' \leq N \).

\paragraph{Global token merging.}
After local merging, the tokens are processed through standard transformer layers up to layer \(\mathcal{L}_m\), where the GBM module is applied. Similar to the steps of the BSM \cite{bolya2023tome} algorithm: (1) In the first step, the tokens in \( T'_m \) are split into two sets: \( A = \{t_1, t_3, \dots, t_{N'-1}\} \) and \( B = \{t_2, t_4, \dots, t_{N'}\} \). A fully-connected bipartite graph is then constructed between the tokens within these sets based on their cosine similarity. (2) Then, GBM only retains the edges that represent the highest similarity from a token in set $A$ to any token in set $B$. This means that for a given token \( A_{i} \) , the edge to token \( B_{j} \) is only retained if \( B_{j} \) is the most similar token to \( A_{i} \) when compared to all other tokens in set \( B \). (3) Next, unlike the original BSM which employs a constant number of merged tokens, GBM uses a similarity threshold \( \tau \). Thus, edges are only retained if their similarity exceeds $\tau$. (4) Finally, the tokens with remaining edges are merged by taking their average, and their indices are stored for future unmerging. The two sets are then concatenated, yielding the output embeddings \( T''_m = \{t_1, \dots, t_{N''}\} \) where \( N'' \leq N' \).

\paragraph{Token unmerging.}

Upon completing global merging, the embeddings \( T''_m \) are processed through the remaining \(L-m\) transformer layers, resulting in the final token embeddings \( \mathit{T}_{L} \). These embeddings, along with the indices of merged tokens retained during the merging phases, are provided as inputs to the decoder \( \mathcal{D} \). In this phase, we deploy an \textit{unmerging} module that duplicates the embeddings of the merged tokens at the indices of the tokens from which they were merged. For transformer-based decoders, which are designed to handle tokens, the unmerging module is applied after the decoder. Conversely, for CNN-based decoders, which require spatially organized features, token unmerging is executed prior to the decoder. 

\paragraph{Adaptive token merging.} \label{par:adaptive-token-merging}

As the complexity of images varies, reducing a constant number of tokens can lead to a suboptimal efficiency or segmentation quality.  Merging too many tokens in complex images can lead to insufficient representation of their complexity. Conversely, simpler images can benefit from a more reduced token count, enhancing efficiency. To tackle this challenge, we introduce an adaptive method that automatically determines a similarity threshold. Before training, we take the base segmentation model to which we want to apply ALGM, and then run inference on the training set. We then extract the tokens after the MHSA block in each layer $\mathcal{L}_l$, calculate the cosine similarities between all token pairs, and compute the mean $\mu_\text{sim}$ and standard deviation $\sigma_\text{sim}$ of these similarities across the entire training set. Given these statistics, we then set the threshold $\tau = \mu_\text{sim} + \sigma_\text{sim}$. 

Using this threshold, the number of remaining tokens $N'$ and $N''$ after the CLAP and GBM modules will vary per image. During training, to facilitate batching of images and tokens, we take the maximum number of remaining tokens $N'$ and $N''$ per batch, and apply this to all images in the batch.

\section{Experimental setup}
\label{sec:exp}

\paragraph{Datasets.}
 We conduct our main experiments on ADE20K~\cite{zhou2017ade20k}, which is widely recognized as a challenging scene parsing dataset. Additionally, we show ALGM's general applicability on the Pascal Context~\cite{mottaghi2014pascalcontext}, Cityscapes~\cite{cordts2016cityscapes}, and COCO-Stuff-10K~\cite{caesar2018cocostuff} datasets.

\paragraph{Implementation details.}
ALGM can be applied to any segmentation model that uses plain ViTs. We apply it to three popular networks: Segmenter~\cite{strudel2021segmenter}, SETR~\cite{zheng2021setr}, and SegViT~\cite{zhang2022segvit} using four standard ViT backbones~\cite{dosovitskiy2021an}: ViT-T/S/B/L. For a fair comparison, we train all networks using the original hyperparameters and official implementations. By default, we integrate our CLAP and GBM modules at the 1\textsuperscript{st} and 5\textsuperscript{th} layers for ViT-T/S/B models, and at the 1\textsuperscript{st} and 7\textsuperscript{th} layers for ViT-L, using the automatically generated threshold \(\tau\) for both merging phases. For more details, see \cref{Sup:Sec:implementation_details}. 

\paragraph{Evaluation metrics.} To assess the segmentation quality, we use the standard mean Intersection-over-Union (mIoU) metric, and for computational efficiency we evaluate the throughput in terms of images per second (im/sec) and the number of floating point operations (FLOPs). For the throughput, we calculate the average im/sec on the validation set with a batch size of 32 on an Nvidia A100 GPU, after a 50-iteration warmup. To calculate the number of FLOPs, we use fvcore~\cite{meta2023fvcore} and compute the average number of operations over all images in the validation set. We report the number of GFLOPs, \ie, FLOPs $\times 10^9$.

\section{Experimental results}
\label{sec:results}
\subsection{Main results}
\label{sec:results:main}
To evaluate the effectiveness of ALGM, we apply it to several ViT-based segmentation networks and compare it with existing state-of-the-art token reduction methods. For each existing method, we report the version with the highest efficiency while maintaining the segmentation quality as much as possible. For a more comprehensive analysis across various settings, additional results are provided in \cref{sup:sec:ALGM_existing_works_various_different_settings}. 
For our ALGM, we report two versions: (1) the default \textbf{ALGM}, which uses the automatic threshold during both training and inference, and (2) \textbf{ALGM*}, which is the same trained model, but during inference it uses the smallest threshold $\tau$ for which the mIoU is higher than the baseline. In other words, ALGM* is tuned for optimal efficiency while maintaining the segmentation quality. See \cref{sec:results:analyses} and \cref{sup:sec:Detailed_analysis_of_ALGM*} for more details on the use of different merging thresholds.

 \begin{table}[t]
\centering
\fontsize{6pt}{7pt}\selectfont
\adjustbox{width=1\linewidth}{
\begin{tabular}{lccc}
\toprule
Method &mIoU (\%)$\uparrow$ & Im/sec $\uparrow$ &GFLOPs $\downarrow$ \\
\midrule

Seg-T~\cite{strudel2021segmenter}\ & 38.1 & 287 & 12.8 \\
+ CTS~\cite{lu2023cts} & 38.2 & 309 & 9.8 \\
+ ToMe~\cite{bolya2023tome} & 37.9 & 346 & 9.2  \\
\rowcolor{gray!30}
+ ALGM (ours) & \textbf{38.9} & 388  & 8.4 \\
\rowcolor{gray!30}
+ ALGM* (ours) &  38.4  & \textbf{427}  & \textbf{7.3} \\
\midrule

Seg-S~\cite{strudel2021segmenter} & 45.3 & 134 & 38.6 \\
+ CTS~\cite{lu2023cts} &  45.1 & 174  & 27.2 \\
+ ToMe~\cite{bolya2023tome} & 45.1 & 170  & 28.2 \\
\rowcolor{gray!30}
+ ALGM (ours) & \textbf{46.4} & 192  & 26.3 \\
\rowcolor{gray!30}
+ ALGM* (ours) & 45.5 &  \textbf{235}  & \textbf{20.9} \\

\midrule

Seg-B~\cite{strudel2021segmenter}\ & 48.5 & 51 & 130 \\
+ CTS~\cite{lu2023cts} & 48.7 & 73 & 91 \\
+ ToMe~\cite{bolya2023tome} & 48.5  &  68 & 97 \\
+ ELViT$^\ddag$~\cite{liang2022expediting} & 48.2 &  73 & 92 \\
\rowcolor{gray!30}
+ ALGM (ours) & \textbf{49.4} & 73 & 91 \\
\rowcolor{gray!30}
+ ALGM* (ours) & 48.5 & \textbf{87}  & \textbf{76} \\
\midrule 

Seg-L~\cite{strudel2021segmenter}\ & 51.8 & 10 & 672 \\
+ CTS~\cite{lu2023cts} & 51.8 & 16 & 446  \\
+ ToMe~\cite{bolya2023tome} & 51.6 & 14  & 505  \\
+ ELViT~\cite{liang2022expediting} & 51.4 & 12  & 539 \\
+ ELViT$^\ddag$~\cite{liang2022expediting} & 51.9 & 12 & 539 \\
+ AiluRus$^\ddag$~\cite{li2023ailurus} & 52.2 & - & 479 \\
\rowcolor{gray!30}
+ ALGM (ours)&\textbf{52.7} & 16 & 438 \\
\rowcolor{gray!30}
+ ALGM* (ours) & 51.9 & \textbf{20}  &  \textbf{370}\\

\midrule
SegViT-L~\cite{zhang2022segvit}\ & 54.5 & 10 & 638 \\
\rowcolor{gray!30}
+ ALGM (ours) & \textbf{54.6} & 14 & 476 \\
\rowcolor{gray!30}
+ ALGM* (ours) & 54.5 & \textbf{15}  & \textbf{461} \\

\midrule
SETR-L~\cite{zheng2021setr} & 48.1 & 18 & 363 \\
\rowcolor{gray!30}
+ ALGM (ours) & \textbf{48.4} & 24 & 277 \\
\rowcolor{gray!30}
+ ALGM* (ours) & 48.1 &  \textbf{29} & \textbf{227} \\
\bottomrule
\end{tabular}
}
\caption{\textbf{Main results on ADE20K.} ALGM applied to Segmenter (Seg)~\cite{strudel2021segmenter}, SegViT~\cite{zhang2022segvit}, and SETR~\cite{zheng2021setr} across 4 ViT backbones. ALGM* is the same trained model as ALGM, but uses the threshold $\tau$ during inference that achieves the best efficiency while maintaining the mIoU w.r.t.~the baseline. $^{\ddag}$Indicates a training-free method, applied directly to the baseline model.}
\label{tab:main_results_ade20k}
\vspace{-15pt}
\end{table}

\begin{table*}
    \begin{subtable}{0.33\linewidth}
    \centering
    \adjustbox{width=\linewidth}{
        \begin{tabular}{lccc}
        \toprule
        Method & mIoU (\%)$\uparrow$ &Im/sec$\uparrow$ &GFLOPs$\downarrow$  \\\midrule
        Seg-S~\cite{strudel2021segmenter}\ & 42.3 & 132  & 39  \\
        + CTS~\cite{lu2023cts} & 42.2 & 164 & 28 \\
        \rowcolor{gray!30}
        + ALGM (ours)  & \textbf{43.1} & 182 & 27  \\
        \rowcolor{gray!30}
        + ALGM* (ours) & 42.3 & \textbf{219}  & \textbf{22} \\
        \midrule
        Seg-B~\cite{strudel2021segmenter}\ & 43.6 & 50  & 130  \\
        + CTS~\cite{lu2023cts} & 43.7 & 71 & 89  \\
        \rowcolor{gray!30}
        + ALGM (ours)  & \textbf{44.4} & 69 & 96 \\
        \rowcolor{gray!30}
        + ALGM* (ours) & 43.8 &  \textbf{84} &  \textbf{79}\\
        \midrule
        
        Seg-L~\cite{strudel2021segmenter} & 46.8 & 18 & 401 \\
        + CTS~\cite{lu2023cts} & 46.7 & 27 & 272 \\
        \rowcolor{gray!30}
        + ALGM (ours) & \textbf{47.4} & 25  & 287 \\
        \rowcolor{gray!30}
        + ALGM* (ours) & 46.8 &\textbf{31}  &\textbf{241} \\
        \bottomrule
        \end{tabular}
    }
    \caption{\textbf{COCO-Stuff~\cite{caesar2018cocostuff}.}}
    \label{tab:main_results_other:cocostuff}
    \end{subtable}
\hfill
    \begin{subtable}{0.33\linewidth}
    \centering
    \adjustbox{width=1\linewidth}{
        \begin{tabular}{lccc}\toprule
        Method &mIoU (\%)$\uparrow$ &Im/sec$\uparrow$ &GFLOPs$\downarrow$  \\
        \midrule
        Seg-S~\cite{strudel2021segmenter}\ &76.5 & 41 & 116 \\
        + CTS~\cite{lu2023cts} &  76.5 & \textbf{81} & \textbf{56}   \\
        \rowcolor{gray!30}
        + ALGM (ours) & \textbf{76.9} & 65  & 76 \\
        \rowcolor{gray!30}
        + ALGM* (ours) & 76.5 & 78 & 64 \\
        
        \midrule 
        Seg-L~\cite{strudel2021segmenter}\ & 79.1 & 6.2 & 1005 \\
        + CTS~\cite{lu2023cts} & \textbf{79.5} & \textbf{13.2} & \textbf{523} \\
        + ToMe~\cite{bolya2023tome} & 78.7 & 8.1 & 822 \\
        + ELViT~\cite{liang2022expediting} &  78.7 & 7.3 & 857 \\
        + ELViT$^\ddag$~\cite{liang2022expediting} & 78.8 & 7.3 & 857 \\
        + AiluRus$^\ddag$~\cite{li2023ailurus} & 78.8 & -& 711 \\
        \rowcolor{gray!30}
        + ALGM (ours) & \textbf{79.5} & 8.6 & 766 \\
        \rowcolor{gray!30}
        + ALGM* (ours) & 79.1 & 10.3  & 654 \\
        \bottomrule
        \end{tabular}
    }
    \caption{\textbf{Cityscapes~\cite{cordts2016cityscapes}.}}
    \label{tab:main_results_other:cityscapes}
    \end{subtable}
\hfill
    \begin{subtable}{0.33\linewidth}
    \centering
    \adjustbox{width=1\linewidth}{
        \begin{tabular}{lccc}\toprule
        Method &mIoU (\%)$\uparrow$ &Im/sec$\uparrow$ &GFLOPs$\downarrow$  \\
        \midrule
        Seg-S~\cite{strudel2021segmenter}\ &53.0 & 172 & 32.1 \\
        + CTS~\cite{lu2023cts} & 52.9 & 220 & 22.4 \\
        \rowcolor{gray!30}
        + ALGM (ours) & \textbf{53.2} & 217 & 24.6 \\
        \rowcolor{gray!30}
        + ALGM* (ours) & 53.0 & \textbf{263}  & \textbf{20.2}\\\midrule 
        
        Seg-L~\cite{strudel2021segmenter}\ &57.7 & 21  & 343 \\
        + CTS~\cite{lu2023cts} & 57.6 & 32 & 233  \\
        + ToMe~\cite{bolya2023tome} &  57.6 & 22 & 263  \\
        + ELViT~\cite{liang2022expediting} & 57.5 & 25 & 257 \\
        + ELViT$^\ddag$~\cite{liang2022expediting} & 57.9 & 25 & 257 \\
        \rowcolor{gray!30}
        + ALGM (ours) & \textbf{58.0} & 30 & 247 \\
        \rowcolor{gray!30}
        + ALGM* (ours) & 57.7 &  \textbf{34}  & \textbf{222} \\
        \bottomrule
        \end{tabular}
    }
    \caption{\textbf{Pascal-Context~\cite{mottaghi2014pascalcontext}.}}
    \label{tab:main_results_other:pascal}
    \end{subtable}
\vspace{-18pt}
\caption{\textbf{Main results on COCO-Stuff, Cityscapes and Pascal-Context.} ALGM applied to Segmenter (Seg) across 3 ViT backbones and 3 datasets. ALGM* is the same trained model as ALGM, but uses the threshold $\tau$ during inference that achieves the best efficiency while maintaining the mIoU w.r.t.~the baseline. $^{\ddag}$Indicates a training-free method, applied directly to the baseline model.}
\label{tab:main_results_other}
\end{table*}

\begin{table}
    \centering
    \adjustbox{width=1.0\linewidth}{
    \begin{tabular}{lcccc}
    \toprule
    \multirow{2}[2]{*}{Method}& \multicolumn{2}{c}{No token reduction} & \multicolumn{2}{c}{With token reduction}\\
    \cmidrule(r){2-3} \cmidrule(l){4-5}
         & mIoU$\uparrow$ & GFLOPs$\downarrow$ & mIoU$\uparrow$ & GFLOPs$\downarrow$ \\
        \midrule

        SETR-B + DToP~\cite{tang2023dtop}  & 47.0 & 108 & 47.0 (\textbf{+0.0}) & 81 (\textbf{-25\%}) \\
        \rowcolor{gray!30}
        SETR-B + ALGM (ours)  & 46.4 & 108 &  47.1 (\textbf{+0.7}) & 86 (\textbf{-20\%}) \\
        \rowcolor{gray!30}
        SETR-B + ALGM* (ours)  & 46.4 & 108 & 46.5 (\textbf{+0.1}) & 75 (\textbf{-30\%}) \\
        
        \midrule 
        Seg-S + DoViT~\cite{liu2023dovit} & 46.2 & 26.6 & 45.8 (\textbf{-0.4}) & 21.8 (\textbf{-18\%}) \\
        \rowcolor{gray!30}
        Seg-S + ALGM (ours)  & 45.3 & 38.6 & 46.4 (\textbf{+1.1}) & 26.3 (\textbf{-32\%}) \\
        \rowcolor{gray!30}
        Seg-S + ALGM* (ours)  & 45.3 & 38.6 & 45.5 (\textbf{+0.2}) & 20.9 (\textbf{-46\%}) \\
        \bottomrule
    \end{tabular}
    }
    \caption{\textbf{ALGM vs.~DToP and DoViT~\cite{tang2023dtop,liu2023dovit}.} Applied to SETR~\cite{zheng2021setr}, and Segmenter (Seg)~\cite{strudel2021segmenter} on ADE20K~\cite{zhou2017ade20k}.}
\label{tab:algm_vs_dtop_dovit}
\vspace{-5pt}
\end{table}

\paragraph{ADE20K.}
\cref{tab:main_results_ade20k} presents the results of ALGM and existing token reduction methods for different segmentation models and ViT backbones on the ADE20K dataset~\cite{zhou2017ade20k}. We find that, across all settings, ALGM is able to considerably improve the throughput and number of GFLOPs with respect to the base segmentation networks, and also achieve a substantial mIoU increase. This shows that our token merging approach does not only improve the efficiency, but also boosts the segmentation quality. In \cref{sec:results:analyses}, we provide a more detailed analysis into the cause of this mIoU improvement. The ALGM* variant, which is optimized for efficiency, is able to improve the throughput even further (up to +100\% for Seg-L), while consistently achieving an mIoU that is the same or slightly higher than the base networks. Moreover, ALGM and ALGM* consistently outperform all existing token reduction works, in terms of both the mIoU and the efficiency metrics. This shows that our method can find a better balance between segmentation quality and efficiency, which is the objective of this work.

\paragraph{Other datasets.}
When applying ALGM to COCO-Stuff, Cityscapes and Pascal-Context~\cite{caesar2018cocostuff,cordts2016cityscapes,mottaghi2014pascalcontext} in \cref{tab:main_results_other}, we observe very similar results as for ADE20K. For all datasets, our default ALGM significantly improves the throughput with respect to the base segmentation networks, while also obtaining a better segmentation quality. Again, ALGM* can improve the throughput even further, obtaining throughput improvements of +90\% on Cityscapes for Seg-S and +72\% on COCO-Stuff for Seg-L without any drop in segmentation quality. 

On the COCO-Stuff and Pascal-Context datasets, ALGM and ALGM* also consistently outperform all existing methods. For Cityscapes, ALGM outperforms ToMe~\cite{bolya2023tome}, AiluRus~\cite{li2023ailurus}, and ELViT~\cite{liang2022expediting}, but it does not achieve the same efficiency improvements as CTS~\cite{lu2023cts}. We observe that the visual homogeneity of the Cityscapes images causes tokens to be similar in the first transformer layer even if they do not belong to the same category, requiring a higher merging threshold and limiting the efficiency improvement. Overall, taking into account all datasets, we can conclude that ALGM is a robust and generally applicable method that consistently improves the efficiency of ViT-based segmentation models while also enhancing their accuracy.

\paragraph{Comparison with other works.}
In \cref{tab:algm_vs_dtop_dovit}, we compare ALGM to DToP~\cite{tang2023dtop} and DoViT~\cite{liu2023dovit}. However, these works report mIoU and GFLOPs results for SETR~\cite{zheng2021setr} and Segmenter~\cite{strudel2021segmenter} without token reduction that differ from the results we obtain from the official code of SETR and Segmenter, while DToP and DoViT do not release their code. Therefore, we can only compare the relative performance differences obtained due to token reduction. In \cref{tab:algm_vs_dtop_dovit}, we observe that DToP can maintain the mIoU while reducing the GFLOPs by 25\%, whereas ALGM* maintains the mIoU and reduces the GFLOPs by 30\%. Compared to DoViT, ALGM considerably improves both the segmentation quality and efficiency. For further comparisons, see \cref{sup:sec:Comparison_with_DToP_Pascal-Context,sup:sec:Comparison_with_PAUMER}.

\begin{table}
\centering
\adjustbox{width=1.0\linewidth}{
\begin{tabular}{lcccc}\toprule
Method &Decoder   & mIoU (\%)$\uparrow$ &Im/sec$\uparrow$& GFLOPs$\downarrow$\\
\midrule
EVA~\cite{fang2023eva}& Mask2Former & 61.5 & 1.9 & 4080 \\
\rowcolor{gray!30}
EVA~\cite{fang2023eva} + ALGM$^\ddag$ & Mask2Former & 61.5 & \textbf{2.4} & 3538 \\
\rowcolor{gray!30}
EVA~\cite{fang2023eva} + ALGM & Mask2Former &\textbf{61.7}& \textbf{2.4} &\textbf{3537}  \\
\midrule

\textcolor{gray}{BEiT-3~\cite{wang2023beit3,chen2022vitadapter}}& \textcolor{gray}{Mask2Former}   &\textcolor{gray}{\textbf{62.0}} & \textcolor{gray}{-}  & \textcolor{gray}{-}\\
\textcolor{gray}{BEiTv2-L~\cite{chen2022vitadapter,peng2022beitv2}}& \textcolor{gray}{Mask2Former} &\textcolor{gray}{61.2} & \textcolor{gray}{-} & \textcolor{gray}{-} \\
\textcolor{gray}{BEiT-L~\cite{bao2022beit,chen2022vitadapter}}& \textcolor{gray}{Mask2Former} &\textcolor{gray}{59.4} & \textcolor{gray}{-} & \textcolor{gray}{-} \\
\textcolor{gray}{SwinV2-G~\cite{liu2021swinv2}}& \textcolor{gray}{UperNet~\cite{xiao2018upernet}}&\textcolor{gray}{59.3} & \textcolor{gray}{-} & \textcolor{gray}{-} \\
\bottomrule
\end{tabular}
}
\caption{\textbf{Application to state-of-the-art.} ALGM is applied to SOTA method EVA + ViT-Adapter + Mask2Former~\cite{fang2023eva,chen2022vitadapter,cheng2021mask2former} and evaluated over the ADE20K validation set, with single-scale testing. $^{\ddag}$Directly applied to the backbone without fine-tuning.}
\label{tab:sota_application}
\vspace{-8pt}
\end{table}

\begin{table*}
    \begin{subtable}[h]{0.33\linewidth}
        \centering
        \scalebox{0.7}{
        \begin{tabular}{cccc}\toprule
        Window size & mIoU (\%)$\uparrow$ &Im/sec$\uparrow$ &GFLOPs$\downarrow$ \\\midrule
        \multicolumn{1}{l}{\textit{Baseline}} & \textit{45.3} & \textit{134} & \textit{38.6}  \\
        \midrule
         \rowcolor{gray!30}
        2$\times$2 & 46.4 & 192 & 26.3    \\
        4$\times$4 & 44.6 & 212 & 22.5 \\
        8$\times$8 & 37.2 & 264 &  13.3  \\
        2$\times$1 & 46.5 & 181 & 28.3    \\
        2$\times$4 & 45.4 & 208 & 23.6    \\
        
        \bottomrule
        \end{tabular}

        }
        \caption{\textbf{CLAP module window size.}}
        \label{tab:ablations:window_size}
    \end{subtable}
\hfill
    \begin{subtable}[h]{0.33\linewidth}
        \centering
        \scalebox{0.7}{
        \begin{tabular}{ccccc}\toprule
            Layer 1 & Layer 5 & mIoU (\%)$\uparrow$ & Im/sec$\uparrow$ & GFLOPs$\downarrow$ \\
        \midrule
        \multicolumn{2}{l}{\textit{Baseline}} & \textit{45.3} & \textit{134} & \textit{38.6}  \\
        \midrule
        CLAP & -  & 45.6 & 161  & 31.9    \\
        GBM & - & 45.0 & 183  & 28.5 \\
        - & GBM  & 45.8 & 144  &  36.1  \\
        GBM & GBM & 44.2 & 216 & 23.7  \\
         \rowcolor{gray!30}
        CLAP & GBM & 46.4 & 192 & 26.3    \\

        \bottomrule
        \end{tabular}
        }
        \caption{\textbf{Merging modules.}}
        \label{tab:ablations:merging_modules}
    \end{subtable}
\hfill
          \begin{subtable}[h]{0.33\linewidth}
        \centering
        \scalebox{0.7}{
        \begin{tabular}{lccc}\toprule
        Layer & mIoU (\%)$\uparrow$ &Im/sec$\uparrow$ &GFLOPs$\downarrow$ \\\midrule
        \textit{Baseline}  & \textit{45.3} & \textit{134} & \textit{38.6}  \\
        \midrule
        {2} & 45.1 & 201 & 25.1    \\
        {4}& 45.8 & 195  & 25.9 \\
        \rowcolor{gray!30}
        {5}& 46.4 & 192  & 26.2 \\
        {6}& 46.3 & 185 & 26.8 \\
        {8}& 46.5 & 176  & 27.6 \\
        
        \bottomrule
        \end{tabular}
        }
        \caption{\textbf{GBM module position.}}
        \label{tab:ablations:merging_positions}
        \end{subtable}

\vspace{-5pt}

\caption{\textbf{Ablations for ALGM.} We apply ALGM to Segmenter~\cite{strudel2021segmenter} with ViT-S~\cite{dosovitskiy2021an} and evaluate on the ADE20K validation set~\cite{zhou2017ade20k}. \textbf{CLAP}: Conditional Local Average Pooling (local merging); \textbf{GBM}: Global Bipartite Matching~\cite{bolya2023tome} (global merging).}
\label{tab:ablations}
\vspace{-5pt}
\end{table*}

\subsection{Application to state-of-the-art model}
\label{sec:results:sota_application}
To demonstrate the effectiveness of ALGM on a large-scale state-of-the-art network, we apply it to the EVA backbone~\cite{fang2023eva}  with a ViT-Adapter + Mask2Former decoder~\cite{chen2022vitadapter,cheng2021mask2former}. Here, we calculate the average throughput on 4 Nvidia A6000 GPUs due to memory requirements. 
The results in \cref{tab:sota_application} show that without training, we can improve the throughput by 26\% while keeping the mIoU constant. When we also train the model, we achieve the same efficiency gains but now also further improve the mIoU with +0.2. These results show the general effectiveness and compatibility of ALGM with large-scale pre-trained networks.

\subsection{Ablations}
\label{sec:results:ablations}

\paragraph{CLAP module window size.}
In \cref{tab:ablations:window_size}, we evaluate the effect of using different window sizes for our CLAP module. We find that smaller window sizes yield higher mIoU scores, whereas larger window sizes result in a better efficiency. This is as expected, as we have found in \cref{fig:local_sim} that, the smaller the local window is, the more likely it is that a high token similarity indicates that tokens depict the same category and can thus share a token without harming the segmentation quality. On the other hand, using smaller window sizes means that fewer tokens are merged, limiting the overall efficiency improvement. 

\paragraph{Impact of merging modules.}
To assess the impact of the individual CLAP and GBM merging modules, we evaluate various configurations in \cref{tab:ablations:merging_modules}. We find that only applying CLAP leads to modest improvements in both the mIoU and the throughput, showing that local merging is effective. If we use the GBM module in the first layer instead, we find that the throughput increases but at the cost of segmentation quality, confirming our findings in \cref{sec:method:similarity_analysis,fig:global_sim} that global token similarities in the first layer should not be used to identify tokens for merging. Conversely, placing the GBM module in layer 5 does yield an improved mIoU, albeit with a lower efficiency gain. Applying GBM in both layer 1 and layer 5 results in a significantly better efficiency, but the incorrectly merged tokens in layer 1 are then merged even further in layer 5, harming the segmentation quality. Finally, combining CLAP with GBM yields the best mIoU while achieving a significant efficiency gain, showing the power of applying both early local merging and later global merging.

\paragraph{GBM module position.}
\cref{tab:ablations:merging_positions} shows the effect of positioning the GBM module at various transformer layers. We find that, for ViT-S, layer 5 yields the best trade-off between mIoU and efficiency, which again shows that early layers should not be used for global merging, and applying it in later layers gives diminishing efficiency returns. For more ablations, see \cref{Sup:Sec:Additional_ablations}.

\begin{table}[t]
\centering
\adjustbox{width=1.\linewidth}{
\begin{tabular}{ccccc}\toprule
 Denoising & Balancing & Token selection & Token reorganization &mIoU (\%)$\uparrow$ \\\midrule
 \rowcolor{gray!30}
\checkmark & \checkmark & Take average & Reduction &  46.4  \\
 \checkmark & \ding{55}  &Take average & No reduction & 45.6 \\
 \ding{55} & \checkmark &Pick random token & Reduction &  45.0  \\
\ding{55} & \ding{55} & Pick random token & No reduction &  44.4   \\

\bottomrule
\end{tabular}
}
\vspace{-5pt}
\caption{\textbf{Analyzing mIoU improvement.} ALGM is applied to Segmenter with ViT-S~\cite{strudel2021segmenter, dosovitskiy2021an} on ADE20K~\cite{zhou2017ade20k}. \textbf{Balancing} refers to attention balancing; \textbf{Denoising} refers to token denoising.}
\label{tab:balancing_denoising}
\vspace{-10pt}
\end{table}

\subsection{Detailed analyses} 
\label{sec:results:analyses}
\paragraph{Cause of segmentation quality improvement.}
The main results in \cref{tab:main_results_ade20k} and \cref{tab:main_results_other} have shown that ALGM not only enhances efficiency, but also improves the segmentation quality. This improvement is most significant for complex datasets, and we hypothesize that it has two causes: (1) \textbf{Balancing}: As tokens that depict the same category are merged, large and frequently-occurring categories are represented by fewer tokens, meaning that they play a less dominant role in the self-attention operation, causing a more balanced attention distribution with respect to rare classes. To assess if this is true, in \cref{tab:balancing_denoising}, we evaluate a setting in which we do not reduce the number of tokens and therefore do not balance the attention process, but instead replicate the average token embedding across the tokens that would otherwise be merged. We find that this causes a drop in mIoU, indicating that attention balancing is indeed a factor in the mIoU improvement of ALGM. (2) \textbf{Denoising}: As tokens are merged, we take the average of their values. This denoises the tokens, which could facilitate the learning process. To evaluate this aspect, we evaluate a configuration where we do not take the average of tokens but instead pick one random token from each set of mergeable tokens. \cref{tab:balancing_denoising} shows that doing so also results in a considerable drop in mIoU. Finally, when disabling both denoising and balancing, the mIoU is at its worst. This implies that both denoising and balancing play an important role, and that the merging of dominant tokens in ALGM is a potent approach to rectify attention imbalances and reduce token noise to enhance the segmentation quality.

\begin{figure}
    \centering
    \includegraphics[width=\linewidth]{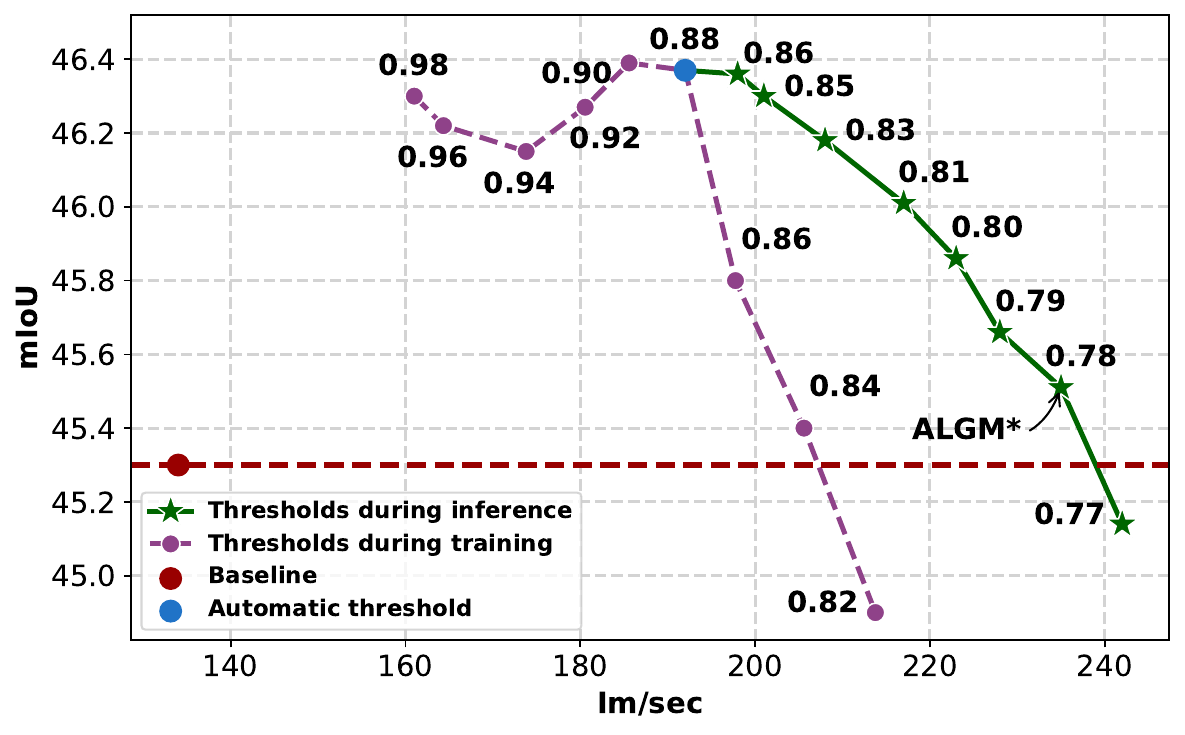}
    \vspace{-20pt}
      \caption{\textbf{Similarity thresholds for token merging.} ALGM applied to Segmenter~\cite{strudel2021segmenter} with ViT-S~\cite{dosovitskiy2021an} on ADE20K~\cite{zhou2017ade20k}. }
    \label{fig:joint_thresholds_ade20k}
    \vspace{-5pt}
\end{figure}

\paragraph{Similarity threshold.}
As explained in \cref{sec:method:algm}, we compute an automatic similarity threshold $\tau$ to select tokens that can be merged. In \cref{fig:joint_thresholds_ade20k}, we show the performance of ALGM with different thresholds. We find that our automatic threshold finds an optimal balance between efficiency and segmentation quality. Interestingly, taking the automatic threshold during training and using lower thresholds during inference leads to less significant mIoU drops than training with a lower $\tau$. We expect that this difference arises from the fact that early-training embeddings are less accurate, leading to overly aggressive merging during training, which the network cannot recover from; naturally, this problem does not occur during inference. Overall, these results show the versatility of ALGM at test time, as it can be used to optimize efficiency while keeping the mIoU the same like we do with ALGM*, but also to achieve a higher mIoU than the baseline. We observe similar results for other datasets, see \cref{Sup:Sec:ALGM_different_training_inference_thresholds}.
\vspace{-5pt}

\begin{figure}[t]
    \centering
    \subcaptionbox{Input image}[0.32\linewidth]{
        \includegraphics[width=\linewidth]{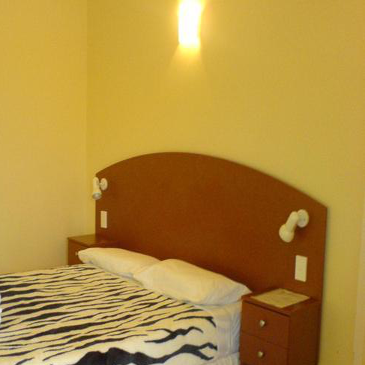}
    }
    \subcaptionbox{Local merging}[0.32\linewidth]{
        \includegraphics[width=\linewidth]{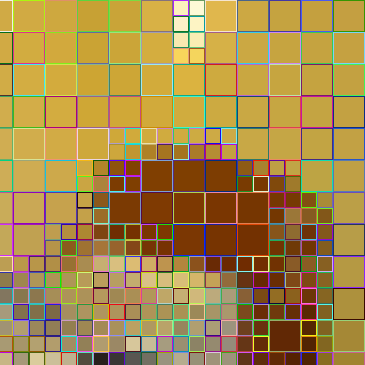}
        
    }
    \subcaptionbox{Global merging}[0.32\linewidth]{
        \includegraphics[width=\linewidth]{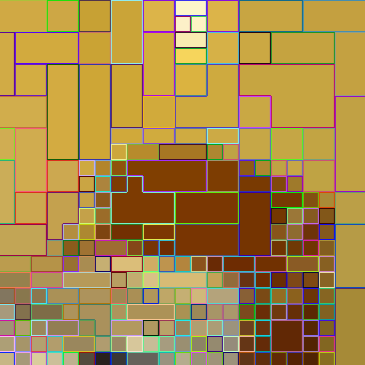}
        }
    \vspace{-7pt}
    \caption{\textbf{Merged tokens.} We depict tokens that are merged as a result of the local CLAP and global GBM merging modules.}
    \label{fig:visualization}
    \vspace{-10pt}
\end{figure}
 
\paragraph{Visualization.}
In~\cref{fig:visualization}, we visualize which tokens are merged as a result of the CLAP local merging and GBM global merging modules. We observe that tokens with visual similarity are mostly merged by CLAP, whereas category correspondence plays a larger role for GBM. For more qualitative results, see \cref{Sup:sec:qualitative_results}.

\section{Discussion}
In this work, we propose a token reduction method for semantic segmentation that combines early local merging with later global merging. This is motivated by the finding that, using this merging strategy, we predominantly merge tokens that contain redundant information, meaning that they can be merged without compromising the segmentation quality. With extensive experiments, we show that our approach is indeed able to find a very good balance between efficiency and segmentation quality, outperforming existing work. Interestingly, we also find that using ALGM for token reduction leads to substantial mIoU improvements for complex datasets. With some first analyses, we find that this is likely caused by improved attention balancing and token denoising. However, further research is required to fully understand the causes of these phenomena and their potential applicability to other networks and tasks. Another interesting avenue for future research could be to examine whether token reduction is similarly effective on more complex tasks like panoptic and video segmentation.

\label{sec:conclusion}

\vspace{-3mm}
{\small \paragraph{Acknowledgements}
This work was funded by the EU project MODI, grant no.~101076810, and the KDT JU EdgeAI project, grant no.~101097300, and utilized the Dutch national e-infrastructure, supported by the SURF Cooperative under grant no.~EINF-5197, funded by the Dutch Research Council (NWO). }


{
    \small
    \bibliographystyle{ieeenat_fullname}
    \bibliography{main}

\begin{thebibliography}{56}
\providecommand{\natexlab}[1]{#1}
\providecommand{\url}[1]{\texttt{#1}}
\expandafter\ifx\csname urlstyle\endcsname\relax
  \providecommand{\doi}[1]{doi: #1}\else
  \providecommand{\doi}{doi: \begingroup \urlstyle{rm}\Url}\fi

\bibitem[Bao et~al.(2022)Bao, Dong, Piao, and Wei]{bao2022beit}
Hangbo Bao, Li Dong, Songhao Piao, and Furu Wei.
\newblock {BEiT: BERT Pre-Training of Image Transformers}.
\newblock In \emph{ICLR}, 2022.

\bibitem[Bolya et~al.(2023)Bolya, Fu, Dai, Zhang, Feichtenhofer, and Hoffman]{bolya2023tome}
Daniel Bolya, Cheng-Yang Fu, Xiaoliang Dai, Peizhao Zhang, Christoph Feichtenhofer, and Judy Hoffman.
\newblock {Token Merging: Your ViT But Faster}.
\newblock In \emph{ICLR}, 2023.

\bibitem[Bonnaerens and Dambre(2023)]{bonnaerens2023learned_threshold}
Maxim Bonnaerens and Joni Dambre.
\newblock {Learned Thresholds Token Merging and Pruning for Vision Transformers}.
\newblock \emph{TMLR}, 2023.

\bibitem[Caesar et~al.(2018)Caesar, Uijlings, and Ferrari]{caesar2018cocostuff}
Holger Caesar, Jasper Uijlings, and Vittorio Ferrari.
\newblock {COCO-Stuff: Thing and Stuff Classes in Context}.
\newblock In \emph{CVPR}, 2018.

\bibitem[Cao et~al.(2023)Cao, Paranjape, and Hajishirzi]{cao2023pumer}
Qingqing Cao, Bhargavi Paranjape, and Hannaneh Hajishirzi.
\newblock {PuMer: Pruning and Merging Tokens for Efficient Vision Language Models}.
\newblock In \emph{ACL}, 2023.

\bibitem[Chang et~al.(2023)Chang, Wang, Lin, Wang, Zhang, Jin, and Shou]{chang2023making}
Shuning Chang, Pichao Wang, Ming Lin, Fan Wang, David~Junhao Zhang, Rong Jin, and Mike~Zheng Shou.
\newblock {Making Vision Transformers Efficient from A Token Sparsification View}.
\newblock In \emph{CVPR}, 2023.

\bibitem[Chen et~al.(2023{\natexlab{a}})Chen, Lin, Li, Shen, Wu, Chao, and Ji]{chen2023cfvit}
Mengzhao Chen, Mingbao Lin, Ke Li, Yunhang Shen, Yongjian Wu, Fei Chao, and Rongrong Ji.
\newblock {CF-ViT: A General Coarse-to-Fine Method for Vision Transformer}.
\newblock In \emph{AAAI}, 2023{\natexlab{a}}.

\bibitem[Chen et~al.(2023{\natexlab{b}})Chen, Shao, Xu, Lin, Zhang, Fei~Chao, Qiao, and Luo]{chen2023diffrate}
Mengzhao Chen, Wenqi Shao, Peng Xu, Mingbao Lin, Kaipeng Zhang, Rongrong~Ji Fei~Chao, Yu Qiao, and Ping Luo.
\newblock {DiffRate: Differentiable Compression Rate for Efficient Vision Transformers}.
\newblock In \emph{ICCV}, 2023{\natexlab{b}}.

\bibitem[Chen et~al.(2023{\natexlab{c}})Chen, Duan, Wang, He, Lu, Dai, and Qiao]{chen2022vitadapter}
Zhe Chen, Yuchen Duan, Wenhai Wang, Junjun He, Tong Lu, Jifeng Dai, and Yu Qiao.
\newblock {Vision Transformer Adapter for Dense Predictions}.
\newblock In \emph{ICLR}, 2023{\natexlab{c}}.

\bibitem[Cheng et~al.(2022)Cheng, Misra, Schwing, Kirillov, and Girdhar]{cheng2021mask2former}
Bowen Cheng, Ishan Misra, Alexander~G. Schwing, Alexander Kirillov, and Rohit Girdhar.
\newblock {Masked-attention Mask Transformer for Universal Image Segmentation}.
\newblock In \emph{CVPR}, 2022.

\bibitem[Cordts et~al.(2016)Cordts, Omran, Ramos, Rehfeld, Enzweiler, Benenson, Franke, Roth, and Schiele]{cordts2016cityscapes}
Marius Cordts, Mohamed Omran, Sebastian Ramos, Timo Rehfeld, Markus Enzweiler, Rodrigo Benenson, Uwe Franke, Stefan Roth, and Bernt Schiele.
\newblock {The Cityscapes Dataset for Semantic Urban Scene Understanding}.
\newblock In \emph{CVPR}, 2016.

\bibitem[Courdier et~al.(2022)Courdier, Sivaprasad, and Fleuret]{courdier2022paumer}
Evann Courdier, Prabhu~Teja Sivaprasad, and François Fleuret.
\newblock {PAUMER: Patch Pausing Transformer for Semantic Segmentation}.
\newblock In \emph{BMVC}, 2022.

\bibitem[Dosovitskiy et~al.(2021)Dosovitskiy, Beyer, Kolesnikov, Weissenborn, Zhai, Unterthiner, Dehghani, Minderer, Heigold, Gelly, Uszkoreit, and Houlsby]{dosovitskiy2021an}
Alexey Dosovitskiy, Lucas Beyer, Alexander Kolesnikov, Dirk Weissenborn, Xiaohua Zhai, Thomas Unterthiner, Mostafa Dehghani, Matthias Minderer, Georg Heigold, Sylvain Gelly, Jakob Uszkoreit, and Neil Houlsby.
\newblock {An Image is Worth 16x16 Words: Transformers for Image Recognition at Scale}.
\newblock In \emph{ICLR}, 2021.

\bibitem[Fang et~al.(2023)Fang, Wang, Xie, Sun, Wu, Wang, Huang, Wang, and Cao]{fang2023eva}
Yuxin Fang, Wen Wang, Binhui Xie, Quan Sun, Ledell Wu, Xinggang Wang, Tiejun Huang, Xinlong Wang, and Yue Cao.
\newblock {EVA: Exploring the Limits of Masked Visual Representation Learning at Scale}.
\newblock In \emph{CVPR}, 2023.

\bibitem[Fayyaz et~al.(2022)Fayyaz, Abbasi~Kouhpayegani, Rezaei~Jafari, Sommerlade, Vaezi~Joze, Pirsiavash, and Gall]{fayyaz2022ats}
Mohsen Fayyaz, Soroush Abbasi~Kouhpayegani, Farnoush Rezaei~Jafari, Eric Sommerlade, Hamid~Reza Vaezi~Joze, Hamed Pirsiavash, and Juergen Gall.
\newblock Adaptive token sampling for efficient vision transformers.
\newblock \emph{ECCV}, 2022.

\bibitem[Haurum et~al.(2023)Haurum, Escalera, Taylor, and Moeslund]{Haurum_2023_which_token_to_use}
Joakim~Bruslund Haurum, Sergio Escalera, Graham~W. Taylor, and Thomas~B. Moeslund.
\newblock {Which Tokens to Use? Investigating Token Reduction in Vision Transformers}.
\newblock In \emph{ICCV Workshops}, 2023.

\bibitem[Heo et~al.(2023)Heo, Fayyazi, Nazemi, and Pedram]{heo2023local_global}
Jung~Hwan Heo, Arash Fayyazi, Mahdi Nazemi, and Massoud Pedram.
\newblock {A Fast Training-Free Compression Framework for Vision Transformers}.
\newblock \emph{arXiv preprint arXiv:2303.02331}, 2023.

\bibitem[Huang et~al.(2023)Huang, Zhou, Cao, He, and Tan]{huang2022super_token_sampling}
Huaibo Huang, Xiaoqiang Zhou, Jie Cao, Ran He, and Tieniu Tan.
\newblock {Vision Transformer with Super Token Sampling}.
\newblock In \emph{CVPR}, 2023.

\bibitem[Kirillov et~al.(2023)Kirillov, Mintun, Ravi, Mao, Rolland, Gustafson, Xiao, Whitehead, Berg, Lo, Dollár, and Girshick]{kirillov2023segment}
Alexander Kirillov, Eric Mintun, Nikhila Ravi, Hanzi Mao, Chloe Rolland, Laura Gustafson, Tete Xiao, Spencer Whitehead, Alexander~C. Berg, Wan-Yen Lo, Piotr Dollár, and Ross Girshick.
\newblock {Segment Anything}.
\newblock In \emph{ICCV}, 2023.

\bibitem[Kong et~al.(2022)Kong, Dong, Ma, Meng, Niu, Sun, Shen, Yuan, Ren, Tang, et~al.]{kong2022spvit}
Zhenglun Kong, Peiyan Dong, Xiaolong Ma, Xin Meng, Wei Niu, Mengshu Sun, Xuan Shen, Geng Yuan, Bin Ren, Hao Tang, et~al.
\newblock {SPViT: Enabling Faster Vision Transformers via Latency-Aware Soft Token Pruning}.
\newblock In \emph{ECCV}, 2022.

\bibitem[Li et~al.(2023)Li, Wang, ZhangHANG, Shi, Jiang, Li, Dai, Xiong, and Tian]{li2023ailurus}
Jin Li, Yaoming Wang, XiaopengIAOPENG ZhangHANG, Bowen Shi, Dongsheng Jiang, Chenglin Li, Wenrui Dai, Hongkai Xiong, and Qi Tian.
\newblock {AiluRus: A Scalable ViT Framework for Dense Prediction}.
\newblock In \emph{NeurIPS}, 2023.

\bibitem[Li et~al.(2022)Li, Thorsley, and Hassoun]{li2022sait}
Ling Li, David Thorsley, and Joseph Hassoun.
\newblock {SaiT: Sparse Vision Transformers through Adaptive Token Pruning}.
\newblock \emph{arXiv preprint arXiv:2210.05832}, 2022.

\bibitem[Liang et~al.(2022{\natexlab{a}})Liang, Yuan, Ding, Luo, Lin, Jia, Zhang, Zhang, and Hu]{liang2022expediting}
Weicong Liang, Yuhui Yuan, Henghui Ding, Xiao Luo, Weihong Lin, Ding Jia, Zheng Zhang, Chao Zhang, and Han Hu.
\newblock {Expediting Large-Scale Vision Transformer for Dense Prediction without Fine-tuning}.
\newblock In \emph{NeurIPS}, 2022{\natexlab{a}}.

\bibitem[Liang et~al.(2022{\natexlab{b}})Liang, Ge, Tong, Song, Wang, and Xie]{liang2022evit}
Youwei Liang, Chongjian Ge, Zhan Tong, Yibing Song, Jue Wang, and Pengtao Xie.
\newblock {Not All Patches are What You Need: Expediting Vision Transformers via Token Reorganizations}.
\newblock In \emph{ICLR}, 2022{\natexlab{b}}.

\bibitem[Liu et~al.(2023)Liu, Gehrig, Messikommer, Cannici, and Scaramuzza]{liu2023revisiting}
Yifei Liu, Mathias Gehrig, Nico Messikommer, Marco Cannici, and Davide Scaramuzza.
\newblock {Revisiting Token Pruning for Object Detection and Instance Segmentation}.
\newblock \emph{arXiv preprint arXiv:2306.07050}, 2023.

\bibitem[Liu et~al.(2024)Liu, Zhou, Wang, Wang, Wang, and Zhang]{liu2023dovit}
Yuang Liu, Qiang Zhou, Jing Wang, Fan Wang, Jun Wang, and Wei Zhang.
\newblock {Dynamic Token-Pass Transformers for Semantic Segmentation}.
\newblock In \emph{WACV}, 2024.

\bibitem[Liu et~al.(2022)Liu, Hu, Lin, Yao, Xie, Wei, Ning, Cao, Zhang, Dong, Wei, and Guo]{liu2021swinv2}
Ze Liu, Han Hu, Yutong Lin, Zhuliang Yao, Zhenda Xie, Yixuan Wei, Jia Ning, Yue Cao, Zheng Zhang, Li Dong, Furu Wei, and Baining Guo.
\newblock {Swin Transformer V2: Scaling Up Capacity and Resolution}.
\newblock In \emph{CVPR}, 2022.

\bibitem[Long et~al.(2023)Long, Zhao, Pi, Wang, and Wang]{long2022beyond}
Sifan Long, Zhen Zhao, Jimin Pi, Shengsheng Wang, and Jingdong Wang.
\newblock {Beyond Attentive Tokens: Incorporating Token Importance and Diversity for Efficient Vision Transformers}.
\newblock In \emph{CVPR}, 2023.

\bibitem[Lu et~al.(2023)Lu, {de Geus}, and Dubbelman]{lu2023cts}
Chenyang Lu, Daan {de Geus}, and Gijs Dubbelman.
\newblock {Content-aware Token Sharing for Efficient Semantic Segmentation with Vision Transformers}.
\newblock In \emph{CVPR}, 2023.

\bibitem[Meng et~al.(2022)Meng, Li, Chen, Lan, Wu, Jiang, and Lim]{meng2022adavit}
Lingchen Meng, Hengduo Li, Bor-Chun Chen, Shiyi Lan, Zuxuan Wu, Yu-Gang Jiang, and Ser-Nam Lim.
\newblock {AdaViT: Adaptive Vision Transformers for Efficient Image Recognition}.
\newblock In \emph{CVPR}, 2022.

\bibitem[Mottaghi et~al.(2014)Mottaghi, Chen, Liu, Cho, Lee, Fidler, Urtasun, and Yuille]{mottaghi2014pascalcontext}
Roozbeh Mottaghi, Xianjie Chen, Xiaobai Liu, Nam-Gyu Cho, Seong-Whan Lee, Sanja Fidler, Raquel Urtasun, and Alan Yuille.
\newblock {The Role of Context for Object Detection and Semantic Segmentation in the Wild}.
\newblock In \emph{CVPR}, 2014.

\bibitem[Peng et~al.(2022)Peng, Dong, Bao, Ye, and Wei]{peng2022beitv2}
Zhiliang Peng, Li Dong, Hangbo Bao, Qixiang Ye, and Furu Wei.
\newblock {BEiT v2: Masked Image Modeling with Vector-Quantized Visual Tokenizers}.
\newblock \emph{arXiv preprint arXiv:2208.06366}, 2022.

\bibitem[Rao et~al.(2021)Rao, Zhao, Liu, Lu, Zhou, and Hsieh]{rao2021dynamicvit}
Yongming Rao, Wenliang Zhao, Benlin Liu, Jiwen Lu, Jie Zhou, and Cho-Jui Hsieh.
\newblock {DynamicViT: Efficient Vision Transformers with Dynamic Token Sparsification}.
\newblock In \emph{NeurIPS}, 2021.

\bibitem[Renggli et~al.(2022)Renggli, Pinto, Houlsby, Mustafa, Puigcerver, and Riquelme]{renggli2022tokenmereger}
Cedric Renggli, Andre~Susano Pinto, Neil Houlsby, Basil Mustafa, Joan Puigcerver, and Carlos Riquelme.
\newblock {Learning to Merge Tokens in Vision Transformers}.
\newblock \emph{arXiv preprint arXiv:2202.12015}, 2022.

\bibitem[Research(2023)]{meta2023fvcore}
Meta Research.
\newblock {fvcore}.
\newblock \url{https://github.com/facebookresearch/fvcore}, 2023.

\bibitem[Russakovsky et~al.(2015)Russakovsky, Deng, Su, Krause, Satheesh, Ma, Huang, Karpathy, Khosla, Bernstein, Berg, and Fei-Fei]{russakovsky2015imagenet}
Olga Russakovsky, Jia Deng, Hao Su, Jonathan Krause, Sanjeev Satheesh, Sean Ma, Zhiheng Huang, Andrej Karpathy, Aditya Khosla, Michael Bernstein, Alexander~C. Berg, and Li Fei-Fei.
\newblock {ImageNet} {Large} {Scale} {Visual} {Recognition} {Challenge}.
\newblock \emph{IJCV}, 115\penalty0 (3):\penalty0 211--252, 2015.

\bibitem[Ryoo et~al.(2021)Ryoo, Piergiovanni, Arnab, Dehghani, and Angelova]{ryoo2021tokenlearner}
Michael~S. Ryoo, AJ Piergiovanni, Anurag Arnab, Mostafa Dehghani, and Anelia Angelova.
\newblock {TokenLearner: What Can 8 Learned Tokens Do for Images and Videos?}
\newblock In \emph{NeurIPS}, 2021.

\bibitem[Shi~Hengcan(2023)]{hengcan2023scalegate}
Cai~Jianfei Shi~Hengcan, Hayat~Munawar.
\newblock {Transformer Scale Gate for Semantic Segmentation}.
\newblock In \emph{CVPR}, 2023.

\bibitem[Strudel et~al.(2021)Strudel, Garcia, Laptev, and Schmid]{strudel2021segmenter}
Robin Strudel, Ricardo Garcia, Ivan Laptev, and Cordelia Schmid.
\newblock {Segmenter: Transformer for Semantic Segmentation}.
\newblock In \emph{ICCV}, 2021.

\bibitem[Tang et~al.(2023)Tang, Zhang, Liu, Liu, and Liu]{tang2023dtop}
Quan Tang, Bowen Zhang, Jiajun Liu, Fagiu Liu, and Yifan Liu.
\newblock {Dynamic Token Pruning in Plain Vision Transformers for Semantic Segmentation}.
\newblock In \emph{ICCV}, 2023.

\bibitem[Wang et~al.(2023{\natexlab{a}})Wang, Dedhia, and Jha]{wang2023zerotprune}
Hongjie Wang, Bhishma Dedhia, and Niraj~K. Jha.
\newblock {Zero-TPrune: Zero-Shot Token Pruning through Leveraging of the Attention Graph in Pre-Trained Transformers}.
\newblock \emph{arXiv preprint arXiv:2305.17328}, 2023{\natexlab{a}}.

\bibitem[Wang et~al.(2023{\natexlab{b}})Wang, Bao, Dong, Bjorck, Peng, Liu, Aggarwal, Mohammed, Singhal, Som, and Wei]{wang2023beit3}
Wenhui Wang, Hangbo Bao, Li Dong, Johan Bjorck, Zhiliang Peng, Qiang Liu, Kriti Aggarwal, Owais~Khan Mohammed, Saksham Singhal, Subhojit Som, and Furu Wei.
\newblock {Image as a Foreign Language: BEiT Pretraining for Vision and Vision-Language Tasks}.
\newblock In \emph{CVPR}, 2023{\natexlab{b}}.

\bibitem[Wang et~al.(2021)Wang, Huang, Song, Huang, and Huang]{wang2021not_all_worth}
Yulin Wang, Rui Huang, Shiji Song, Zeyi Huang, and Gao Huang.
\newblock {Not All Images are Worth 16x16 Words: Dynamic Transformers for Efficient Image Recognition}.
\newblock In \emph{NeurIPS}, 2021.

\bibitem[Wei et~al.(2023)Wei, Ye, Zhang, Tang, and Liang]{wei2023joint}
Siyuan Wei, Tianzhu Ye, Shen Zhang, Yao Tang, and Jiajun Liang.
\newblock {Joint Token Pruning and Squeezing Towards More Aggressive Compression of Vision Transformers}.
\newblock In \emph{CVPR}, 2023.

\bibitem[Wu et~al.(2023)Wu, Zeng, Wang, Wang, and Chen]{wu2023ppt}
Xinjian Wu, Fanhu Zeng, Xiudong Wang, Yunhe Wang, and Xinghao Chen.
\newblock {PPT: Token Pruning and Pooling for Efficient Vision Transformers}.
\newblock \emph{arXiv preprint arXiv:2310.01812}, 2023.

\bibitem[Xiao et~al.(2018)Xiao, Liu, Zhou, Jiang, and Sun]{xiao2018upernet}
Tete Xiao, Yingcheng Liu, Bolei Zhou, Yuning Jiang, and Jian Sun.
\newblock {Unified Perceptual Parsing for Scene Understanding}.
\newblock In \emph{ECCV}, 2018.

\bibitem[Xie et~al.(2021)Xie, Wang, Yu, Anandkumar, Alvarez, and Luo]{xie2021segformer}
Enze Xie, Wenhai Wang, Zhiding Yu, Anima Anandkumar, Jose~M Alvarez, and Ping Luo.
\newblock {SegFormer: Simple and Efficient Design for Semantic Segmentation with Transformers}.
\newblock In \emph{NeurIPS}, 2021.

\bibitem[Xu et~al.(2022)Xu, De~Mello, Liu, Byeon, Breuel, Kautz, and Wang]{xu2022groupvit}
Jiarui Xu, Shalini De~Mello, Sifei Liu, Wonmin Byeon, Thomas Breuel, Jan Kautz, and Xiaolong Wang.
\newblock {GroupViT: Semantic Segmentation Emerges from Text Supervision}.
\newblock In \emph{CVPR}, 2022.

\bibitem[Yin et~al.(2022)Yin, Vahdat, Alvarez, Mallya, Kautz, and Molchanov]{yin2022avit}
Hongxu Yin, Arash Vahdat, Jose Alvarez, Arun Mallya, Jan Kautz, and Pavlo Molchanov.
\newblock {A-ViT: Adaptive Tokens for Efficient Vision Transformer}.
\newblock In \emph{CVPR}, 2022.

\bibitem[Yu and Xiang(2023)]{yu2023xpruner}
Lu Yu and Wei Xiang.
\newblock {X-Pruner: eXplainable Pruning for Vision Transformers}.
\newblock \emph{arXiv preprint arXiv:2303.04935}, 2023.

\bibitem[Zeng et~al.(2022)Zeng, Jin, Liu, Qian, Luo, Ouyang, and Wang]{zeng2022TCFormer}
Wang Zeng, Sheng Jin, Wentao Liu, Chen Qian, Ping Luo, Wanli Ouyang, and Xiaogang Wang.
\newblock {Not All Tokens Are Equal: Human-centric Visual Analysis via Token Clustering Transformer}.
\newblock In \emph{CVPR}, 2022.

\bibitem[Zhang et~al.(2022)Zhang, Tian, Tang, Chu, Wei, Shen, et~al.]{zhang2022segvit}
Bowen Zhang, Zhi Tian, Quan Tang, Xiangxiang Chu, Xiaolin Wei, Chunhua Shen, et~al.
\newblock {SegViT: Semantic Segmentation with Plain Vision Transformers}.
\newblock In \emph{NeurIPS}, 2022.

\bibitem[Zheng et~al.(2020)Zheng, Gao, Zhang, Li, Wang, Li, and Dong]{zheng2020end}
Minghang Zheng, Peng Gao, Renrui Zhang, Kunchang Li, Xiaogang Wang, Hongsheng Li, and Hao Dong.
\newblock End-to-end object detection with adaptive clustering transformer.
\newblock \emph{arXiv preprint arXiv:2011.09315}, 2020.

\bibitem[Zheng et~al.(2021)Zheng, Lu, Zhao, Zhu, Luo, Wang, Fu, Feng, Xiang, Torr, and Zhang]{zheng2021setr}
Sixiao Zheng, Jiachen Lu, Hengshuang Zhao, Xiatian Zhu, Zekun Luo, Yabiao Wang, Yanwei Fu, Jianfeng Feng, Tao Xiang, Philip~H.S. Torr, and Li Zhang.
\newblock {Rethinking Semantic Segmentation from a Sequence-to-Sequence Perspective with Transformers}.
\newblock In \emph{CVPR}, 2021.

\bibitem[Zhou et~al.(2017)Zhou, Zhao, Puig, Fidler, Barriuso, and Torralba]{zhou2017ade20k}
Bolei Zhou, Hang Zhao, Xavier Puig, Sanja Fidler, Adela Barriuso, and Antonio Torralba.
\newblock {Scene Parsing through ADE20K Dataset}.
\newblock In \emph{CVPR}, 2017.

\bibitem[Zong et~al.(2022)Zong, Li, Song, Wang, Qiao, Leng, and Liu]{zong2022self-slimmed}
Zhuofan Zong, Kunchang Li, Guanglu Song, Yali Wang, Yu Qiao, Biao Leng, and Yu Liu.
\newblock {Self-slimmed Vision Transformer}.
\newblock In \emph{ECCV}, 2022.

\end{thebibliography}
}

\clearpage
\appendix 

\section*{Appendix}

In this appendix, we provide the following additional material:
\begin{itemize}
    \item More elaborate implementation details (see \cref{Sup:Sec:implementation_details}).
    \item Additional experimental results (see \cref{sup:sec:additional_results}).
    \begin{itemize}
        \item[•] Comparisons with existing work for a range of token reduction settings, to evaluate the trade-off between efficiency and segmentation quality (see \cref{sup:sec:ALGM_existing_works_various_different_settings}).
        \item[•] Additional comparisons with existing token reduction methods (see \cref{sup:sec:Comparison_with_DToP_Pascal-Context} and \cref{sup:sec:Comparison_with_PAUMER}).
        \item[•] An analysis of the effect of different similarity thresholds for more datasets (see \cref{Sup:Sec:ALGM_different_training_inference_thresholds}).
        \item[•] An analysis of obtaining ALGM* on different segmentation models (see \cref{sup:sec:Detailed_analysis_of_ALGM*}).
        \item[•] Additional ablations (see \cref{Sup:Sec:Additional_ablations}).
    \end{itemize}
    \item Qualitative results (see \cref{Sup:sec:qualitative_results}).
\end{itemize}

\section{Implementation details}
\label{Sup:Sec:implementation_details}

For our main experiments, we implement ALGM on top of the publicly available code of  Segmenter\footnote{\scriptsize \url{https://github.com/rstrudel/segmenter}}~\cite{strudel2021segmenter}, SETR\footnote{\scriptsize \url{https://github.com/fudan-zvg/SETR}}~\cite{zheng2021setr}, SegViT\footnote{\scriptsize \url{https://github.com/zbwxp/SegVit}}~\cite{zhang2022segvit},  and EVA\footnote{\scriptsize \url{https://github.com/baaivision/EVA/tree/master/EVA-01/seg}}~\cite{fang2023eva}.
For Segmenter, SETR, and SegViT, we ensure a fair comparison by training all networks using the original hyperparameters and official implementations. When training models with ViT-T/S/B backbones, we utilize 2 Nvidia A100 GPUs, and for ViT-L, we use 4 Nvidia A100 GPUs.
When training EVA~\cite{fang2023eva}, which comprises 40 transformer layers and 1.0B parameters, a batch size of 32 necessitates 32 GPUs with 40GB VRAM. Given our compute limitations, instead, we finetune this model using 4 Nvidia A6000 GPUs with a batch size of 8 for an additional 10k iterations.

\paragraph{GBM module position.}
\cref{tabels:CLAP_GBM_positions} lists the positions of the CLAP and GBM modules in different segmentation networks. As can be seen, the CLAP module is always applied in layer 1 only. To determine the layer where the GBM module should be applied, we implement GBM in several layers on a pre-trained segmentation model without fine-tuning or training, and find where it yields the best results. Specifically, we select the earliest layer for which the baseline mIoU is maintained, since (a) our objective is to maintain the segmentation quality, and (b) the potential efficiency gain is higher if GBM is applied earlier, since more layers will process a reduced set of tokens. This approach aligns with strategies followed in existing works~\cite{li2023ailurus, liang2022expediting, chang2023making, wei2023joint,long2022beyond}. For EVA~\cite{fang2023eva}, we apply the GBM module in more than one layer, which is further discussed in \cref{Sup:Sec:Additional_ablations}.
The similarity threshold $\tau$ is calculated automatically for all models, given the strategy explained in~\cref{sec:method:algm}.

\paragraph{Adaptive token merging.}
Our method adaptively determines the number of merged tokens for each image using the similarity threshold $\tau$. As a result, each image in the training and validation sets has a different number of remaining tokens $N'$ and $N''$. This variability introduces challenges in batch processing. To solve this during training, ALGM is adaptive on a batch level by using the largest value of $N'$ or $N''$ in the batch. In simpler terms, it merges the same number of tokens for each image in a batch, and this number is the minimum number of mergeable tokens across all images in a batch. This ensures that token reduction is guided by the most complex image in the batch, retaining essential details and eliminating the need for padding.  During inference, with a batch size of 1, which is common in real-world situations, ALGM is adaptive per image.

\paragraph{Throughput evaluation.}
As mentioned in the previous section, the adaptive nature of our method results in a different number of remaining tokens $N'$ and $N''$ for each image in the validation set.
This variability complicates batch processing, which we use for stable throughput evaluation following existing work~\cite{lu2023cts}. Specifically, if we would pad the sets of tokens, or use the largest $N'$ or $N''$ in the batch like during training, this would give an incorrect image of the obtainable throughput. To solve this, and still allow for batched evaluation, we use batches of 32 duplicates of the same image, so that the number of reduced tokens is equal throughout the batch. We apply this to each image, and calculate the average throughput over the entire validation set. 
To ensure a fair comparison, we apply the same approach to evaluate the throughput of existing work, the image crop size used for these calculations is the same that is used during training.

\begin{table}
\centering
\adjustbox{width=1\linewidth}{
\begin{tabular}{lccc}\toprule[1pt]
Network & Backbones & CLAP Layer & GBM Layers \\\midrule
Segmenter~\cite{strudel2021segmenter}  & ViT-T/S/B & 1  &  5 \\
Segmenter~\cite{strudel2021segmenter}  & ViT-L & 1  &  7 \\
SegViT~\cite{zhang2022segvit}  & ViT-L & 1  &  12 \\
SETR~\cite{zheng2021setr} & ViT-B & 1  &  5 \\
SETR~\cite{zheng2021setr} & ViT-L & 1  &  13 \\
EVA~\cite{fang2023eva} & ViT-G & 1  &  11, 21, 31 \\
\bottomrule
\end{tabular}
}
\caption{\textbf{CLAP and GBM module positions} in transformer-based semantic segmentation networks Segmenter~\cite{strudel2021segmenter}, SegViT~\cite{zhang2022segvit}, SETR~\cite{zheng2021setr}, and EVA~\cite{fang2023eva}.}
\label{tabels:CLAP_GBM_positions}
\vspace{-5pt}
\end{table}
\section{Additional results}
\label{sup:sec:additional_results}
\subsection{Comparison with existing work across token reduction settings}
\label{sup:sec:ALGM_existing_works_various_different_settings}

\begin{figure}[t]
    \begin{subfigure}[b]{\linewidth}
        \centering
        \includegraphics[width=0.85\linewidth]{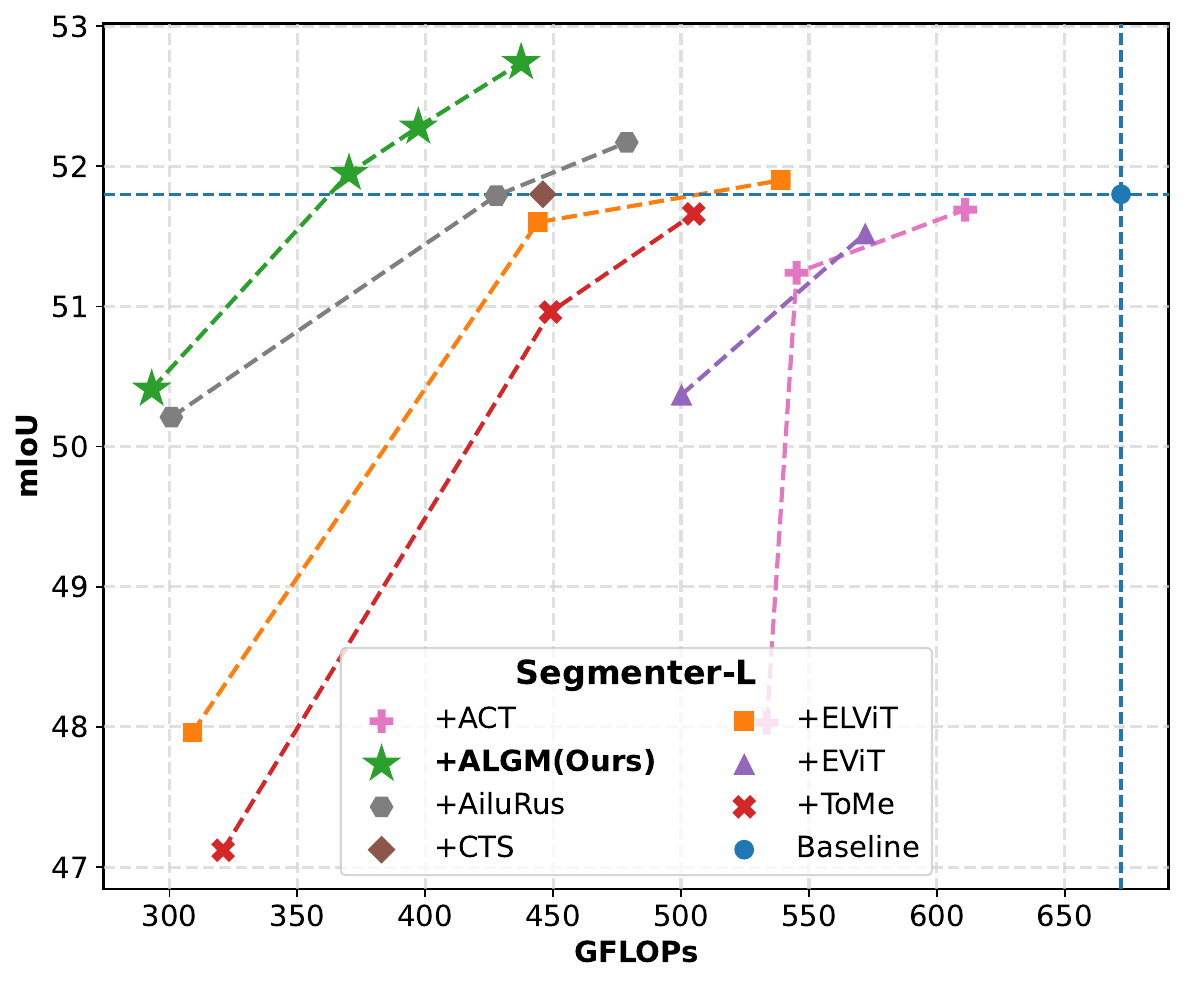}
         \vspace{-2pt}
        \caption{\textbf{ADE20K validation~\cite{zhou2017ade20k}.}}
        \vspace{5pt}
    \end{subfigure}
        \begin{subfigure}[b]{\linewidth}
        \centering
        \includegraphics[width=0.85\linewidth]{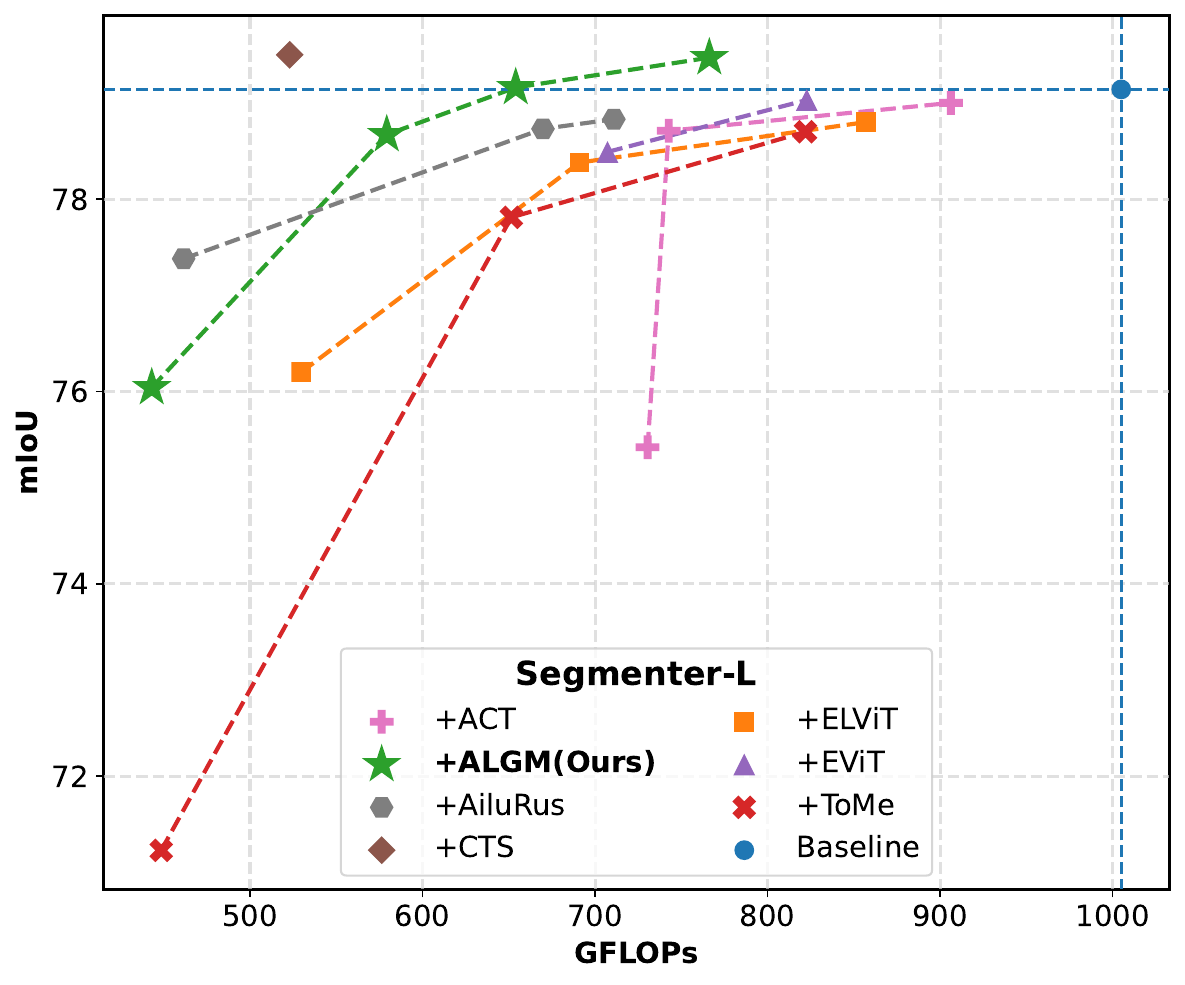}
         \vspace{-2pt}
        \caption{\textbf{Cityscapes val~\cite{cordts2016cityscapes}.}}
        \vspace{5pt}
    \end{subfigure}
        \begin{subfigure}[b]{\linewidth}
        \centering
        \includegraphics[width=0.85\linewidth]{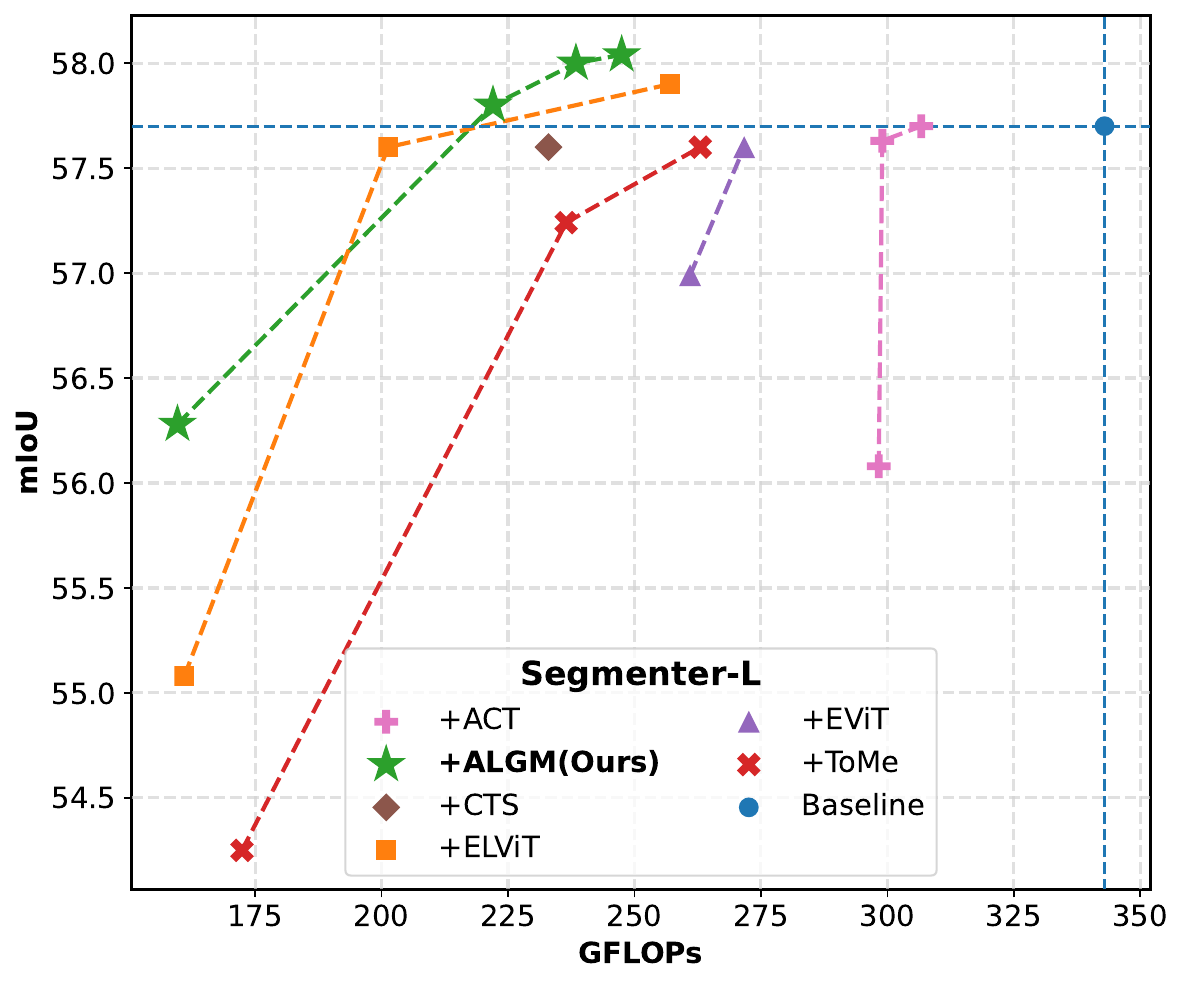}
         \vspace{-2pt}
        \caption{\textbf{Pascal-Context validation~\cite{mottaghi2014pascalcontext}.}}
    \end{subfigure}
    \centering
     \caption{\textbf{Comparison with state-of-the-art methods.} All methods are applied to Segmenter \cite{strudel2021segmenter} with ViT-L~\cite{dosovitskiy2021an}. We compare ALGM to AiluRus~\cite{li2023ailurus}, CTS~\cite{lu2023cts}, ELViT~\cite{liang2022expediting}, ToMe~\cite{bolya2023tome}, EViT~\cite{liang2022evit}, and ACT~\cite{zheng2020end} across different token reduction settings.}
    \label{fig:Comparison_with_existing_methods}
\vspace{-10pt}
\end{figure}

In \cref{sec:results:main}, we present our main results in which, for different existing token reduction methods, we report the version that achieves the highest efficiency while still maintaining the segmentation quality as much as possible. For a more comprehensive comparison, we compare our ALGM with these existing methods across a range of different token reduction settings, essentially evaluating the trade-off between efficiency and segmentation quality. For methods ELViT~\cite{liang2022expediting}, EViT~\cite{liang2022evit}, ToMe~\cite{bolya2023tome}, and ACT~\cite{zheng2020end}, we follow the different token reduction settings specified by Liang \etal~\cite{liang2022expediting}.
For AiluRus~\cite{li2023ailurus}, we report their results across three token reduction settings. For our method, ALGM, we present the results for various similarity threshold values during inference. 
The results for these experiments are provided in \cref{fig:Comparison_with_existing_methods} for ADE20K, Cityscapes and Pascal-Context.

On the ADE20K and Pascal-Context datasets, our ALGM consistently outperforms other methods and achieves a better balance between mIoU and computational efficiency. On the ADE20K dataset, ALGM achieves a mIoU of 51.9, slightly surpassing the baseline, while operating with a 45\% reduction in GFLOPs. When compared to its closest competitor, AiluRus~\cite{li2023ailurus}, our method achieves the same segmentation quality with 14\% fewer GFLOPs. On the other datasets, ALGM also achieves a considerably better balance between the mIoU and GFLOPs than existing works. The only exception is CTS~\cite{lu2023cts} on the Cityscapes dataset. As explained in \cref{sec:results:main}, this is due to the visual homogeneity of the images of this dataset.

\subsection{Comparison with DToP on Pascal-Context}
\label{sup:sec:Comparison_with_DToP_Pascal-Context}
\begin{table}
    \centering
    \adjustbox{width=1.0\linewidth}{
    \begin{tabular}{lcccc}
    \toprule[1pt]
    \multicolumn{1}{c}{}& \multicolumn{2}{c}{No token reduction} & \multicolumn{2}{c}{With token reduction}\\
    \cmidrule(r){2-3} \cmidrule(l){4-5}
         & mIoU$\uparrow$ & GFLOPs$\downarrow$ & mIoU$\uparrow$ & GFLOPs$\downarrow$ \\
        \midrule
        SETR-B + DToP~\cite{tang2023dtop}  & 58.1 & 92 & 58.2 (\textbf{+0.1}) & 69 (\textbf{-25\%}) \\
        \rowcolor{gray!30}
        SETR-B + ALGM (ours)  & 52.2 & 92 & 52.7 (\textbf{+0.5})& 69 (\textbf{-25\%}) \\
        \rowcolor{gray!30}
        SETR-B + ALGM* (ours)  & 52.2 & 92 & 52.3 (\textbf{+0.1})& 59 (\textbf{-36\%}) \\
        \midrule 
        SegViT-L + DToP~\cite{tang2023dtop}  & 63.0 & 315 & 62.7 (\textbf{-0.3}) & 224 (\textbf{-29\%}) \\
        \rowcolor{gray!30}
        SegViT-L + ALGM (ours)  & 64.1 & 334 & 64.2 (\textbf{+0.1}) & 259 (\textbf{-22\%}) \\
        \rowcolor{gray!30}
        SegViT-L + ALGM* (ours)  & 64.1& 334 & 64.1 (\textbf{+0.0})& 237 (\textbf{-29\%}) \\
        \bottomrule
    \end{tabular}
    }
    \caption{\textbf{ALGM vs.~DToP~\cite{tang2023dtop} on Pascal-Context~\cite{mottaghi2014pascalcontext}}, applied to SETR~\cite{zheng2021setr}, and SegViT~\cite{zhang2022segvit}.  ALGM* is the same trained model as ALGM, but uses the threshold $\tau$ during inference that achieves the best efficiency while maintaining the mIoU w.r.t. the baseline.}
\label{supp:tab:algm_vs_dtop_pascal}
\vspace{-10pt}
\end{table}

As mentioned in \cref{sec:results:main}, the performance of SETR~\cite{zheng2021setr} without token reduction as reported by DToP~\cite{tang2023dtop} does not align with the results we obtain from the official code of SETR. Additionally, DToP has not made its code publicly available. As a result, we can only compare to this method in terms of the relative performance differences resulting from token reduction.
In addition to the results presented in \cref{sec:results:main}, \cref{supp:tab:algm_vs_dtop_pascal} compares ALGM to DToP on the Pascal-Context dataset~\cite{mottaghi2014pascalcontext}. For SETR-B, we observe that ALGM achieves a better segmentation quality improvement with the same efficiency. Moreover, ALGM* can achieve a much better efficiency while obtaining the same segmentation quality improvement as DToP. Applied to SegViT-L~\cite{zhang2022segvit}, we find that ALGM* can maintain the mIoU while achieving the same efficiency improvement that DToP obtains while that method causes a mIoU drop. Again, these comparisons highlight that ALGM achieves a better trade-off between segmentation quality and efficiency.

\subsection{Comparison with PAUMER}
\label{sup:sec:Comparison_with_PAUMER}
\begin{table}
\centering
\adjustbox{width=0.9\linewidth}{
\begin{tabular}{lccc}\toprule
Method  & mIoU (\%)$\uparrow$& $\Delta$ mIoU$\uparrow$ & $\Delta$ Im/sec \\\midrule[1pt]
Seg-S~\cite{strudel2021segmenter}  & 45.3 & -  & -  \\\midrule

+ PAUMER~\cite{courdier2022paumer}   & 40.7 & -4.6 & +50\%  \\
+ PAUMER~\cite{courdier2022paumer}  & 34.6 & -10.7 & +100\%  \\
\rowcolor{gray!30}
+ ALGM (ours) & 45.4 & +0.1  & +50\%  \\
\rowcolor{gray!30}
+ ALGM (ours)  & 43.4 & -1.9 & +100\%  \\

\bottomrule
\end{tabular}
}
    \caption{\textbf{ALGM vs.~PAUMER~\cite{courdier2022paumer}.} All models are applied to Segmenter (Seg) \cite{strudel2021segmenter} with ViT-S~\cite{dosovitskiy2021an} and evaluated on the ADE20K~\cite{zhou2017ade20k} validation set; $\Delta$ indicates differences.}

\label{tabels:Vs_PAUMER_Tbl}
\end{table}
When comparing our method to PAUMER~\cite{courdier2022paumer},
we observe similar challenges as for DToP~\cite{tang2023dtop}, 
so we can only compare in terms of relative throughput changes. \cref{tabels:Vs_PAUMER_Tbl} presents a comparative analysis between ALGM and PAUMER when applied to Segmenter~\cite{strudel2021segmenter} on the ADE20K validation set. The results show that, at equal throughput improvements, ALGM achieves significantly better mIoU scores. The difference is especially notable when the throughput is increased by +100\%, where PAUMER causes a mIoU drop of -10.7 but ALGM can limit the decrease to -1.9.

\begin{figure}[!t]
    \begin{subfigure}[b]{\linewidth}
        \centering
        \includegraphics[width=\linewidth]{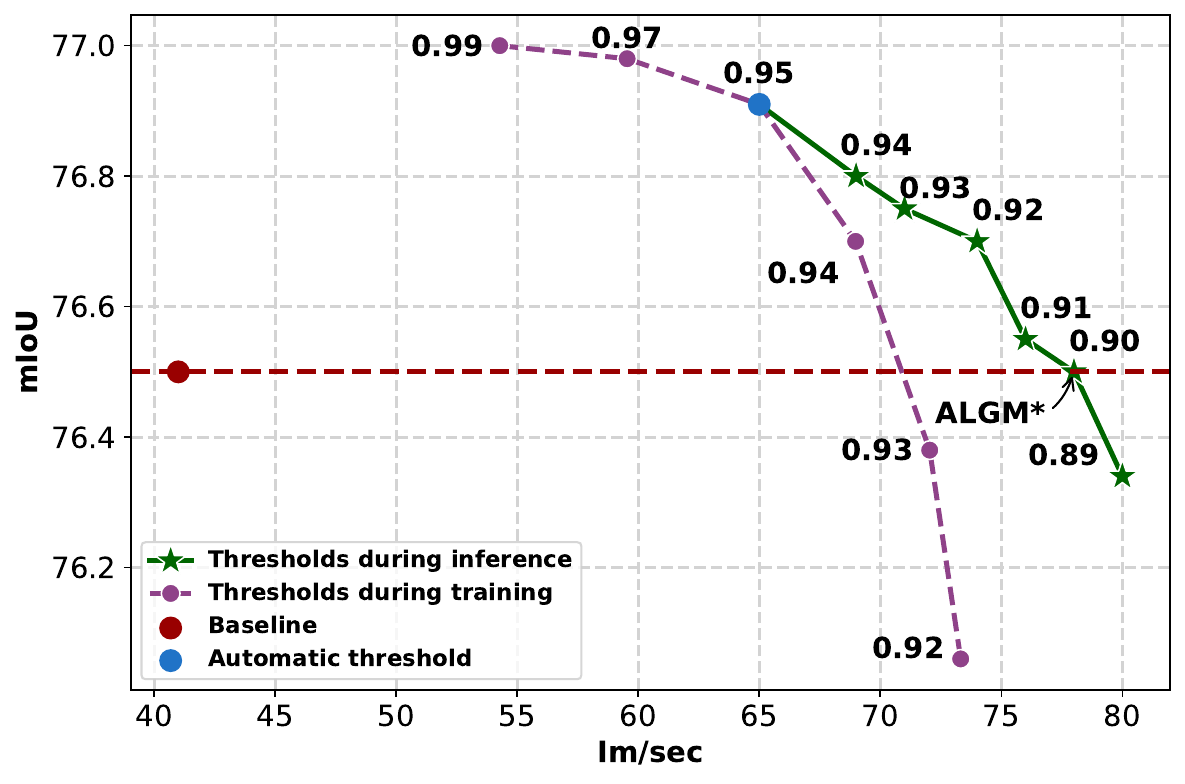}
         \vspace{-15pt}
        \caption{\textbf{Cityscapes val~\cite{cordts2016cityscapes}.}}
        \vspace{8pt}
    \end{subfigure}

    \begin{subfigure}[b]{\linewidth}
        \centering
        \includegraphics[width=\linewidth]{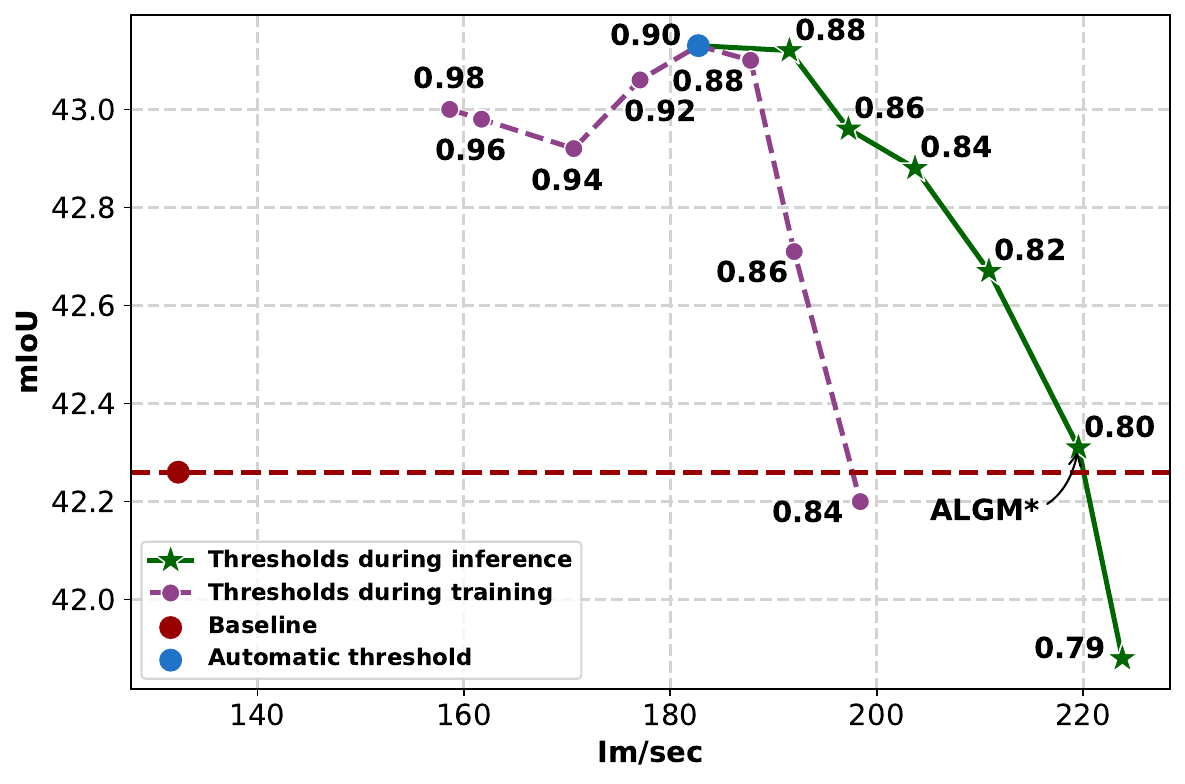}
        \vspace{-15pt}
        \caption{\textbf{COCO-Stuff validation~\cite{caesar2018cocostuff}.}}
        \vspace{8pt}
    \end{subfigure}

    \begin{subfigure}[b]{\linewidth}
        \centering
        \includegraphics[width=\linewidth]{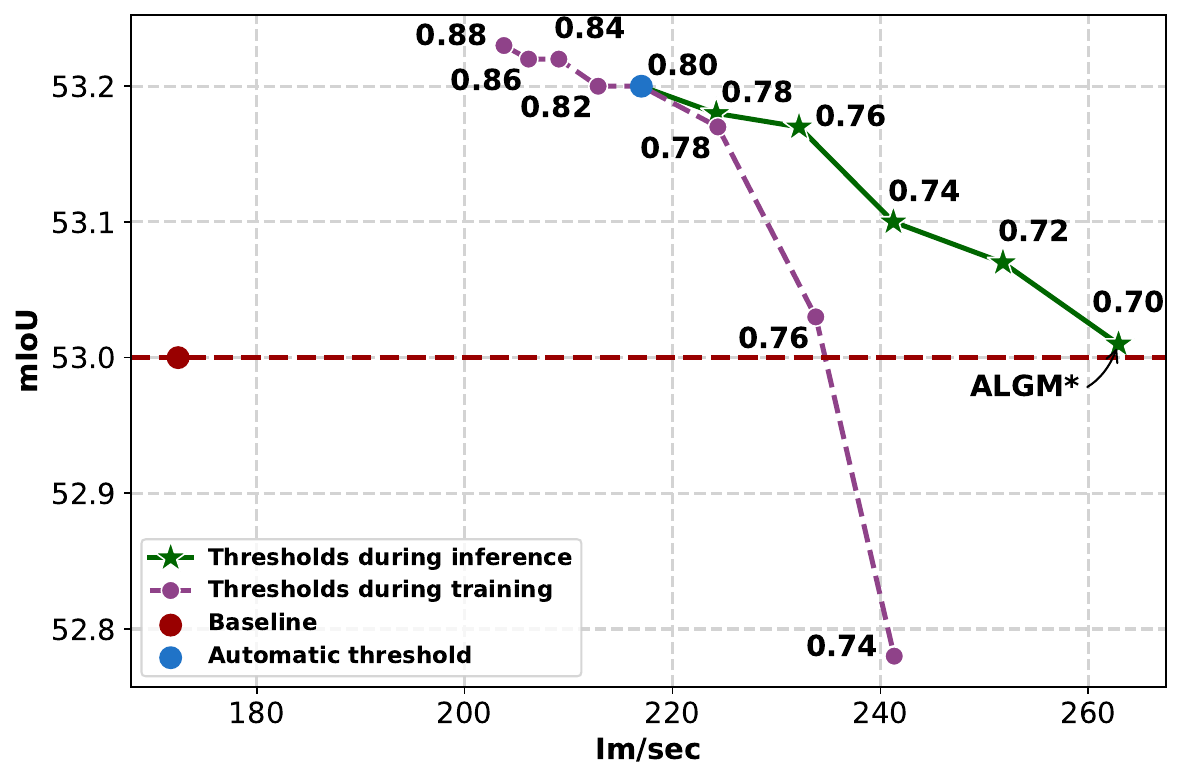}
        \vspace{-15pt}
        \caption{\textbf{Pascal-Context validation~\cite{mottaghi2014pascalcontext}.}}
    \end{subfigure}
 \caption{\textbf{Different similarity thresholds for token merging during training and inference.} ALGM is applied to Segmenter~\cite{strudel2021segmenter} with ViT-S~\cite{dosovitskiy2021an}. ALGM* is the same trained model as ALGM, but uses the threshold $\tau$ during inference that achieves the best efficiency while maintaining the mIoU w.r.t. the baseline.}
\label{fig:different_tresholds}
\end{figure}

\begin{figure*}[!t]
    \centering
    \includegraphics[width=0.32\textwidth]{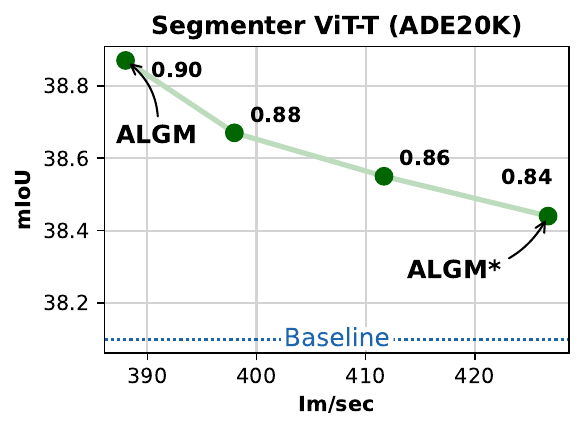}
    \hfill %
    \includegraphics[width=0.32\textwidth]{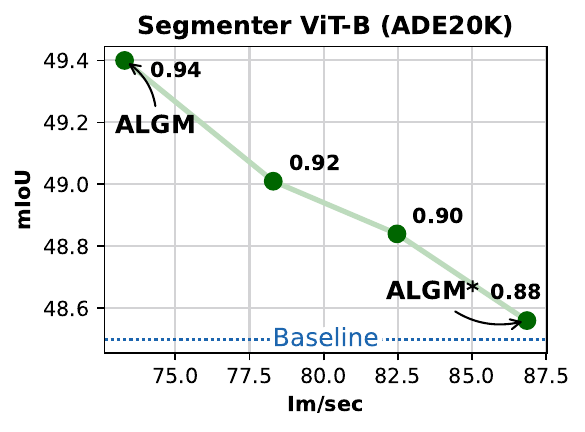}
    \hfill %
    \includegraphics[width=0.32\textwidth]{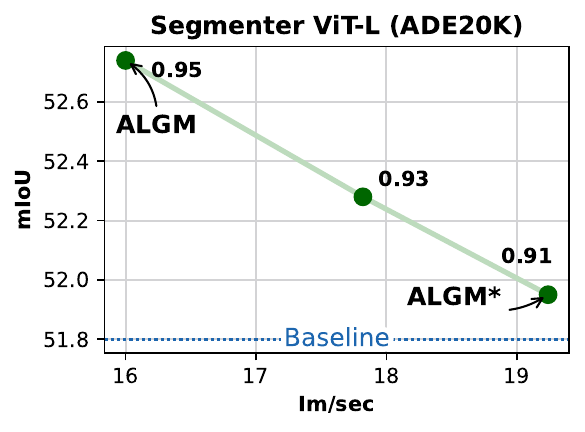}
    \vfill %
    \includegraphics[width=0.32\textwidth]{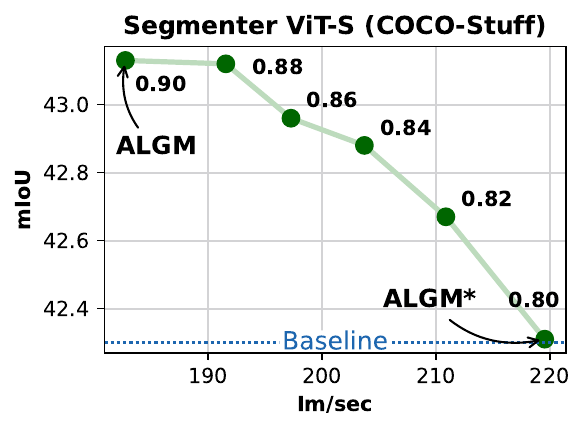}
    \hfill %
    \includegraphics[width=0.32\textwidth]{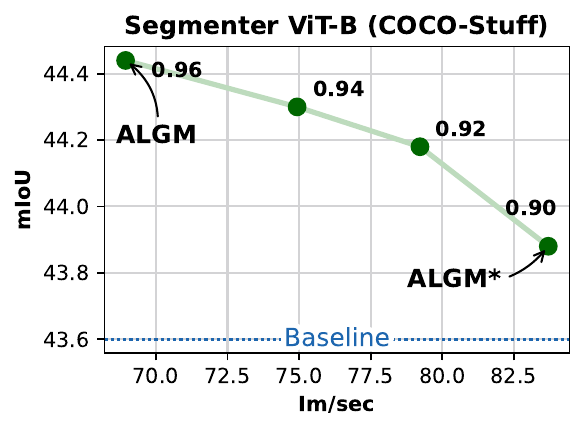}
    \hfill %
    \includegraphics[width=0.32\textwidth]{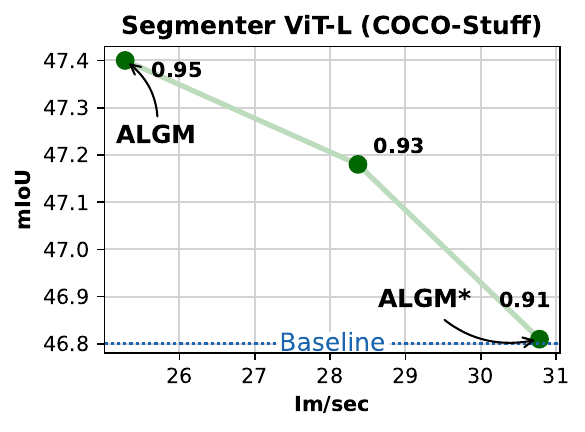}
    \vfill %
    \includegraphics[width=0.32\textwidth]{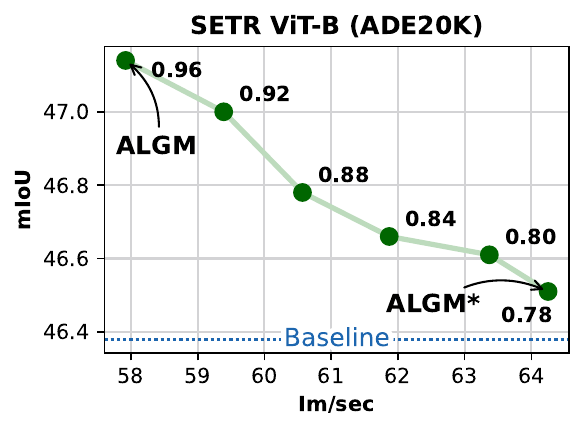}
    \hfill %
    \includegraphics[width=0.32\textwidth]{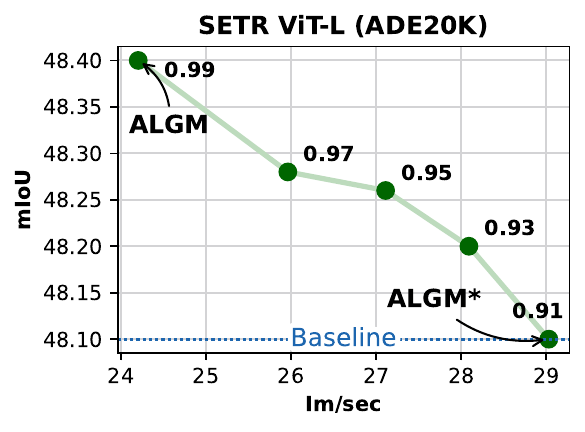}
    \hfill %
    \includegraphics[width=0.32\textwidth]{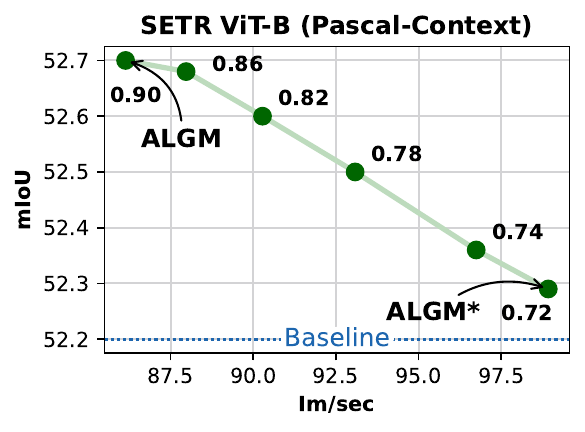}
    \caption{\textbf{Obtaining ALGM*.} ALGM is applied to Segmenter \cite{strudel2021segmenter} and SETR \cite{zheng2021setr} with various backbones (ViT-T, ViT-S, ViT-B and ViT-L)~\cite{dosovitskiy2021an} on the ADE20K~\cite{zhou2017ade20k}, COCO-Stuff~\cite{caesar2018cocostuff} and Pascal-Context~\cite{mottaghi2014pascalcontext} validation sets. These figures show the values of thresholds $\tau$ for the ALGM and ALGM* versions. ALGM* is the same trained model as ALGM, but uses the threshold $\tau$ during inference that achieves the best efficiency while maintaining the mIoU w.r.t. the baseline. }
    \label{fig:Thresholding_all_backbones}
  
\end{figure*}

\subsection{Different similarity thresholds}
\label{Sup:Sec:ALGM_different_training_inference_thresholds}
In \cref{sec:results:analyses}, we explore the impact of different similarity thresholds $\tau$ on the model's performance on ADE20K. Here, we conduct these experiments also for other datasets. \cref{fig:different_tresholds} shows the results for the Cityscapes, COCO-Stuff and Pascal-Context datasets. For all datasets, it is clear that automatic thresholds offer a good balance between efficiency and segmentation quality. Similar to the findings for the ADE20K dataset, we observe that employing a lower threshold during training leads to a significant decrease in mIoU. However, using a lower threshold during inference results in a more modest decline in mIoU while notably improving efficiency. These findings enable a valuable strategy, where we train ALGM with the automatic threshold, but can reduce the threshold $\tau$ during inference to improve the efficiency with minimal impact on the mIoU. This how we obtain ALGM*. This also demonstrates the versatility of our method, as it is suitable for various applications with different demands for efficiency and accuracy.

Notably, the automatically calculated threshold $\tau$ for the Cityscapes dataset is relatively high compared to the thresholds obtained for other datasets. This observation aligns with our results in \cref{sec:results:main}, where we explained that the visual homogeneity of Cityscapes images causes tokens to have high cosine similarities in the first transformer layer, even when they do not depict the same category. This necessitates a higher merging threshold, consequently limiting the potential efficiency improvements.

\subsection{Obtaining ALGM*}
\label{sup:sec:Detailed_analysis_of_ALGM*}
In our main experiments, we present two versions of our method: (1) \textbf{ALGM}, which consistently applies an automatic threshold during both training and inference, and (2) \textbf{ALGM*}, which is the same trained model as ALGM but uses the lowest possible threshold $\tau$ during inference for which the mIoU remains above the baseline. This ALGM* version is designed to optimize efficiency while maintaining the segmentation quality. To illustrate the process of obtaining ALGM*, \cref{fig:Thresholding_all_backbones} shows the results of ALGM with different thresholds during inference, and compares this with the baseline mIoU performance without token reduction.

\subsection{Additional ablations}
\label{Sup:Sec:Additional_ablations}

\begin{table*}[t]
    \begin{subtable}{0.32\linewidth}
        \centering
        \scalebox{0.63}{
        \begin{tabular}{lccc}\toprule[1pt]
        Layer & mIoU (\%)$\uparrow$ &Im/sec$\uparrow$ &GFLOPs$\downarrow$ \\\midrule
        \textit{Baseline} & \textit{45.3} & \textit{134} & \textit{38.6}  \\
        \midrule
        
         5 \& 7 & 45.4 & 214 &  23.1 \\
         \rowcolor{gray!30}
         5 & 46.4 & 192 & 26.3  \\
      
        \bottomrule[1pt]
        \end{tabular}
        }
        \caption{\textbf{Multiple GBM modules.}}
        \label{tab:ablations:multiple_GBM}
    \end{subtable}
    \hfill
    \begin{subtable}{0.32\linewidth}
        \centering
        \scalebox{0.63}{
        \begin{tabular}{lccc}\toprule[1pt]
        Position & mIoU (\%)$\uparrow$ &Im/sec$\uparrow$ &GFLOPs$\downarrow$ \\\midrule
        \textit{Baseline}  & \textit{45.3} & \textit{134} & \textit{38.6}  \\\midrule
        After MLP &  46.0 & 187  &  26.9   \\
        \rowcolor{gray!30}
        Betw. MHSA \& MLP & 46.4 & 192 & 26.3  \\
        
        \bottomrule[1pt]
        \end{tabular}
        }
        \caption{\textbf{Merging module placement.}}
        \label{tab:ablations:merging_module_placement}
    \end{subtable}
    \hfill
    \begin{subtable}{0.32\linewidth}
        \centering
        \scalebox{0.63}{
        \begin{tabular}{lccc}\toprule[1pt]
        Merging method & mIoU (\%)$\uparrow$ &Im/sec$\uparrow$ &GFLOPs$\downarrow$ \\\midrule
        \textit{Baseline}  & \textit{45.3} & \textit{134} & \textit{38.6}  \\\midrule
        Pick random token & 45.0 & 193 & 26.1  \\
        \rowcolor{gray!30}
        Take average & 46.4 & 192 &  26.3   \\
        
        \bottomrule[1pt]
        \end{tabular}
        }
        \caption{\textbf{Token merging operation.}}
        \label{tab:ablations:merging_methods}
    \end{subtable}
    \vspace{5pt}
    \\
    \begin{subtable}{0.32\linewidth}
        \centering
        \scalebox{0.63}{
        \begin{tabular}{lccc}\toprule[1pt]
        Layer & mIoU (\%)$\uparrow$ &Im/sec$\uparrow$ &GFLOPs$\downarrow$ \\\midrule
        \textit{Baseline} & \textit{45.3} & \textit{134} & \textit{38.6}  \\
        \midrule

        + ALGM (Direct)& 45.3 & 193  & 26.2  \\
        + ALGM (Fine-tuning) & 45.7 & 192 &  26.3 \\
         \rowcolor{gray!30}
        + ALGM (Training)& 46.4 & 192 & 26.3  \\
      
        \bottomrule[1pt]
        \end{tabular}
        }
        \caption{\textbf{Effect of training.}}
        \label{tab:ablations:Is_retraining_essential}
    \end{subtable}
    \hfill
    \begin{subtable}{0.32\linewidth}
        \centering
        \scalebox{0.63}{
        \begin{tabular}{lccc}\toprule[1pt]
        Layer & mIoU (\%)$\uparrow$ &Im/sec$\uparrow$ &GFLOPs$\downarrow$ \\\midrule
        \textit{Baseline} & \textit{45.3} & \textit{134} & \textit{38.6}  \\
        \midrule
         \rowcolor{gray!30}
         1 & 46.4 & 192 & 26.3  \\
         2 & 46.4 & 176 & 28.5  \\
         3 & 46.3 & 168 & 29.8  \\

        \bottomrule[1pt]
        \end{tabular}
        }
        \caption{\textbf{CLAP module position.}}
        \label{tab:ablations:CLAP_module_position}
    \end{subtable}
        \hfill
    \begin{subtable}{0.32\linewidth}
            \centering
            \scalebox{0.63}{
            \begin{tabular}{lccc}\toprule[1pt]
            Feature & mIoU (\%)$\uparrow$ &Im/sec$\uparrow$ &GFLOPs$\downarrow$ \\\midrule
            \textit{Baseline} & \textit{45.3} & \textit{134} & \textit{38.6}  \\
            \midrule
             
             K (Key) & 45.9 & 200 & 25.5  \\
             Q (Query) & 45.9 & 200 & 25.5  \\
             V (Value) & 46.0 & 192 & 26.1    \\
             \rowcolor{gray!30}
             X (Token) & 46.4 & 192 & 26.3   \\

            \bottomrule[1pt]
            \end{tabular}
            }
            \caption{\textbf{Feature selection.}}
            \label{tab:ablations:feature_selection}
        \end{subtable}

\vspace{-5pt}
\caption{\textbf{Ablations.} We evaluate different settings for ALGM. We apply ALGM to Segmenter~\cite{strudel2021segmenter} with ViT-S~\cite{dosovitskiy2021an} and evaluate on the ADE20K validation set~\cite{zhou2017ade20k}. MHSA = Multi-head self-attention.}
\label{supp:tab:ablations}
\end{table*}

\paragraph{Multiple GBM modules.} In \cref{tab:ablations:multiple_GBM}, we examine the effect of applying the GBM module in more than one layer. The results indicate that while the application of the GBM module in both the 5\textsuperscript{th} and 7\textsuperscript{th} layer significantly increases throughput, it also results in a noticeable reduction in mIoU compared to its sole application in the 5\textsuperscript{th} layer. This implies that overly aggressive global token merging using the GBM module negatively impacts segmentation quality.

\begin{table}
\centering
\adjustbox{width=0.9\linewidth}{
\begin{tabular}{lcccc}\toprule
Method &Layers   & mIoU (\%)$\uparrow$ &Im/sec$\uparrow$& GFLOPs$\downarrow$\\
\midrule
EVA~\cite{fang2023eva}& - & 61.5 & 1.9 & 4080 \\\midrule
\rowcolor{gray!30}
+ ALGM$^\ddag$ & 11, 21, 31 & 61.5 & 2.4 & 3538 \\
+ ALGM$^\ddag$ & 11, 21 & 61.4 &  2.1 & 3722 \\
+ ALGM$^\ddag$ & 11 & 61.4 & 2 & 3872 \\

\bottomrule
\end{tabular}
}
\caption{\textbf{Multiple GBM modules in EVA.} ALGM is applied to SOTA method EVA + ViT-Adapter + Mask2Former~\cite{fang2023eva,chen2022vitadapter,cheng2021mask2former} and evaluated on the ADE20K validation set, with single-scale testing. $^{\ddag}$Directly applied to the backbone without fine-tuning.}
\label{tab:eva}
\vspace{-8pt}
\end{table}

For EVA~\cite{fang2023eva}, which is a much larger model with 40 transformer layers, we conduct a similar experiment in \cref{tab:eva}. Here, we find that applying the GBM module multiple times does not cause a drop in mIoU. We hypothesize that a very large model like EVA introduces considerable additional redundancies in its many layers, which GBM can then reduce without harming the segmentation quality. However, further research is required to explore this in more detail.

\paragraph{Merging module placement.} 
We conduct an ablation to identify the optimal location for the merging modules within a transformer layer. As shown in \cref{tab:ablations:merging_module_placement}, placing them between the multi-head self-attention (MHSA) block and the MLP yields the best performance in terms of both the segmentation quality and the efficiency. 

\paragraph{Token merging operation.}~\cref{tab:ablations:merging_methods} compares the performance of different token merging operations. \textit{Pick random token} represents the operation where a single random token is picked from each set of tokens that can be merged, and is used to replace these tokens. This approach results in the loss of important information because the selected token might not be the best representation of the collective set of tokens. On the other hand, taking the average of all tokens in each set yields a much better performance. It causes the merged token to be better representative of the original tokens, because it consolidates the information from all tokens in the set. Moreover, it can denoise these token embeddings, as discussed in \cref{sec:results:analyses}.

\paragraph{Effect of training.}
\label{Sup:Sec:is_training_essentia}
Since our method introduces no additional learnable parameters, ALGM can easily be integrated with off-the-shelf pre-trained ViT-based networks to run inference directly while reducing tokens. To assess the impact of training the models after module integration, we explore three scenarios: (a) directly applying the module during inference without any additional training, (b)  fine-tuning the model for an additional 16 epochs, resuming training from the model pre-trained on the ADE20K dataset, and (c) training the model for 64 epochs, starting from the model pre-trained on the ImageNet dataset~\cite{russakovsky2015imagenet}. These are situations that have been evaluated before in earlier work~\cite{tang2023dtop}. The results of these approaches are presented in \cref{tab:ablations:Is_retraining_essential}. We observe that applying ALGM improves efficiency across all scenarios. While direct integration maintains the baseline mIoU, further training, particularly training from scratch, significantly improves the segmentation quality. As we discuss in \cref{sec:results:analyses}, these results indicate that training the model with the ALGM module leverages the benefits of attention balancing and token denoising during training, leading to improvements in segmentation quality. 

\begin{table}
\centering
\scalebox{0.8}{
\begin{tabular}{lccc}\toprule[1pt]
Method   &mIoU (\%)$\uparrow$ &Im/sec$\uparrow$ &GFLOPs$\downarrow$ \\\midrule
Seg-L  & 51.8 & 10 & 672 \\\midrule
\rowcolor{gray!30}
+ ALGM  (Direct)& 51.8 & \textbf{16}  & \textbf{436}  \\
+ ELViT~\cite{liang2022expediting} (Direct)& 51.9 & 12  & 539  \\
+ AiluRus~\cite{li2023ailurus} (Direct)& \textbf{52.2} &  - & 479 \\
+ ToMe~\cite{bolya2023tome} (Direct)& 51.7 & 14  & 505  \\
\midrule
\rowcolor{gray!30}
+ ALGM  (Training) & \textbf{52.7} & \textbf{16} &  \textbf{438 }\\
+ ELViT~\cite{liang2022expediting} (Training)& 51.4 & 12  & 539  \\
+ ToMe~\cite{bolya2023tome} (Training)& 51.6 & 14  & 505  \\
\midrule
+ AiluRus~\cite{li2023ailurus} (Fine-tuning)& 52.1 &  - & -  \\
\bottomrule[1pt]
\end{tabular}
}
\caption{\textbf{Effect of training with ALGM and existing work.} All methods are applied to Segmenter (Seg)~\cite{strudel2021segmenter} with ViT-L~\cite{dosovitskiy2021an} and evaluated on ADE20K~\cite{zhou2017ade20k}. We evaluate three settings. (1) Direct: direct application without further training. (2) Training: training the model from scratch for 160k iterations. (3) Fine-tuning: resuming training from a pre-trained model for 160k iterations.}
\label{tabels:Impact_of_training_Tbl}
\vspace{-5pt}
\end{table}

For further insights into the effect of training, we also compare ALGM against ELViT~\cite{liang2022expediting}, ToMe~\cite{bolya2023tome}, and AiluRus~\cite{li2023ailurus}, which are explicitly designed for direct application without additional training. As shown in \cref{tabels:Impact_of_training_Tbl}, although ALGM is primarily designed for training from scratch, it still achieves competitive mIoU results without training, while being more efficient. Furthermore, we observe that training these existing training-free methods does not cause them to perform better than when applied directly. This shows that training a network with token reduction methods does not automatically give a mIoU boost, and that it is the design of our approach that enables this. 

\paragraph{CLAP module position.}
In this experiment, we investigate the impact of applying the CLAP module in different transformer layers. We keep the GBM module at the 5\textsuperscript{th} layer. The findings, as outlined in \cref{tab:ablations:CLAP_module_position}, show that token embeddings in the first layer are sufficiently representative of class correspondence for effective local merging. Delaying the application of CLAP to the second or third layers does not significantly impact the mIoU, but it does negatively affect the efficiency. Thus, applying the CLAP module in the first layer achieves the optimal balance of efficiency and accuracy.

\paragraph{Feature selection.} 
As mentioned in \cref{sec:method:algm}, our approach utilizes the cosine similarity of token embeddings $X$ to identify tokens for merging. Here, we examine the effect of using cosine similarity of other features -- \ie, keys, queries, and values -- to determine which tokens can be merged. Table~\ref{tab:ablations:feature_selection} shows the results. Although using keys or queries results in the highest throughput, using the tokens $X$ yields a considerably higher mIoU and a throughput that is close to the throughput obtained by using keys or queries. This differs from the findings for ToMe~\cite{bolya2023tome}, where the keys are identified as the best option to identify mergeable tokens for the image classification task. We hypothesize that this can be attributed to the fact that all token embeddings are directly used to make the final semantic segmentation prediction, whereas image classification networks only use a single \textit{CLS} token for the final class prediction~\cite{dosovitskiy2021an}. This gives the tokens a more important role for semantic segmentation, making them the most appropriate feature to use for token reduction.

\section{Qualitative results}
\label{Sup:sec:qualitative_results}
\begin{figure*}[t]
\centering
\includegraphics[width=0.24\linewidth]{X_suppl/images/Vis_simpam_local_global/ade20k/ADE_val_00000118.png}
\includegraphics[width=0.24\linewidth]{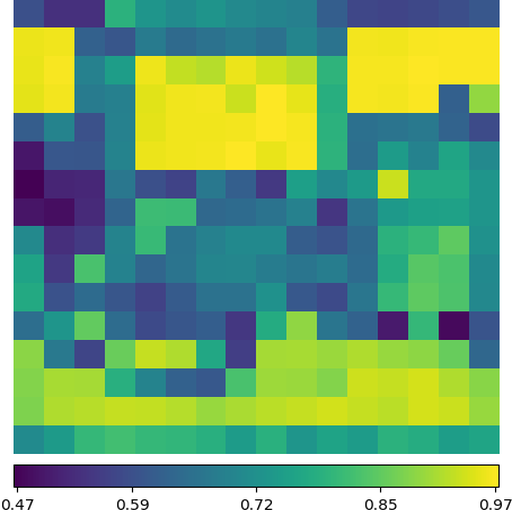}
\includegraphics[width=0.24\linewidth]{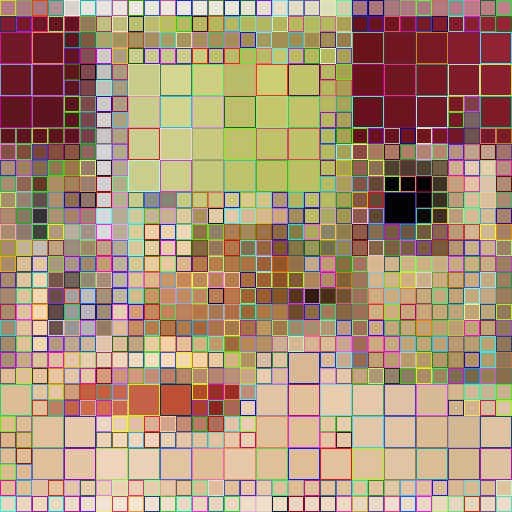}
\includegraphics[width=0.24\linewidth]{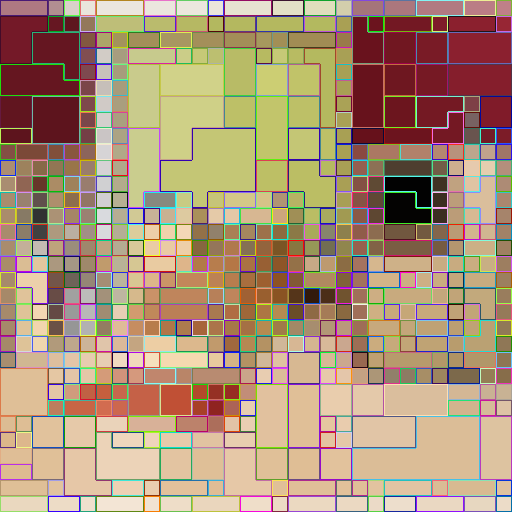}\\

\includegraphics[width=0.24\linewidth]{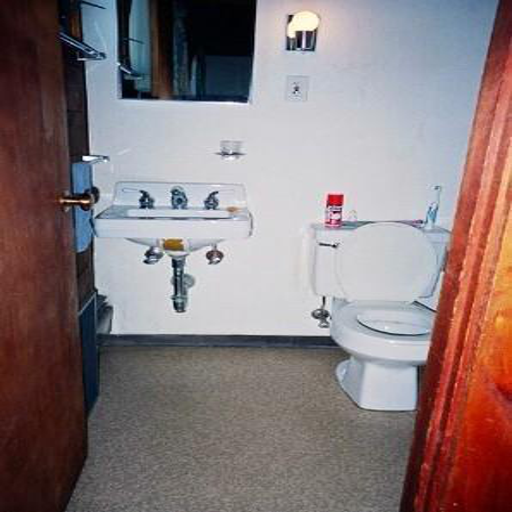}
\includegraphics[width=0.24\linewidth]{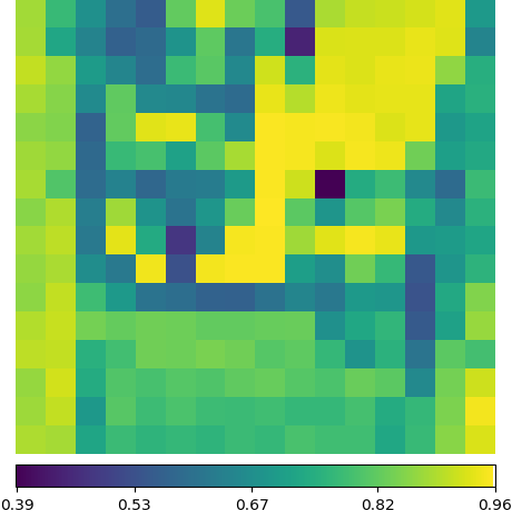}
\includegraphics[width=0.24\linewidth]{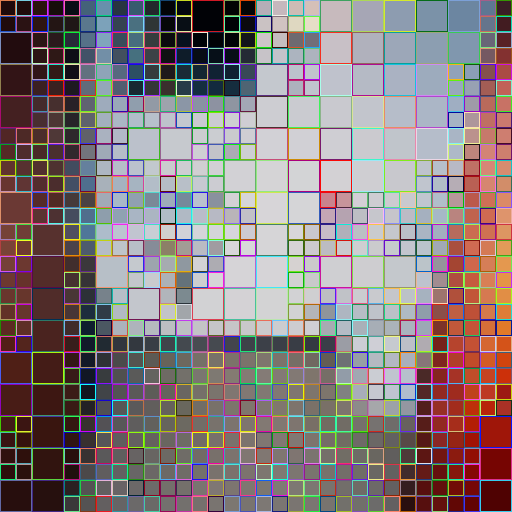}
\includegraphics[width=0.24\linewidth]{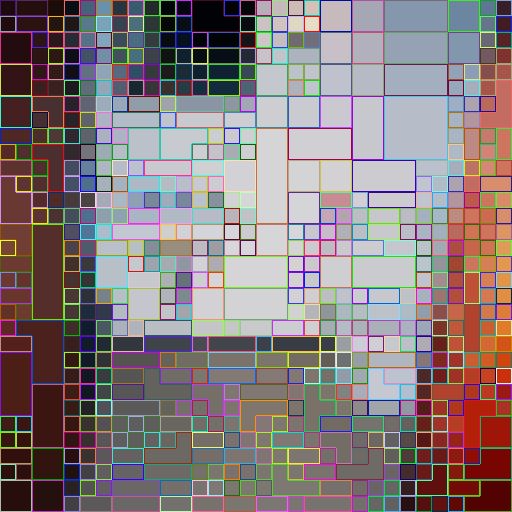}
\\
\includegraphics[width=0.24\linewidth]{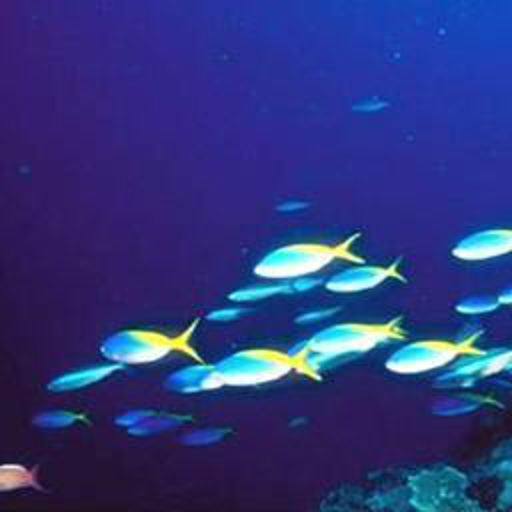}
\includegraphics[width=0.24\linewidth]{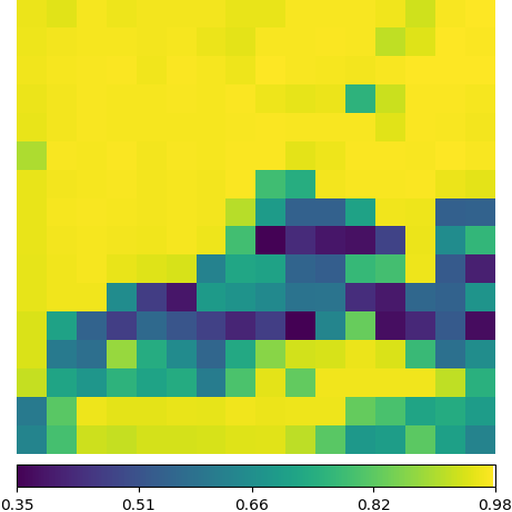}
\includegraphics[width=0.24\linewidth]{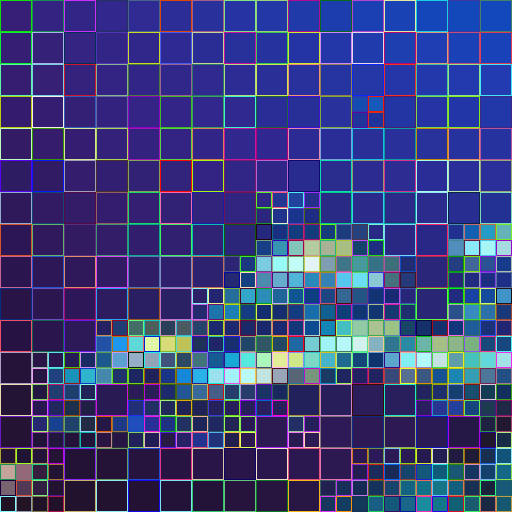}
\includegraphics[width=0.24\linewidth]{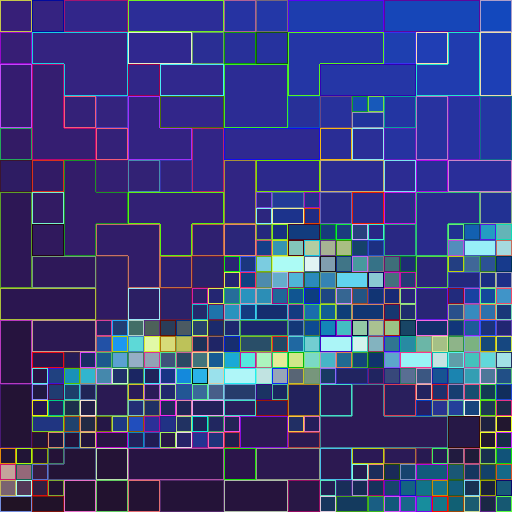}
 \\

\includegraphics[width=0.24\linewidth]{X_suppl/images/Vis_simpam_local_global/ade20k/ADE_val_00001874.png}
\includegraphics[width=0.24\linewidth]{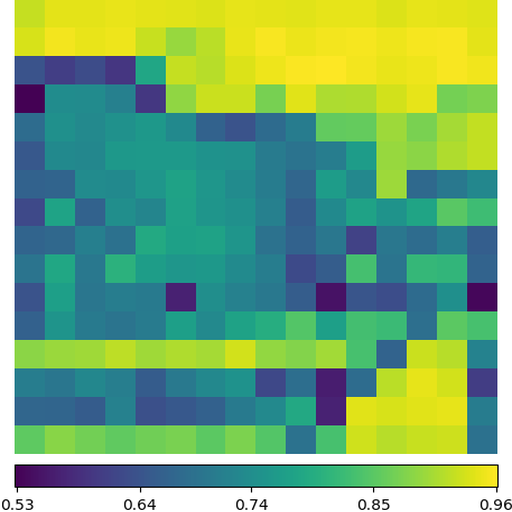}
\includegraphics[width=0.24\linewidth]{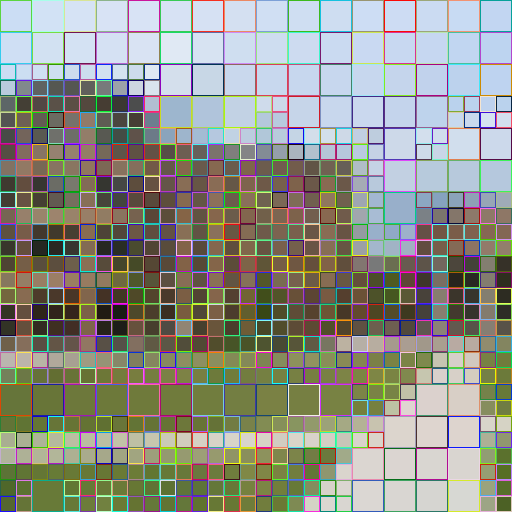}
\includegraphics[width=0.24\linewidth]{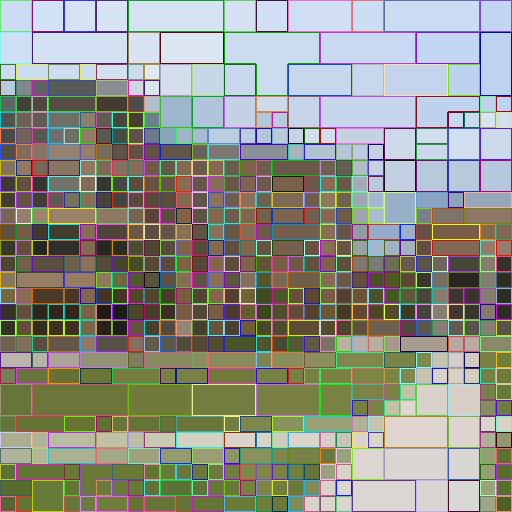}
\\

\includegraphics[width=0.24\linewidth]{X_suppl/images/Vis_simpam_local_global/ade20k/ADE_val_00000090.png}
\includegraphics[width=0.24\linewidth]{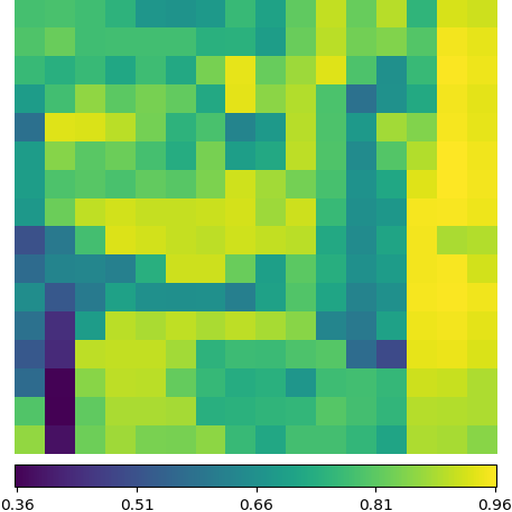}
\includegraphics[width=0.24\linewidth]{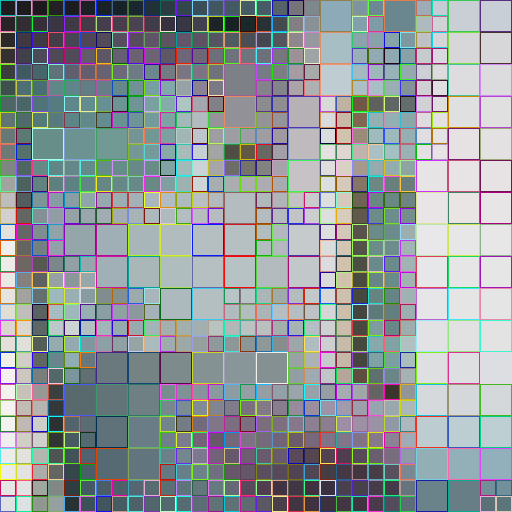}
\includegraphics[width=0.24\linewidth]{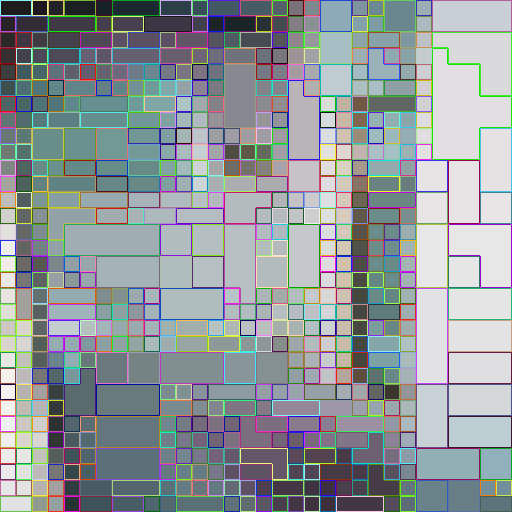}
\\

 \begin{subfigure}[b]{0.24\textwidth}
     \centering
     \caption{Input image}
 \end{subfigure}
 \begin{subfigure}[b]{0.24\textwidth}
     \centering
     \caption{Local token similarities}
 \end{subfigure}
  \begin{subfigure}[b]{0.24\textwidth}
     \centering
     \caption{Local merging}
 \end{subfigure}
 \begin{subfigure}[b]{0.24\textwidth}
     \centering
     \caption{Global merging}
 \end{subfigure}

\caption{\textbf{Qualitative results on ADE20K~\cite{zhou2017ade20k}.} This figure displays (a) the input image, (b) the average cosine similarities between tokens in 2$\times$2 local windows in the first transformer layer, (c) the merged tokens after CLAP local merging, (d) the merged tokens after GBM global merging.} 
\label{fig:simmap_local_global_qual_results_ade20k}
\end{figure*}

\begin{figure*}[t]
\centering

\includegraphics[width=0.24\linewidth]{X_suppl/images/Vis_simpam_local_global/cocostuff/COCO_train2014_000000011077.png}
\includegraphics[width=0.24\linewidth]{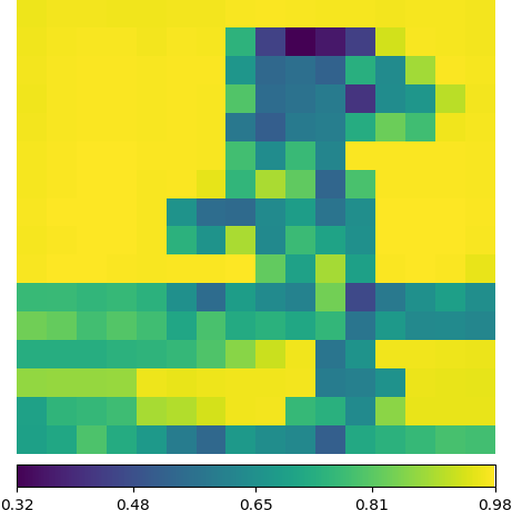}
\includegraphics[width=0.24\linewidth]{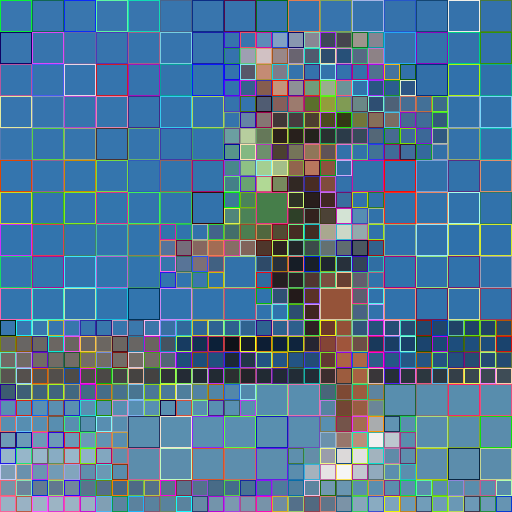}
\includegraphics[width=0.24\linewidth]{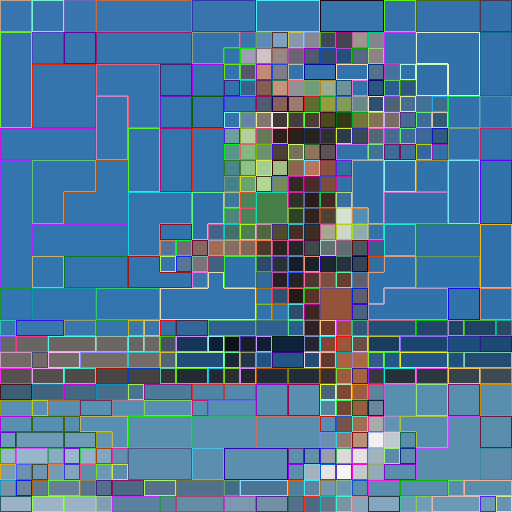}
\\

\includegraphics[width=0.24\linewidth]{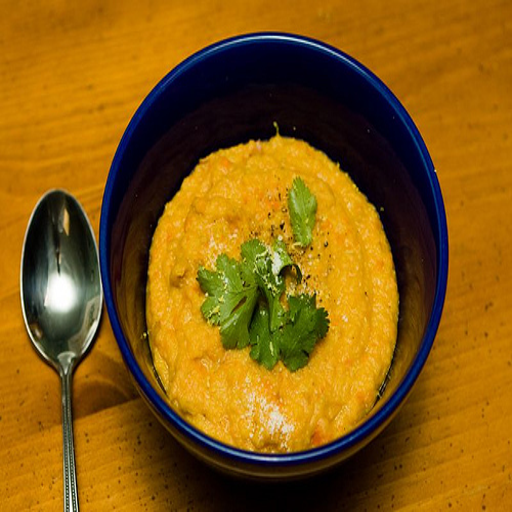}
\includegraphics[width=0.24\linewidth]{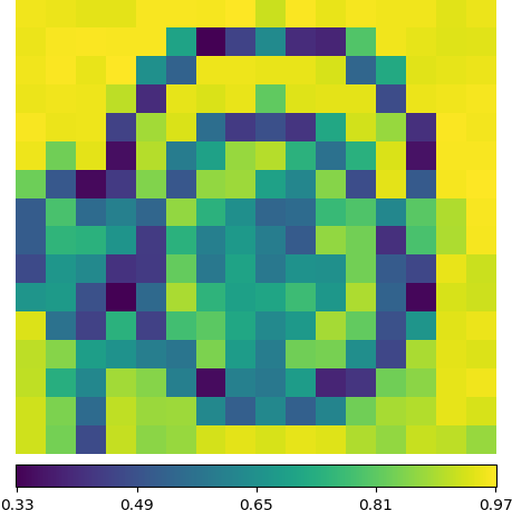}
\includegraphics[width=0.24\linewidth]{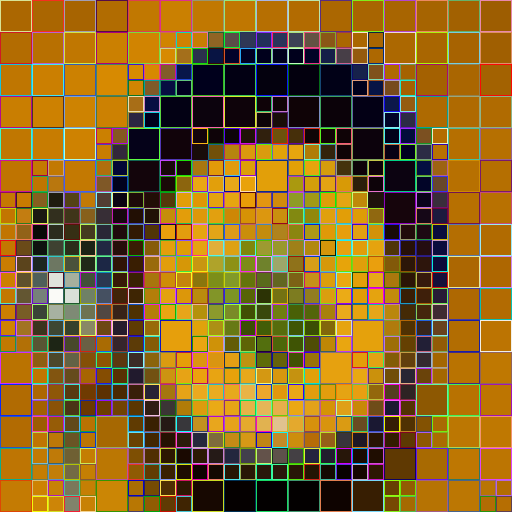}
\includegraphics[width=0.24\linewidth]{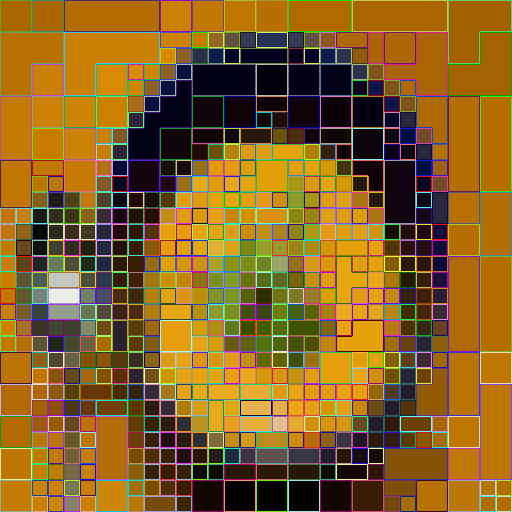}
\\

\includegraphics[width=0.24\linewidth]{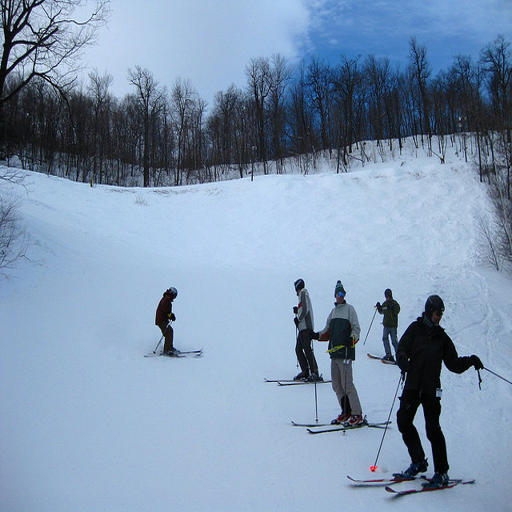}
\includegraphics[width=0.24\linewidth]{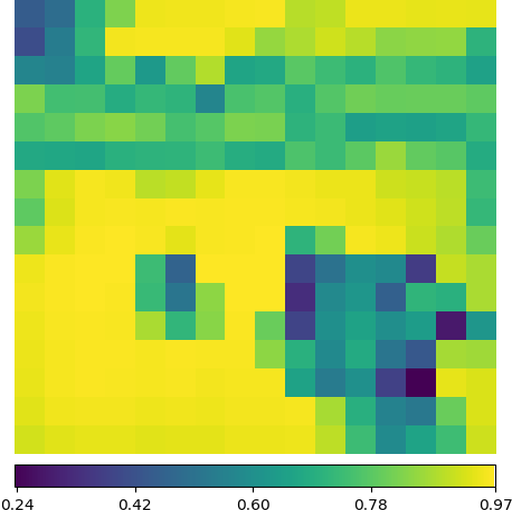}
\includegraphics[width=0.24\linewidth]{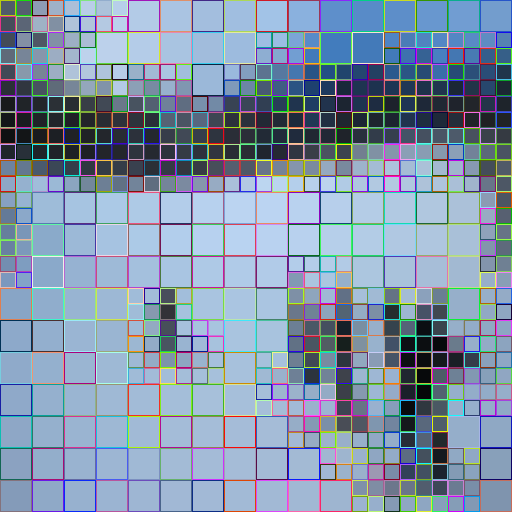}
\includegraphics[width=0.24\linewidth]{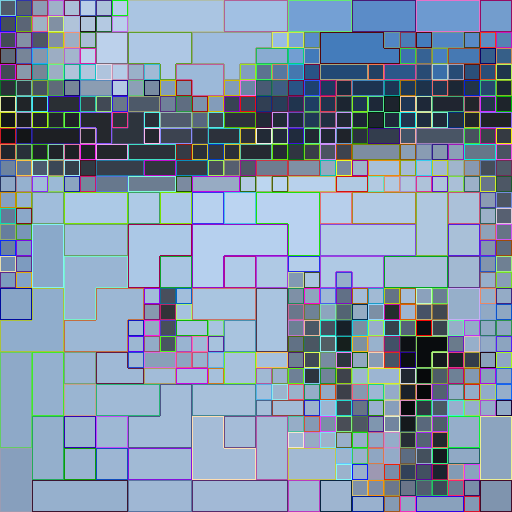}
 \\

\includegraphics[width=0.24\linewidth]{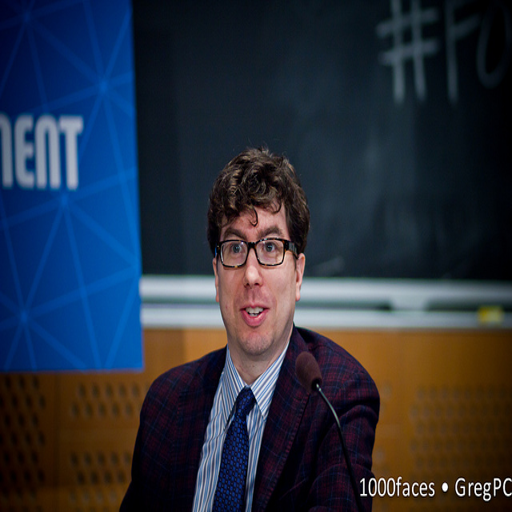}
\includegraphics[width=0.24\linewidth]{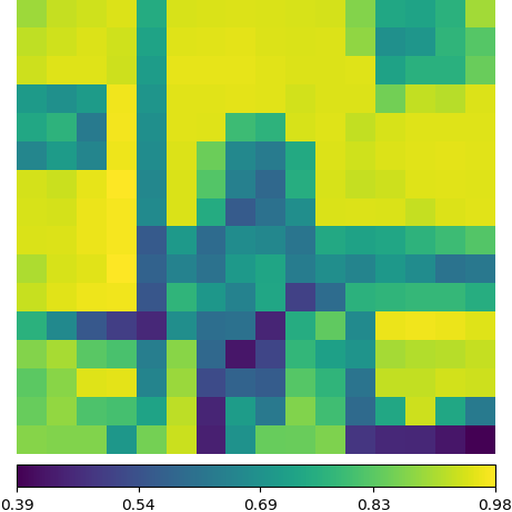}
\includegraphics[width=0.24\linewidth]{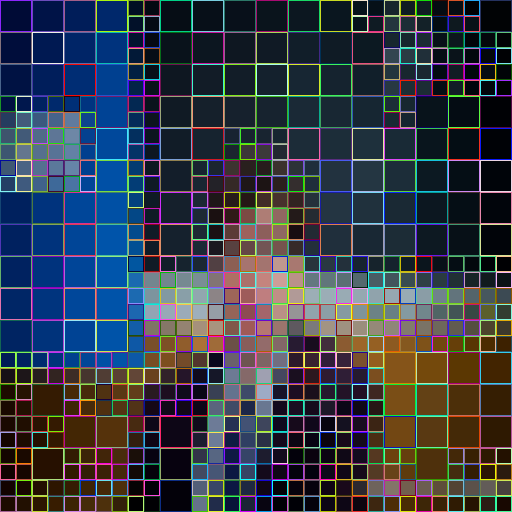}
\includegraphics[width=0.24\linewidth]{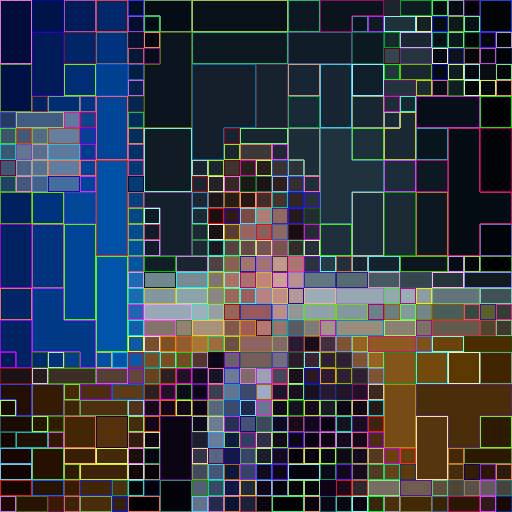}
 \\

\includegraphics[width=0.24\linewidth]{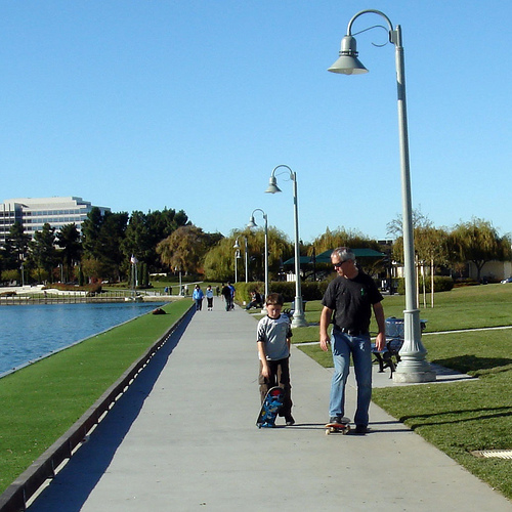}
\includegraphics[width=0.24\linewidth]{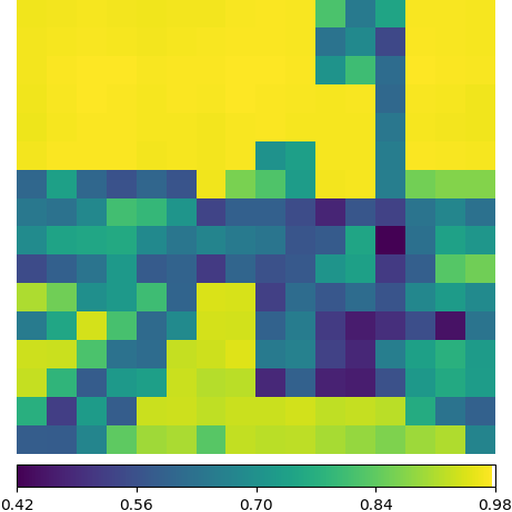}
\includegraphics[width=0.24\linewidth]{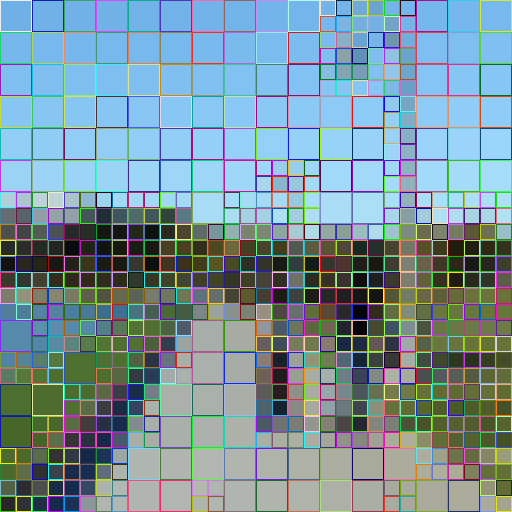}
\includegraphics[width=0.24\linewidth]{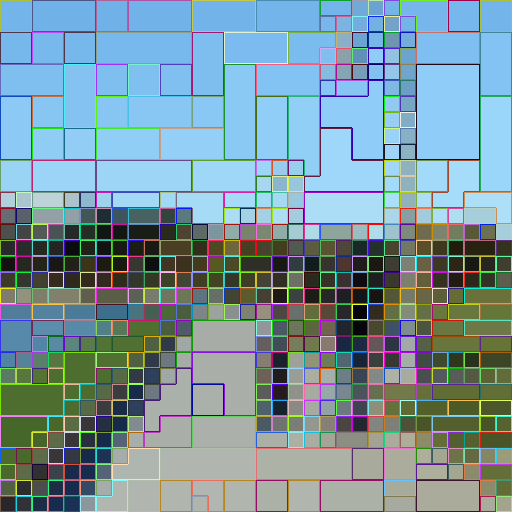}
 \\

 \begin{subfigure}[b]{0.24\textwidth}
     \centering
     \caption{Input image}
 \end{subfigure}
 \begin{subfigure}[b]{0.24\textwidth}
     \centering
     \caption{Local token similarities}
 \end{subfigure}
  \begin{subfigure}[b]{0.24\textwidth}
     \centering
     \caption{Local merging}
 \end{subfigure}
 \begin{subfigure}[b]{0.24\textwidth}
     \centering
     \caption{Global merging}
 \end{subfigure}

\caption{\textbf{Qualitative results on COCO-Stuff~\cite{caesar2018cocostuff}.} This figure displays (a) the input image, (b) the average cosine similarities between tokens in 2$\times$2 local windows in the first transformer layer, (c) the merged tokens after CLAP local merging, (d) the merged tokens after GBM global merging.}
\label{fig:simmap_local_global_qual_results_cocostuff}
\end{figure*}

\begin{figure*}[t]
\centering

\includegraphics[width=0.24\linewidth]{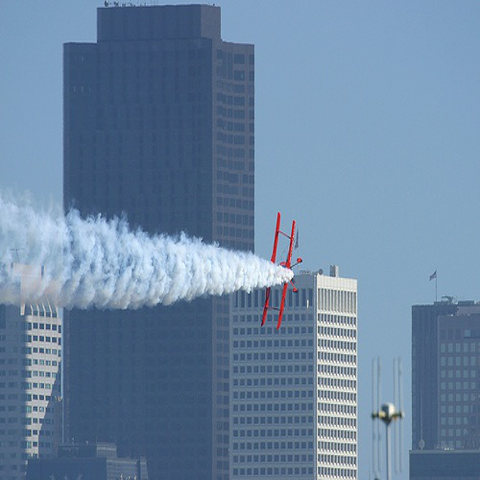}
\includegraphics[width=0.24\linewidth]{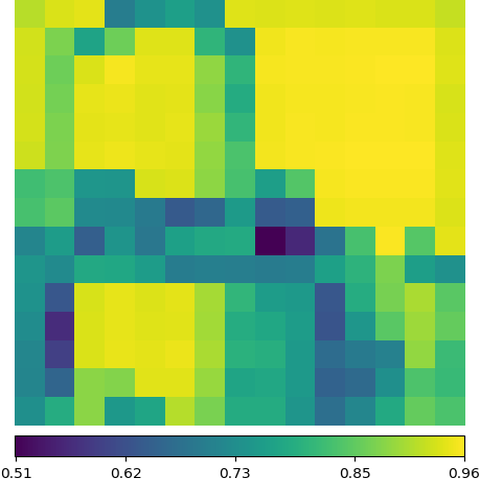}
\includegraphics[width=0.24\linewidth]{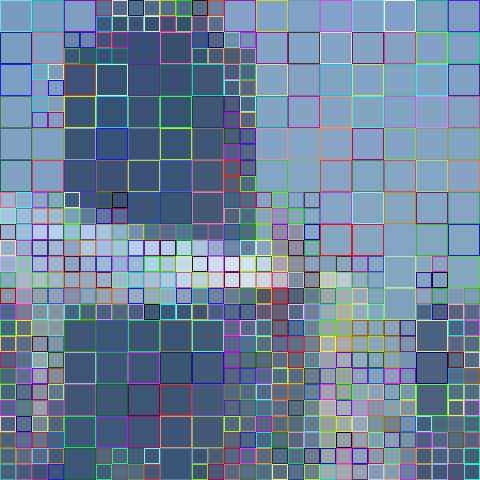}
\includegraphics[width=0.24\linewidth]{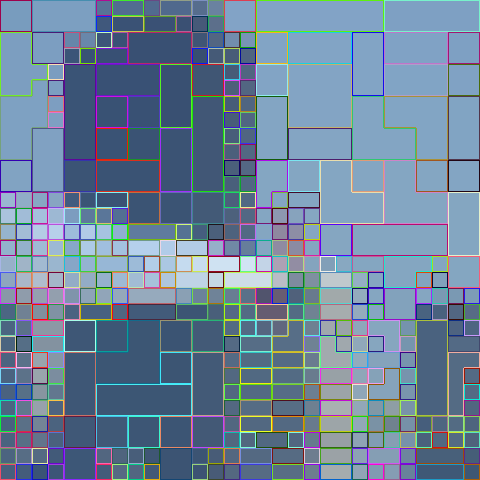}
\\

\includegraphics[width=0.24\linewidth]{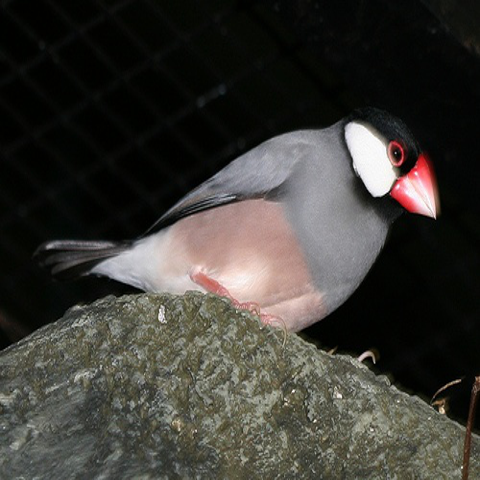}
\includegraphics[width=0.24\linewidth]{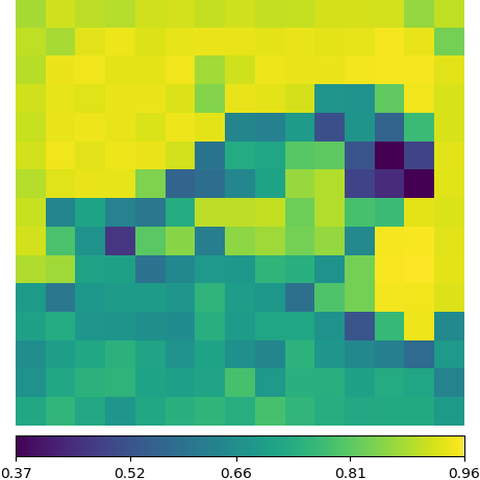}
\includegraphics[width=0.24\linewidth]{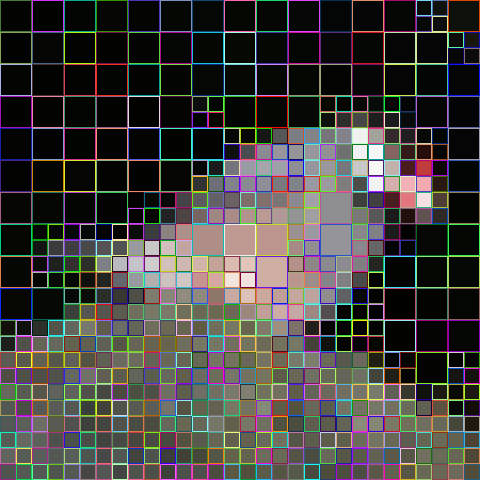}
\includegraphics[width=0.24\linewidth]{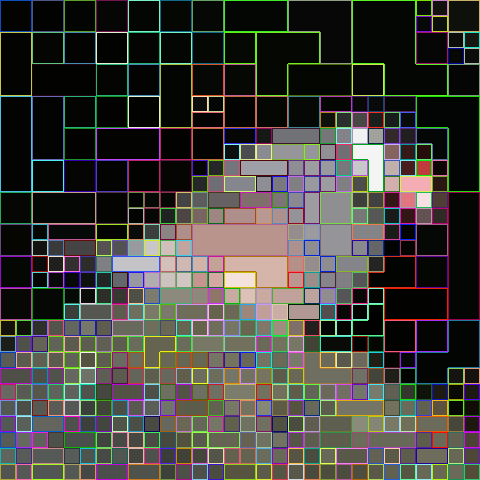}
\\

\includegraphics[width=0.24\linewidth]{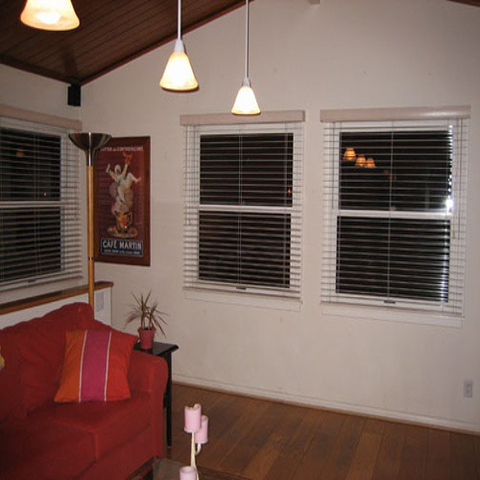}
\includegraphics[width=0.24\linewidth]{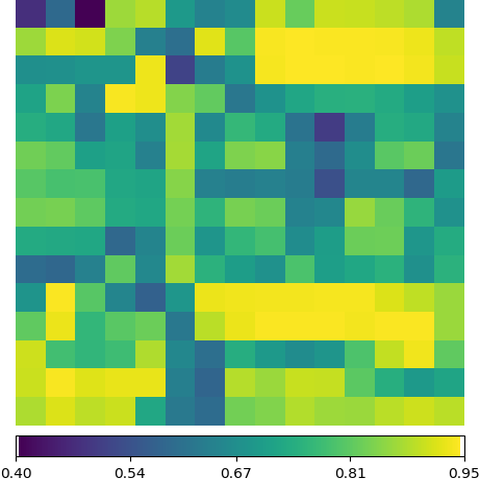}
\includegraphics[width=0.24\linewidth]{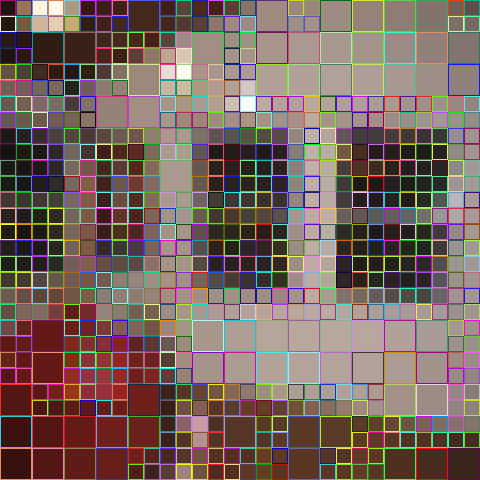}
\includegraphics[width=0.24\linewidth]{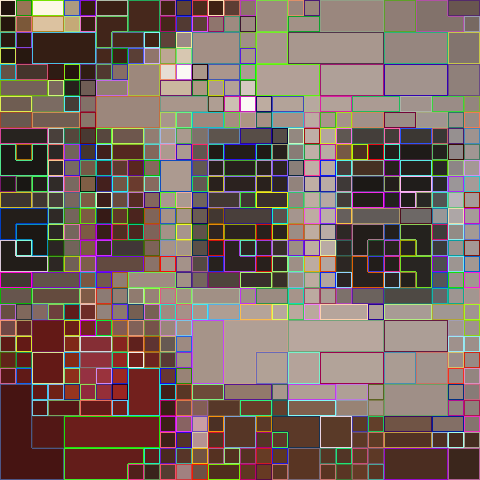}
 \\

\includegraphics[width=0.24\linewidth]{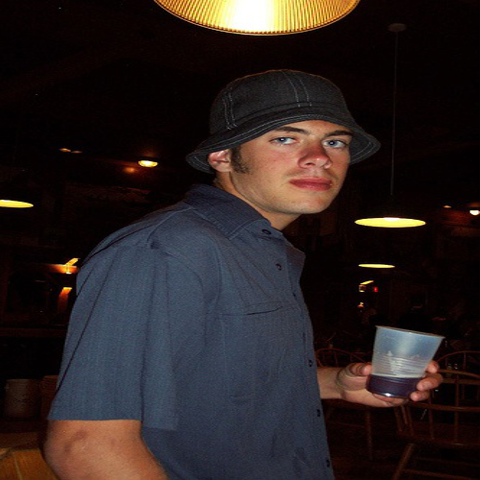}
\includegraphics[width=0.24\linewidth]{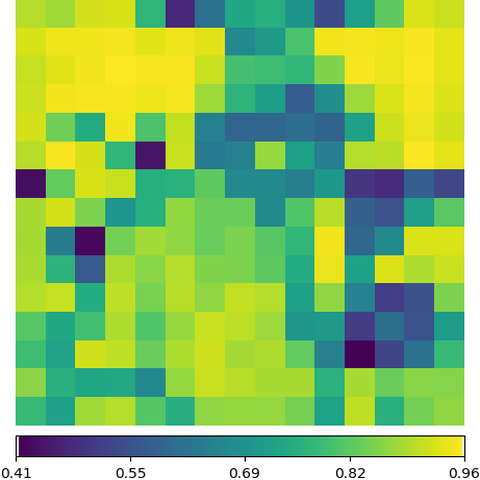}
\includegraphics[width=0.24\linewidth]{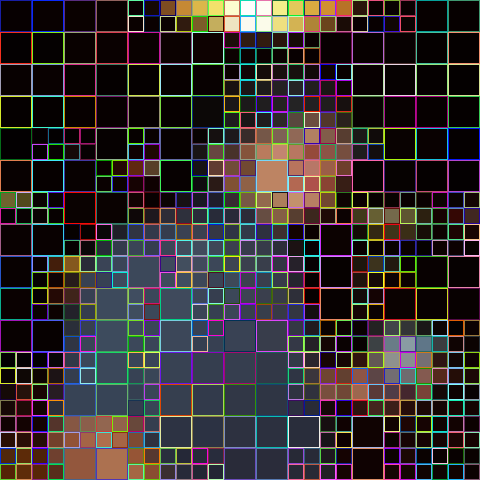}
\includegraphics[width=0.24\linewidth]{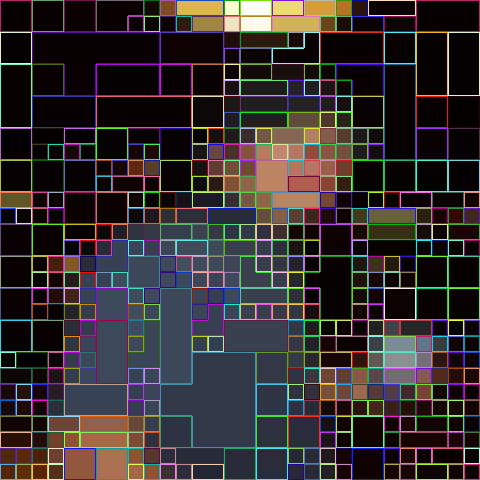}
 \\

\includegraphics[width=0.24\linewidth]{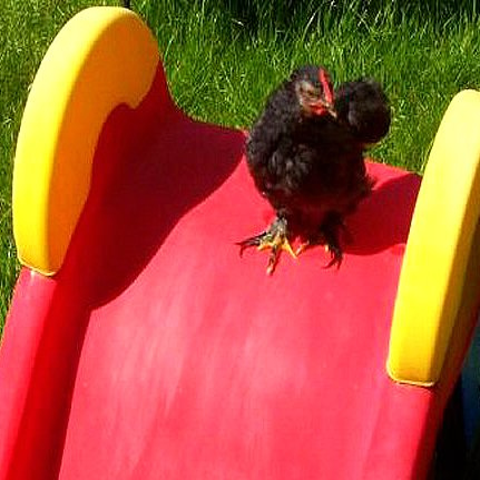}
\includegraphics[width=0.24\linewidth]{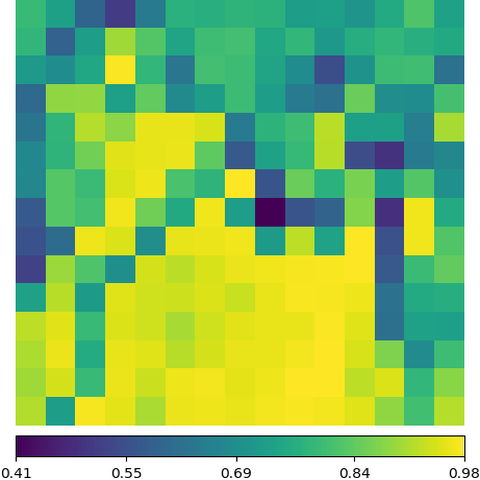}
\includegraphics[width=0.24\linewidth]{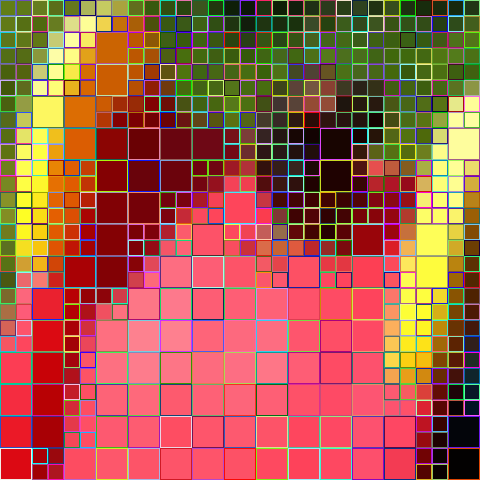}
\includegraphics[width=0.24\linewidth]{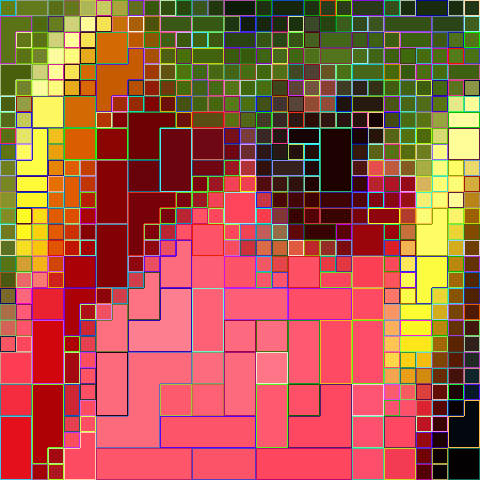}
 \\

 \begin{subfigure}[b]{0.24\textwidth}
     \centering
     \caption{Input image}
 \end{subfigure}
 \begin{subfigure}[b]{0.24\textwidth}
     \centering
     \caption{Local token similarities}
 \end{subfigure}
  \begin{subfigure}[b]{0.24\textwidth}
     \centering
     \caption{Local merging}
 \end{subfigure}
 \begin{subfigure}[b]{0.24\textwidth}
     \centering
     \caption{Global merging}
 \end{subfigure}

\caption{\textbf{Qualitative results on Pascal-Context~\cite{mottaghi2014pascalcontext}.} This figure displays (a) the input image, (b) the average cosine similarities between tokens in 2$\times$2 local windows in the first transformer layer, (c) the merged tokens after CLAP local merging, (d) the merged tokens after GBM global merging.}
\label{fig:simmap_local_global_qual_results_pcontext}
\end{figure*}

\begin{figure*}[t]
\centering

\includegraphics[width=0.24\linewidth]{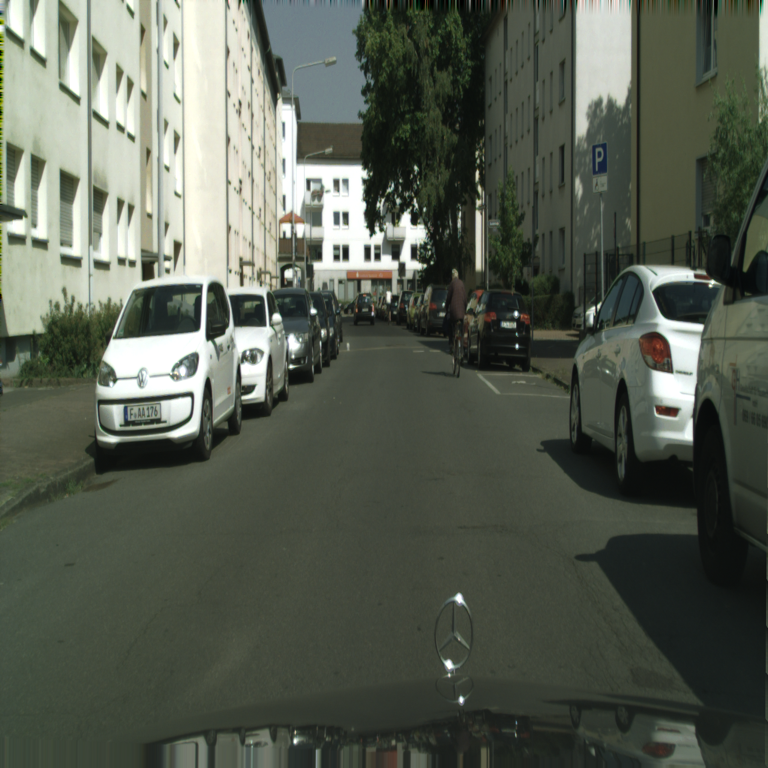}
\includegraphics[width=0.24\linewidth]{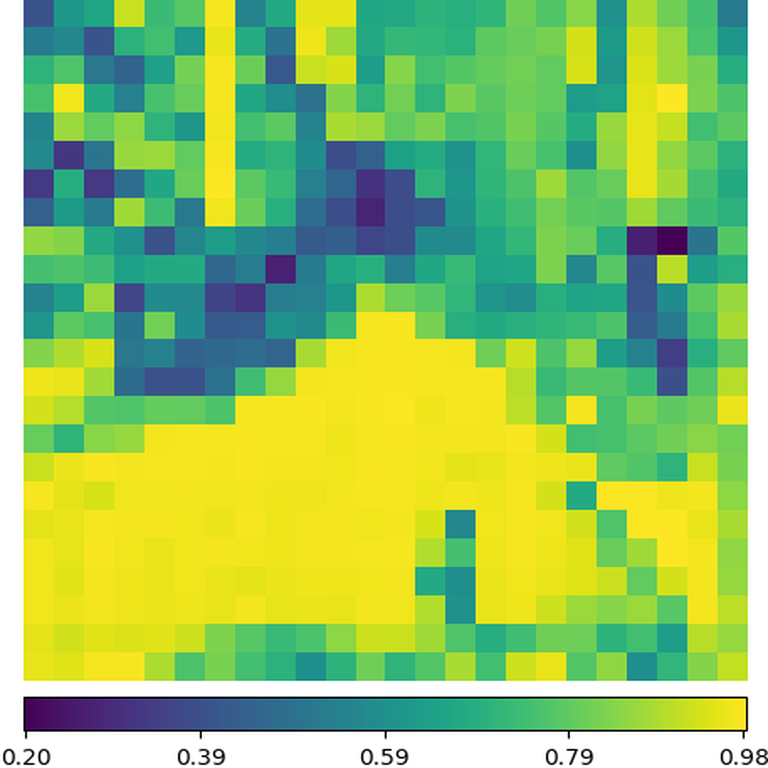}
\includegraphics[width=0.24\linewidth]{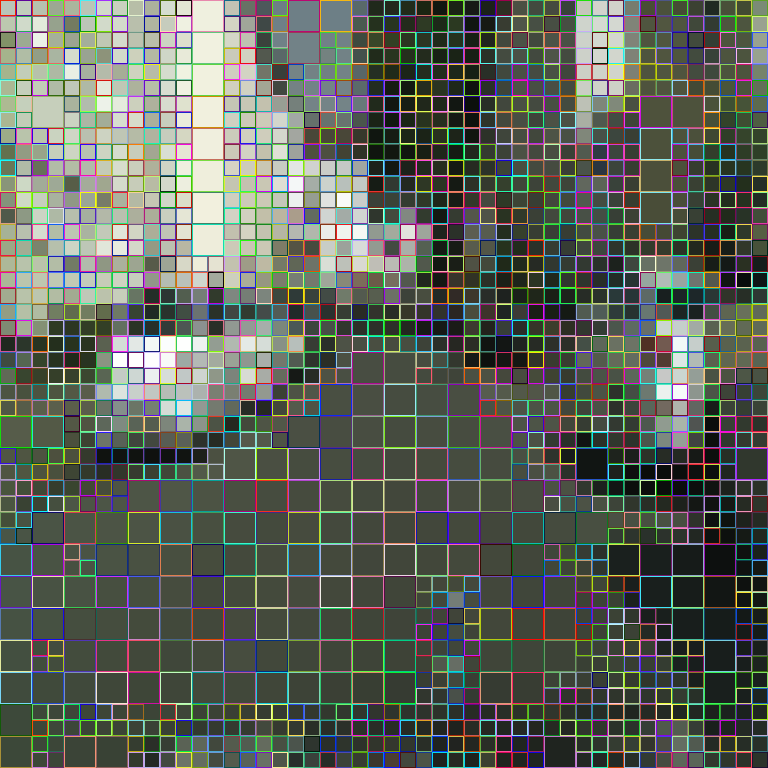}
\includegraphics[width=0.24\linewidth]{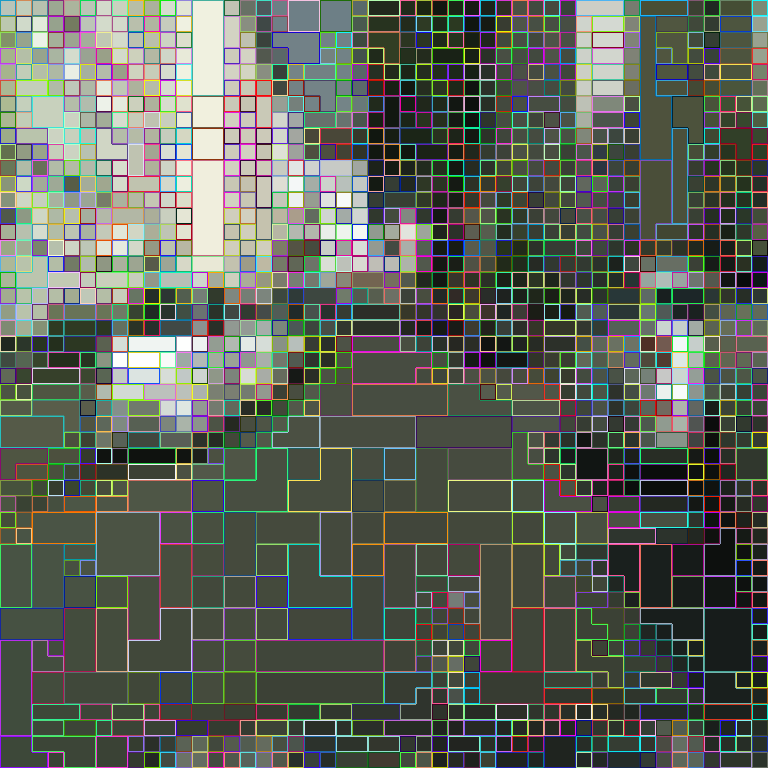}
\\

\includegraphics[width=0.24\linewidth]{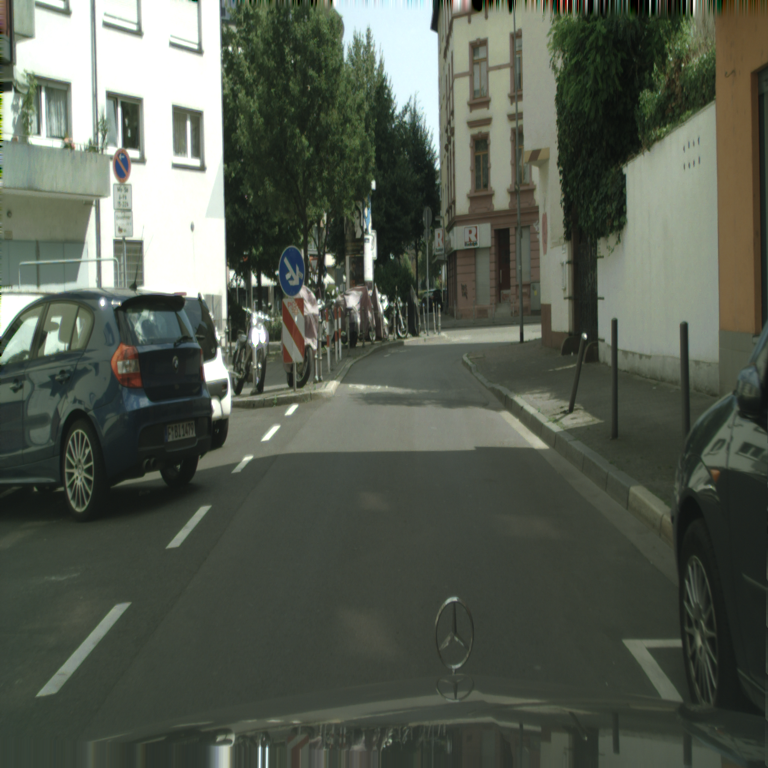}
\includegraphics[width=0.24\linewidth]{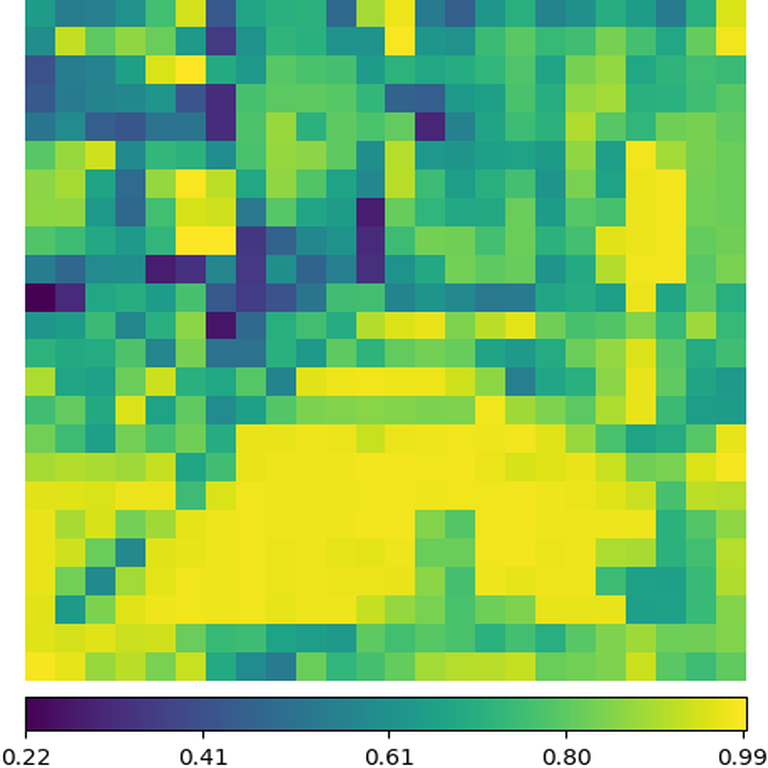}
\includegraphics[width=0.24\linewidth]{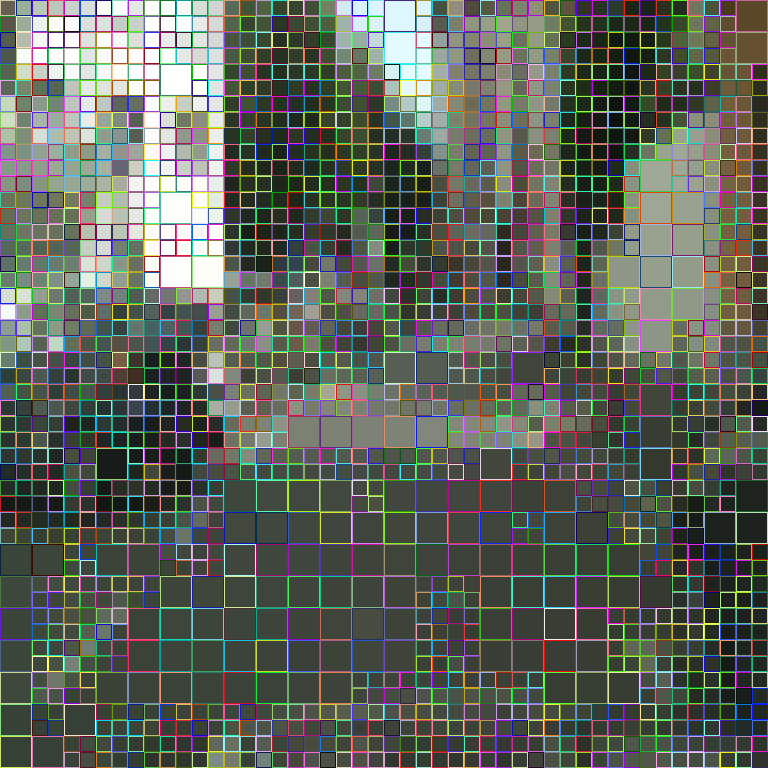}
\includegraphics[width=0.24\linewidth]{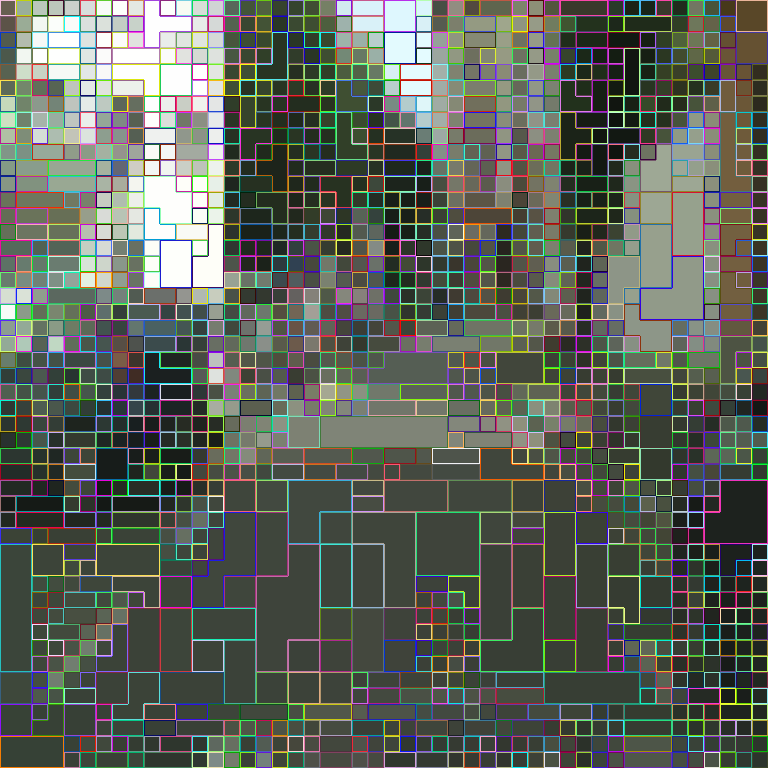}
\\

\includegraphics[width=0.24\linewidth]{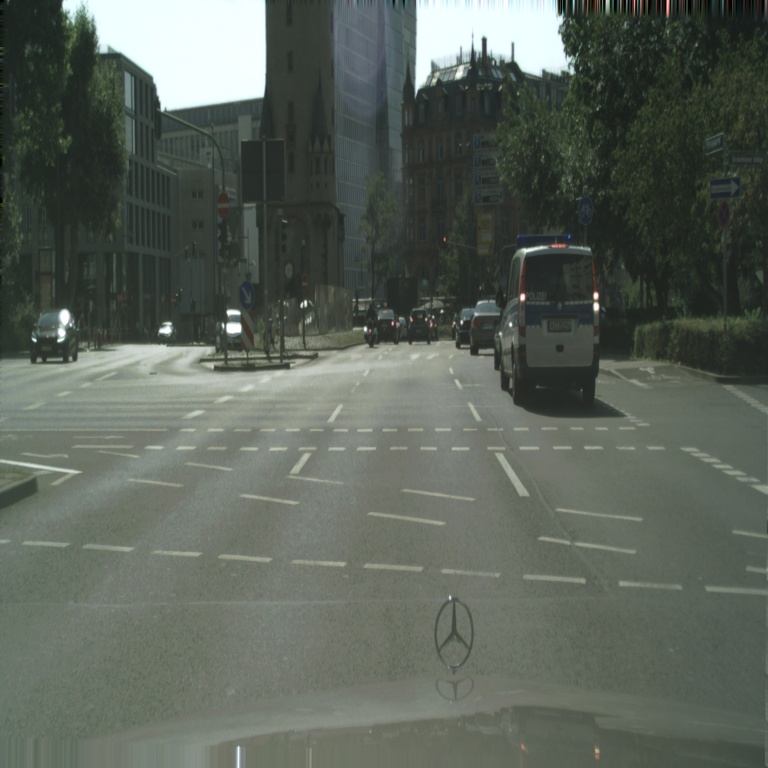}
\includegraphics[width=0.24\linewidth]{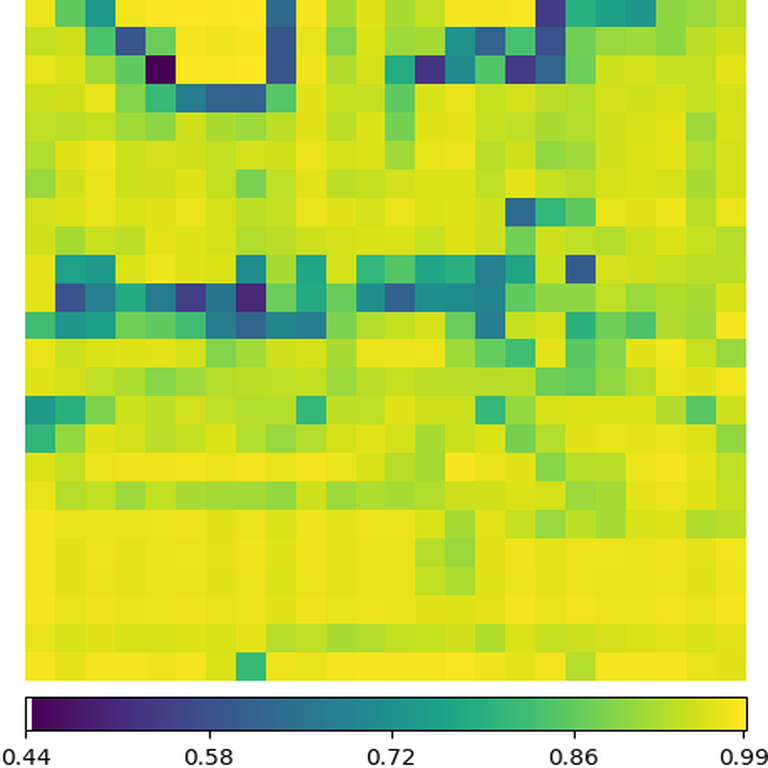}
\includegraphics[width=0.24\linewidth]{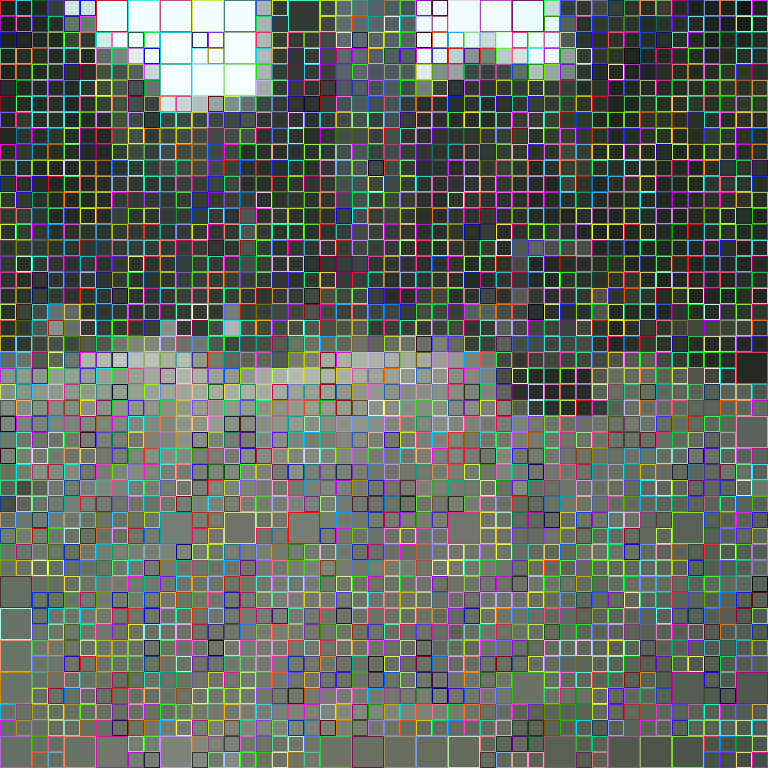}
\includegraphics[width=0.24\linewidth]{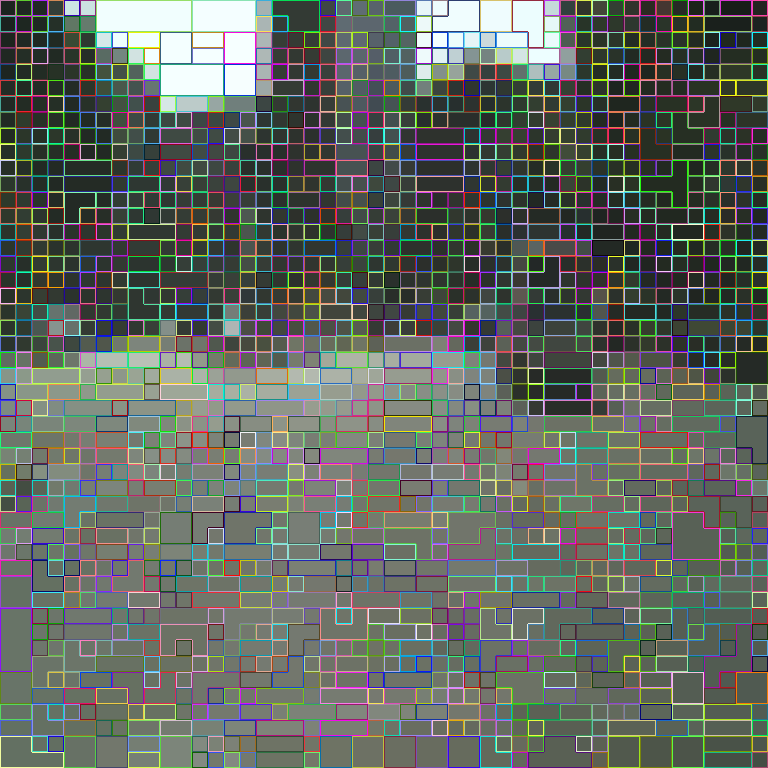}
 \\

\includegraphics[width=0.24\linewidth]{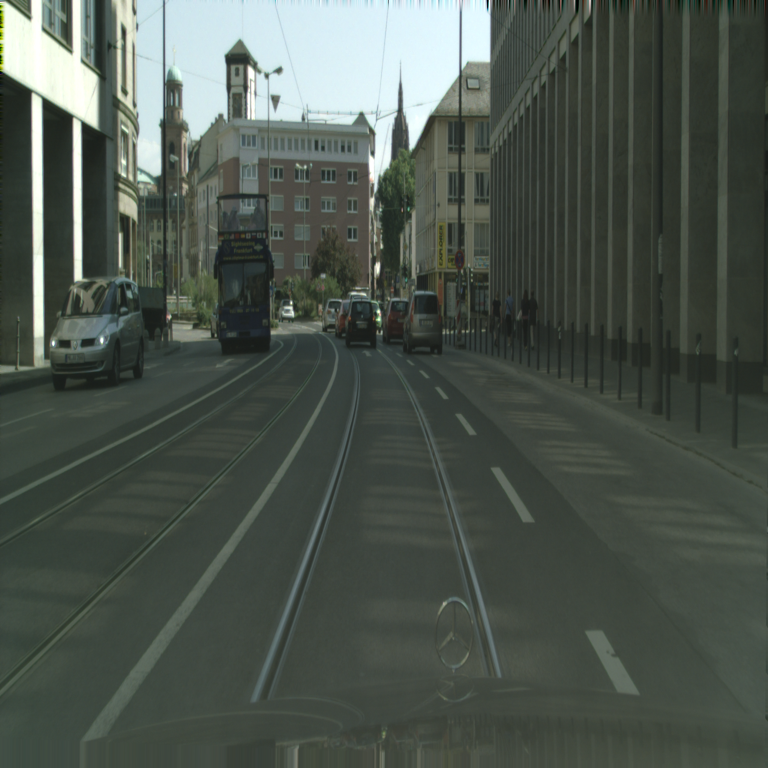}
\includegraphics[width=0.24\linewidth]{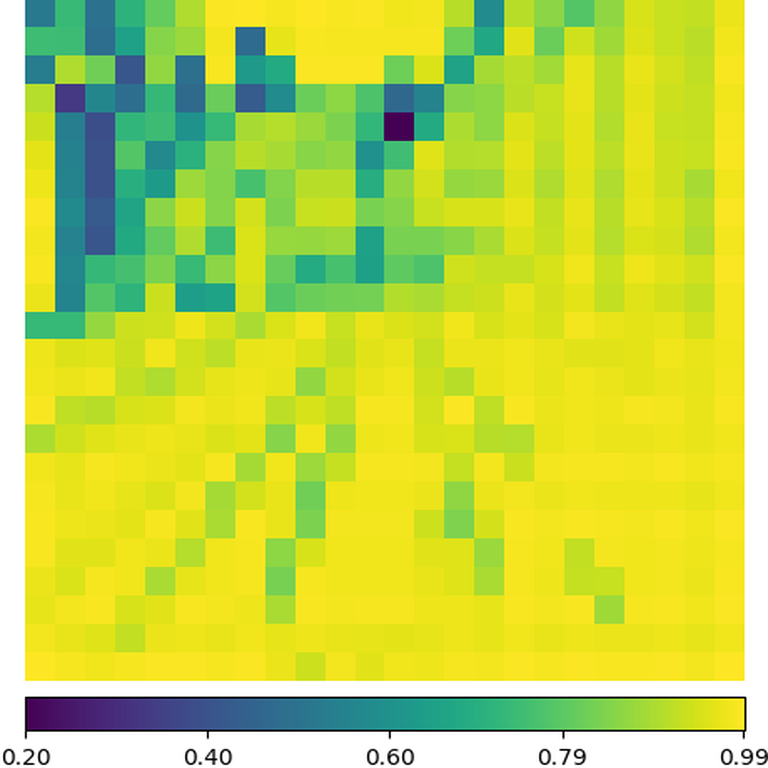}
\includegraphics[width=0.24\linewidth]{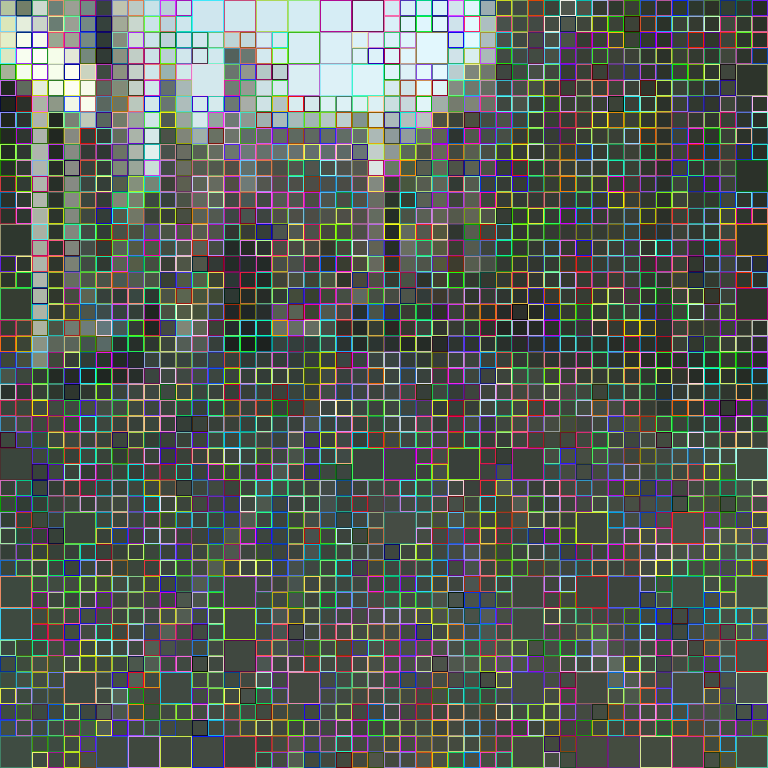}
\includegraphics[width=0.24\linewidth]{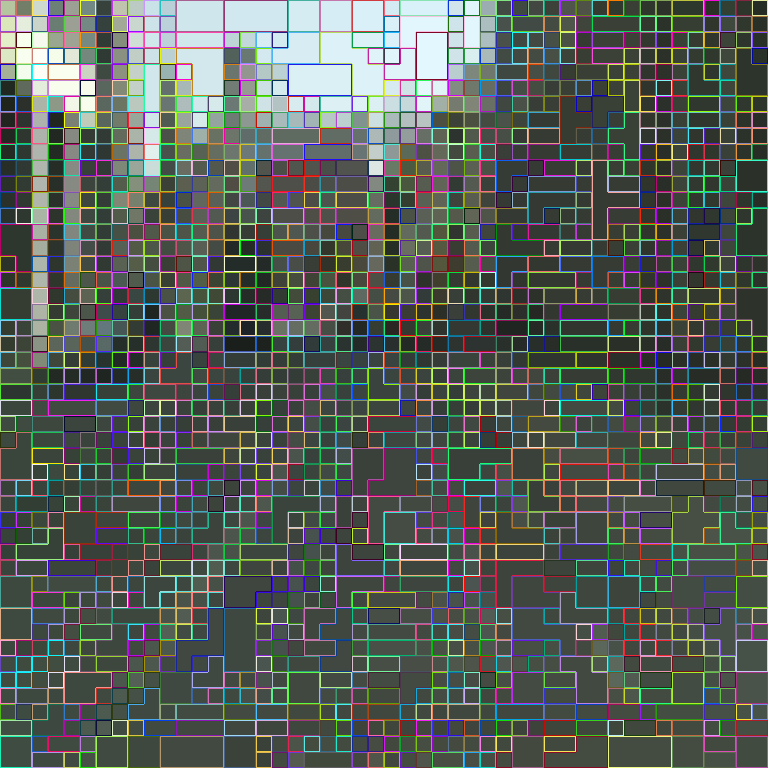}
 \\

 \begin{subfigure}[b]{0.24\textwidth}
     \centering
     \caption{Input image}
 \end{subfigure}
 \begin{subfigure}[b]{0.24\textwidth}
     \centering
     \caption{Local token similarities}
 \end{subfigure}
  \begin{subfigure}[b]{0.24\textwidth}
     \centering
     \caption{Local merging}
 \end{subfigure}
 \begin{subfigure}[b]{0.24\textwidth}
     \centering
     \caption{Global merging}
 \end{subfigure}

\caption{\textbf{Qualitative results on Cityscapes~\cite{cordts2016cityscapes}.} This figure displays (a) the input image, (b) the average cosine similarities between tokens in 2$\times$2 local windows in the first transformer layer, (c) the merged tokens after CLAP local merging, (d) the merged tokens after GBM global merging.}
\label{fig:simmap_local_global_qual_results_cityscapes}
\end{figure*}
\subsection{Merging operations} In \cref{fig:simmap_local_global_qual_results_ade20k}, \cref{fig:simmap_local_global_qual_results_cocostuff}, \cref{fig:simmap_local_global_qual_results_pcontext}, and \cref{fig:simmap_local_global_qual_results_cityscapes}, we show qualitative examples of the effect of CLAP local merging and GBM global merging. The similarity map displays the average cosine similarity between tokens within a 2$\times$2 window, which is used to determine which tokens can be merged in the CLAP module. These figures demonstrate that for all datasets, CLAP predominantly merges tokens with a high visual similarity. On the other hand, GBM also merges tokens that depict the same category but have lower visual similarity, after acquiring more informative embeddings in the middle network layers.
Notably, from row 3 and 4 in \cref{fig:simmap_local_global_qual_results_cityscapes}, depicting Cityscapes examples, it becomes clear that the high visual homogeneity of Cityscapes images causes tokens to have high cosine similarities also if they do not depict the same class. As mentioned in \cref{sec:results:main} and \cref{sup:sec:ALGM_existing_works_various_different_settings}, this high average cosine similarity causes the automatic threshold $\tau$ to be quite high, limiting the efficiency improvement of the ALGM method on this dataset.
     
\subsection{Semantic segmentation predictions}  
\begin{figure*}[t]
\centering
\includegraphics[width=0.19\linewidth]{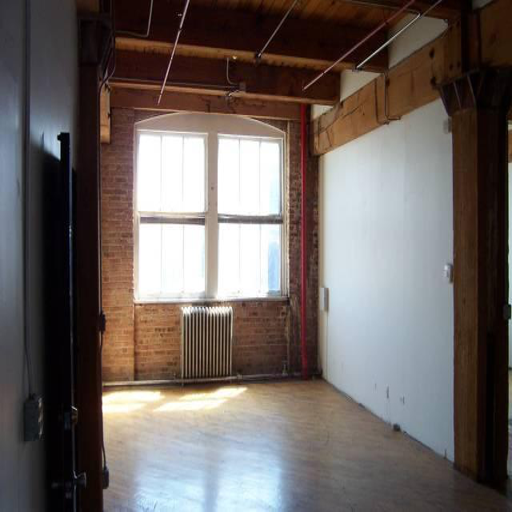}
\includegraphics[width=0.19\linewidth]{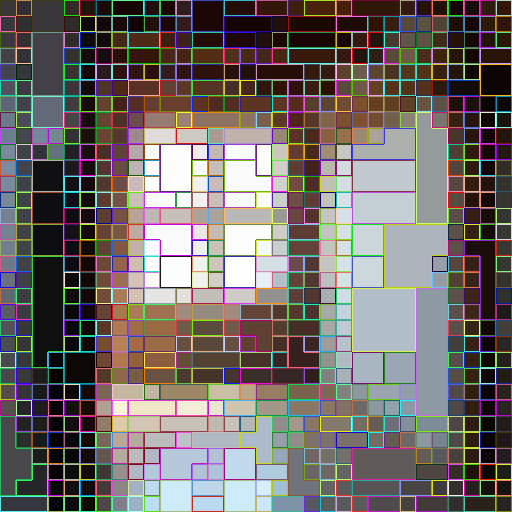}
\includegraphics[width=0.19\linewidth]{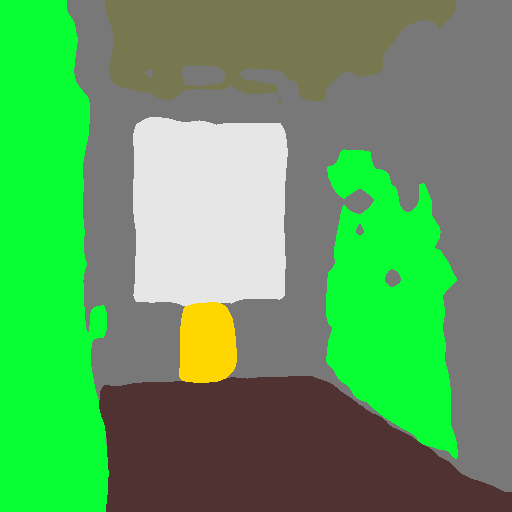}
\includegraphics[width=0.19\linewidth]{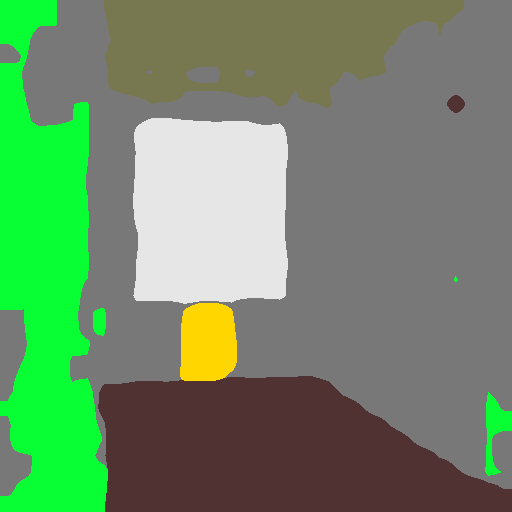}
\includegraphics[width=0.19\linewidth]{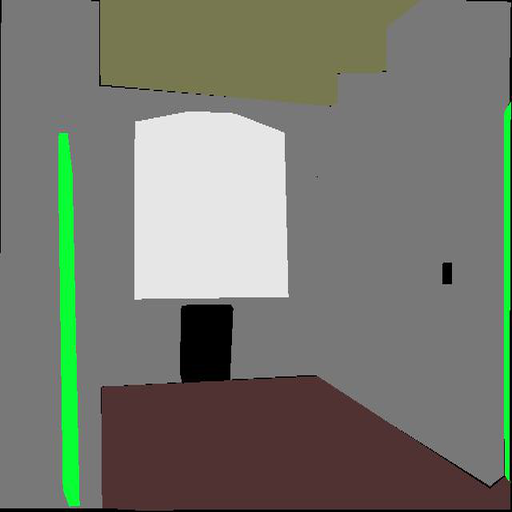}
\\

\includegraphics[width=0.19\linewidth]{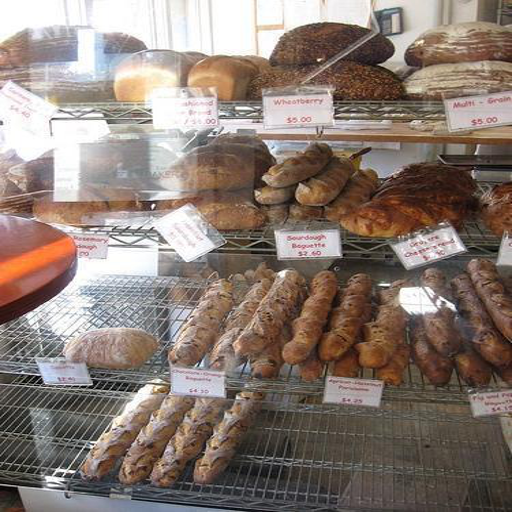}
\includegraphics[width=0.19\linewidth]{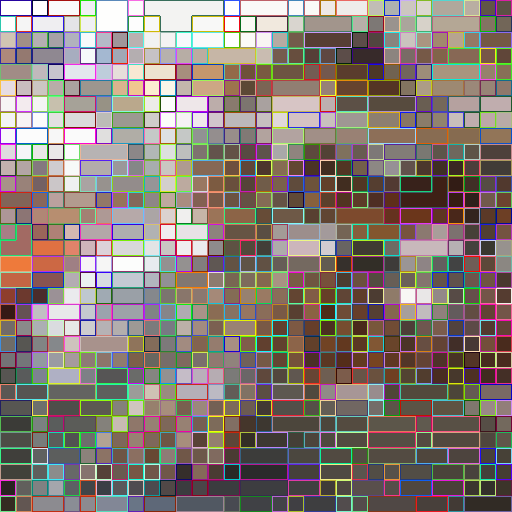}
\includegraphics[width=0.19\linewidth]{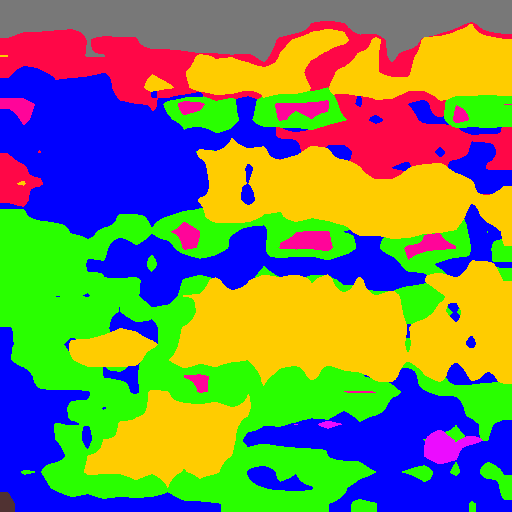}
\includegraphics[width=0.19\linewidth]{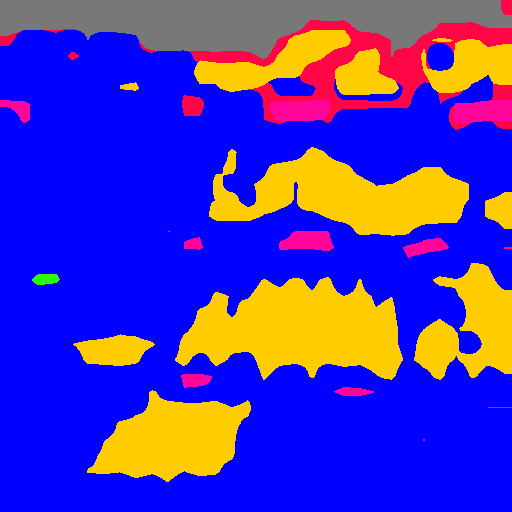}
\includegraphics[width=0.19\linewidth]{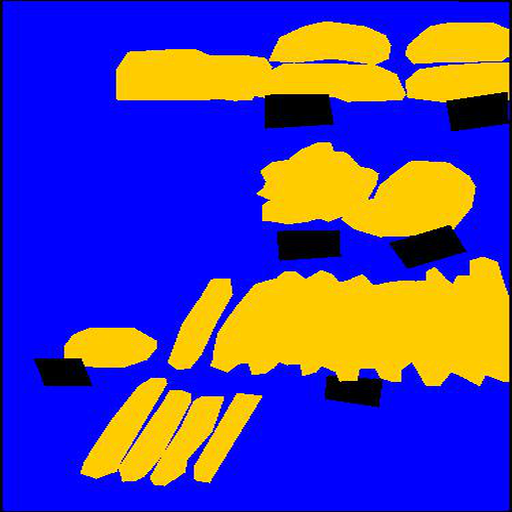}
\\

\includegraphics[width=0.19\linewidth]{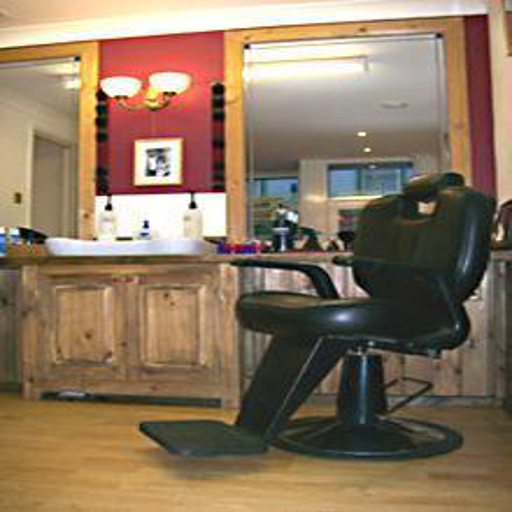}
\includegraphics[width=0.19\linewidth]{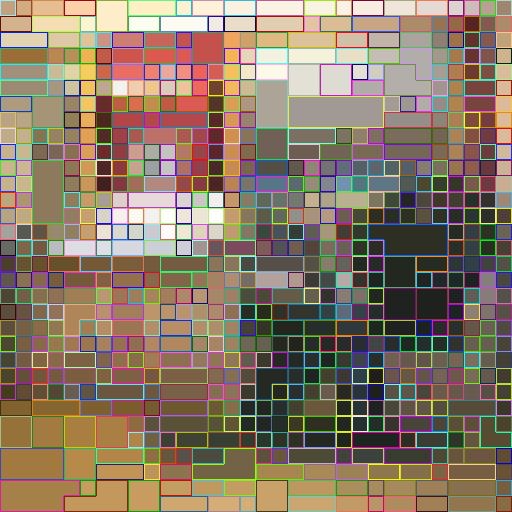}
\includegraphics[width=0.19\linewidth]{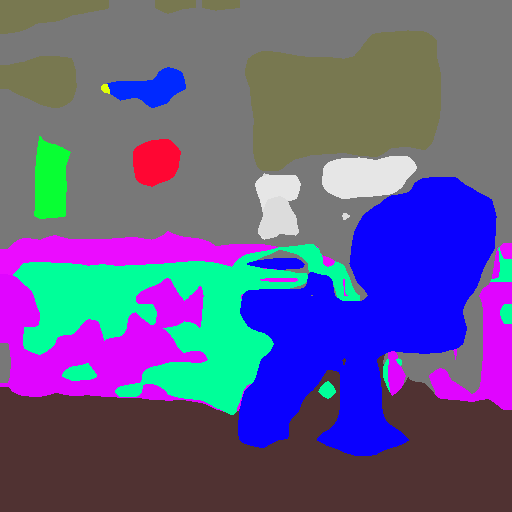}
\includegraphics[width=0.19\linewidth]{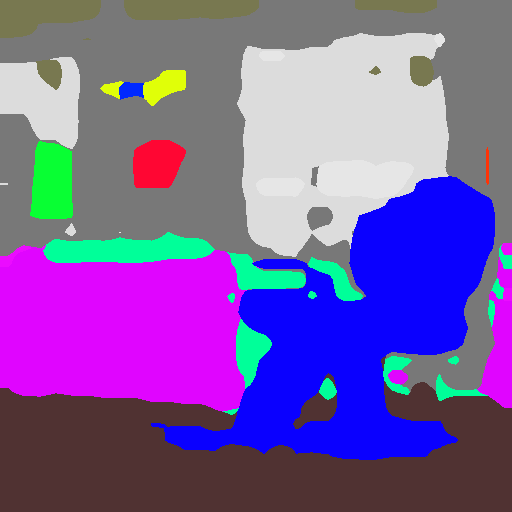}
\includegraphics[width=0.19\linewidth]{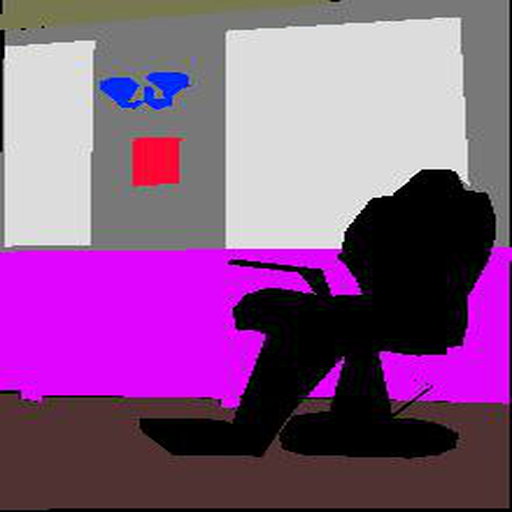}
\\ 

\includegraphics[width=0.19\linewidth]{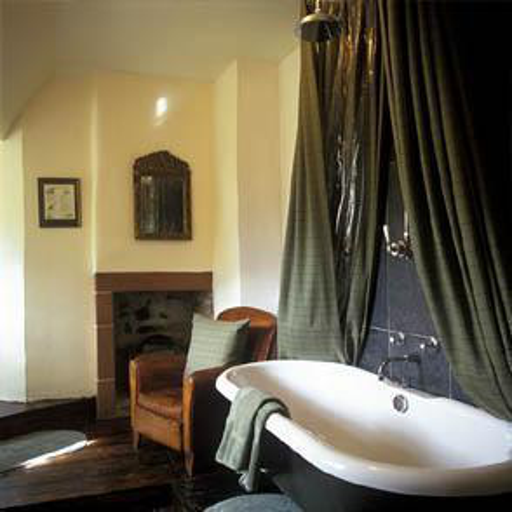}
\includegraphics[width=0.19\linewidth]{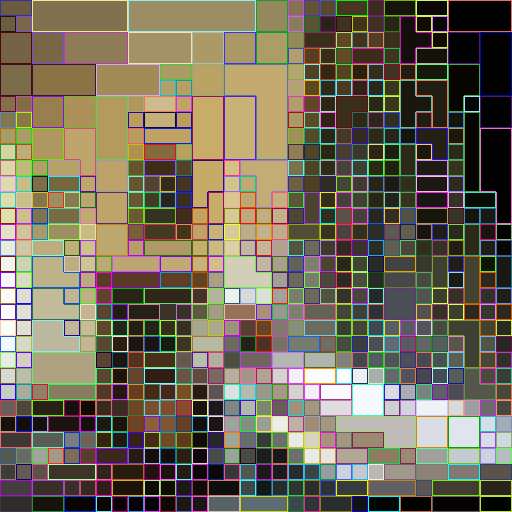}
\includegraphics[width=0.19\linewidth]{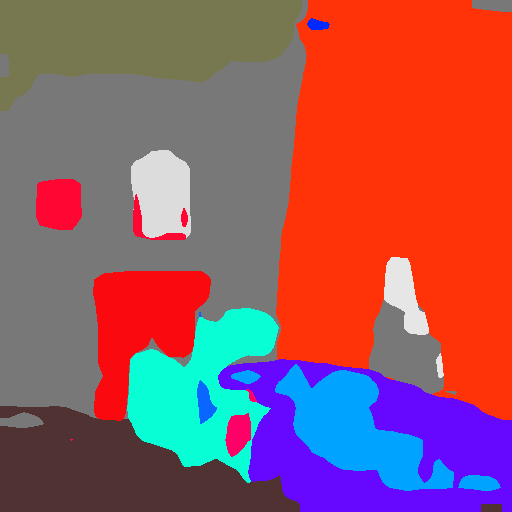}
\includegraphics[width=0.19\linewidth]{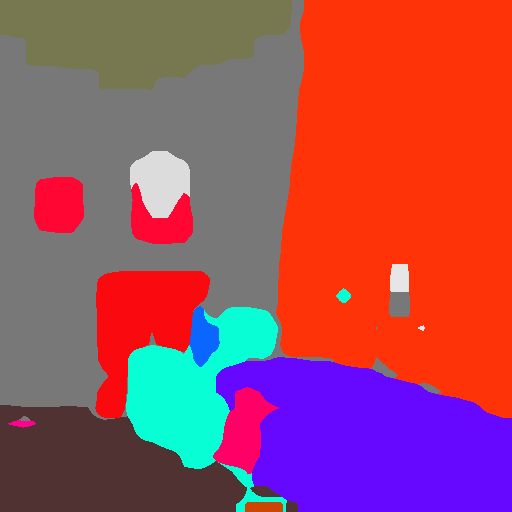}
\includegraphics[width=0.19\linewidth]{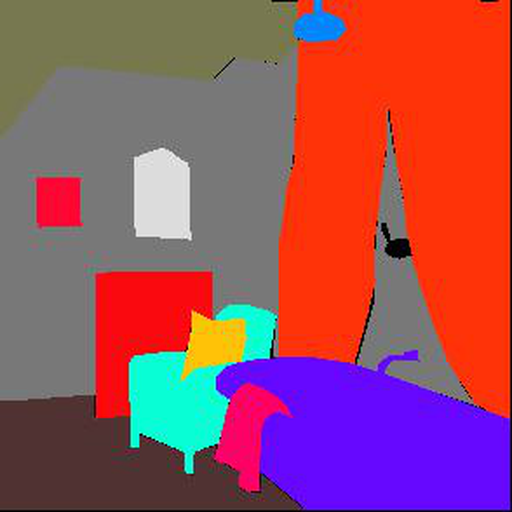}
\\ 

\includegraphics[width=0.19\linewidth]{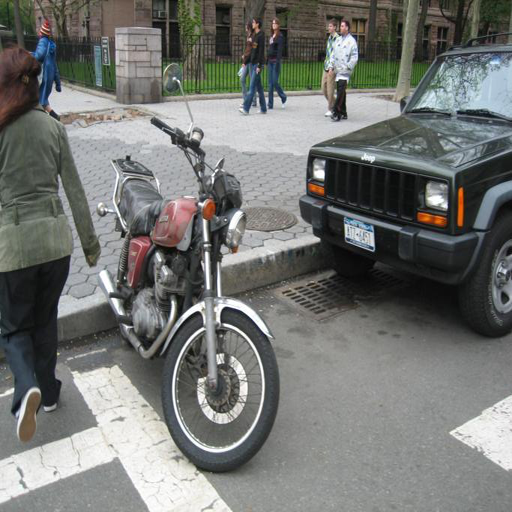}
\includegraphics[width=0.19\linewidth]{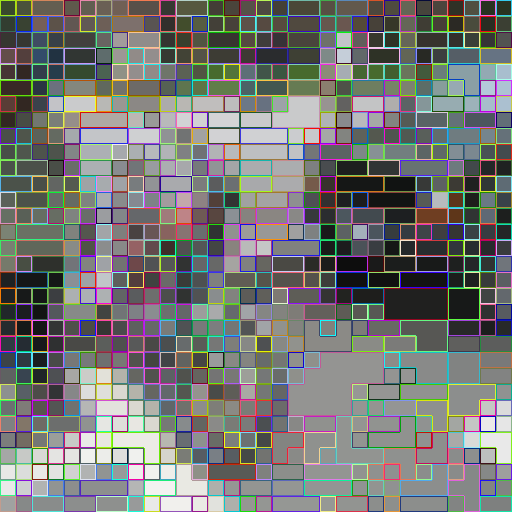}
\includegraphics[width=0.19\linewidth]{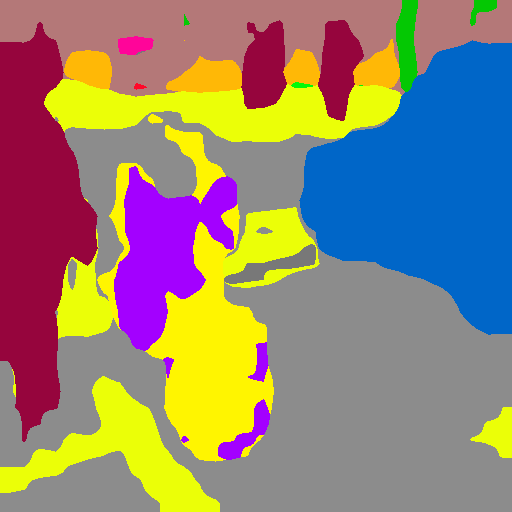}
\includegraphics[width=0.19\linewidth]{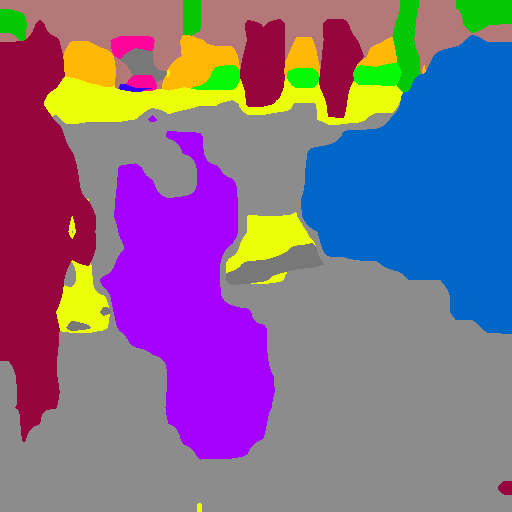}
\includegraphics[width=0.19\linewidth]{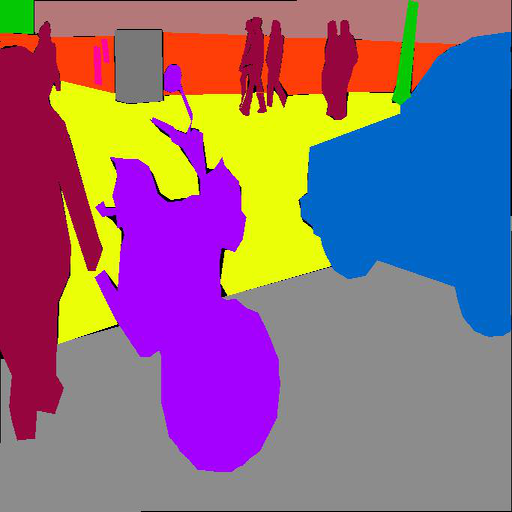}
\\

\includegraphics[width=0.19\linewidth]{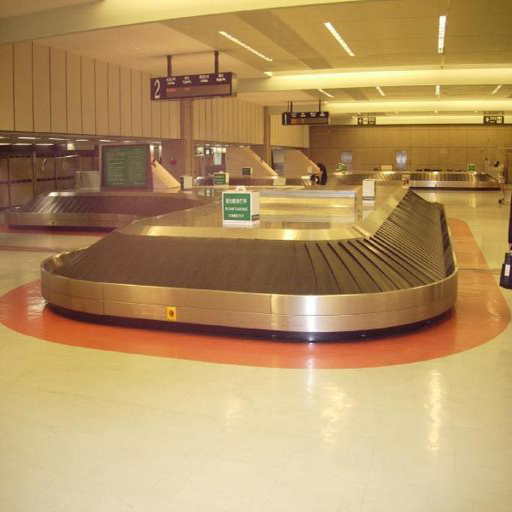}
\includegraphics[width=0.19\linewidth]{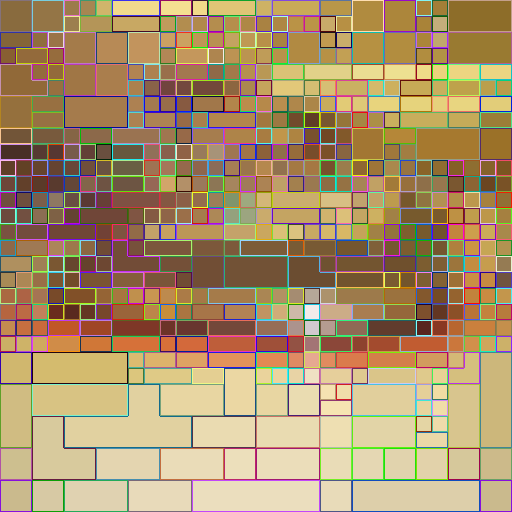}
\includegraphics[width=0.19\linewidth]{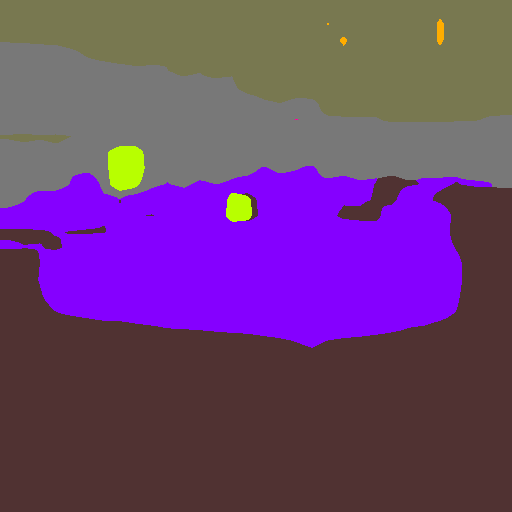}
\includegraphics[width=0.19\linewidth]{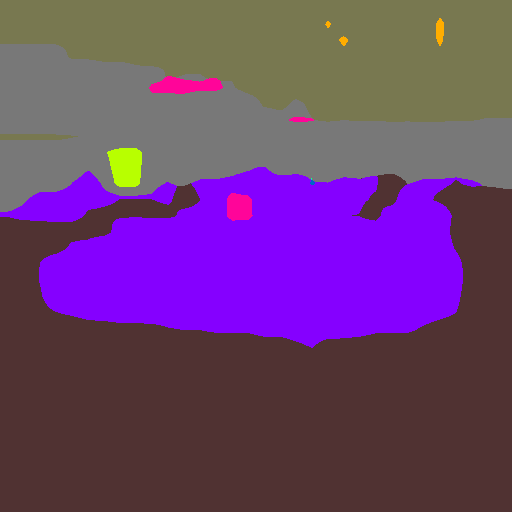}
\includegraphics[width=0.19\linewidth]{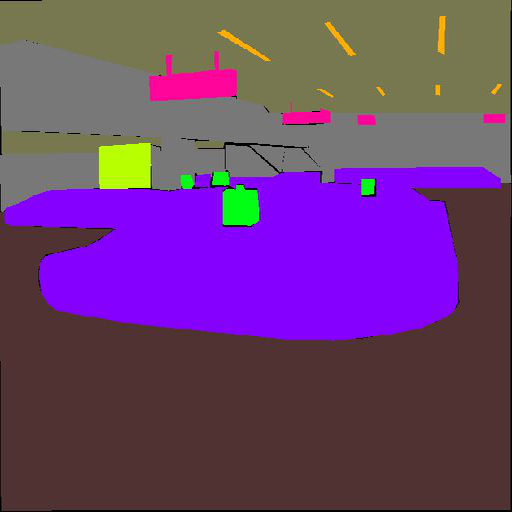}
\\

 \begin{subfigure}[b]{0.19\textwidth}
     \centering
     \caption{Input image}
 \end{subfigure}
  \begin{subfigure}[b]{0.19\textwidth}
     \centering
     \caption{Token merging}
 \end{subfigure}
 \begin{subfigure}[b]{0.19\textwidth}
     \centering
     \caption{Baseline}
 \end{subfigure}
 \begin{subfigure}[b]{0.19\textwidth}
     \centering
     \caption{With ALGM (ours)}
 \end{subfigure}
  \begin{subfigure}[b]{0.19\textwidth}
     \centering
     \caption{Ground truth}
 \end{subfigure}

\caption{\textbf{Examples of predictions by Segmenter with ViT-S and ALGM on ADE20K~\cite{zhou2017ade20k}.}} 
\label{fig:seg_results_ade20k}
\end{figure*}

\begin{figure*}[t]
\centering
\includegraphics[width=0.19\linewidth]{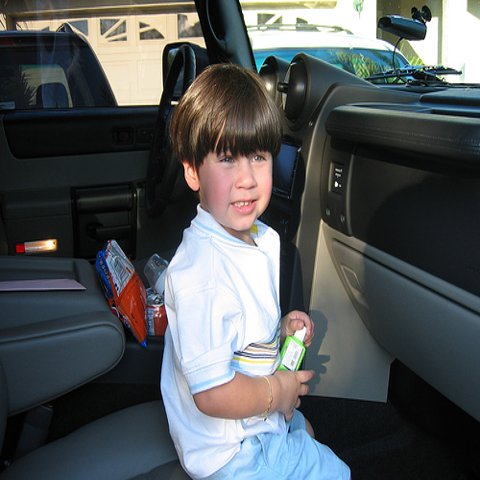}
\includegraphics[width=0.19\linewidth]{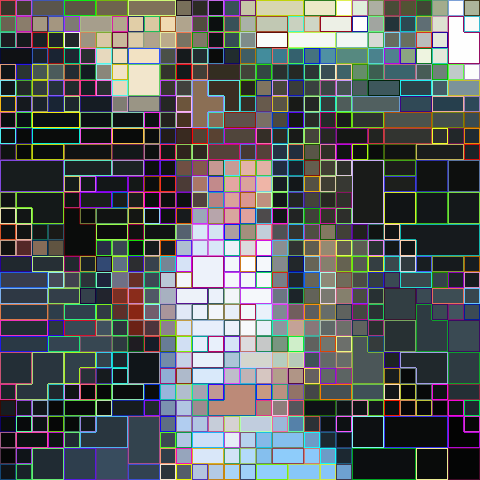}
\includegraphics[width=0.19\linewidth]{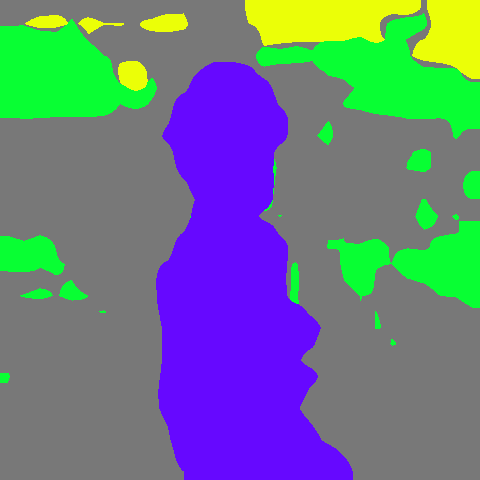}
\includegraphics[width=0.19\linewidth]{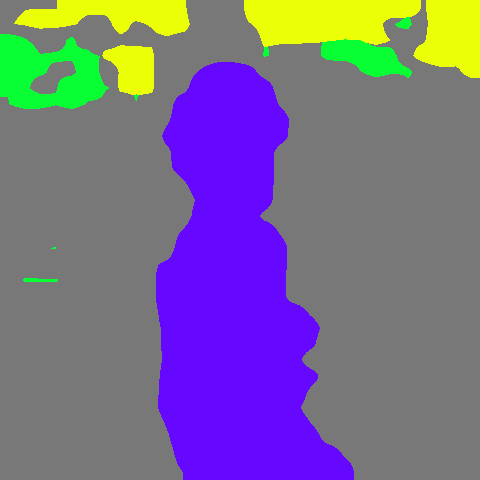}
\includegraphics[width=0.19\linewidth]{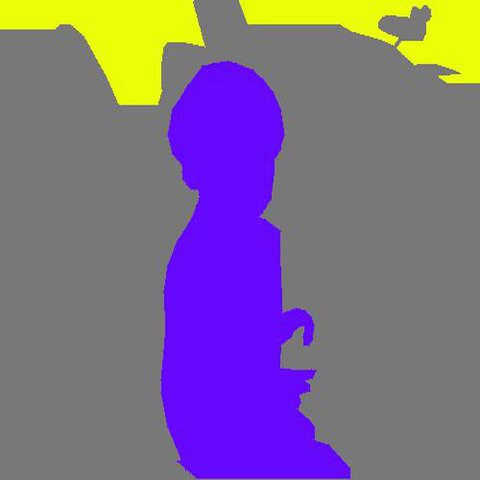}
\\
\includegraphics[width=0.19\linewidth]{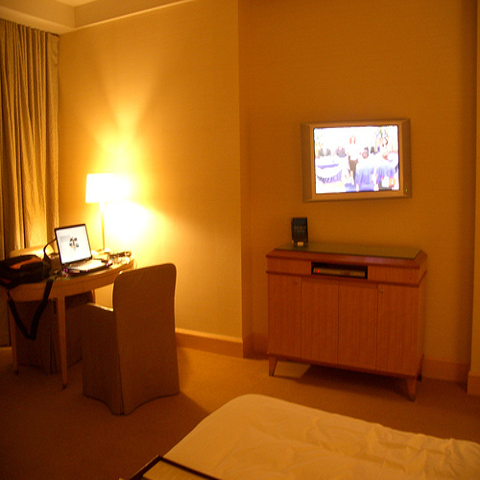}
\includegraphics[width=0.19\linewidth]{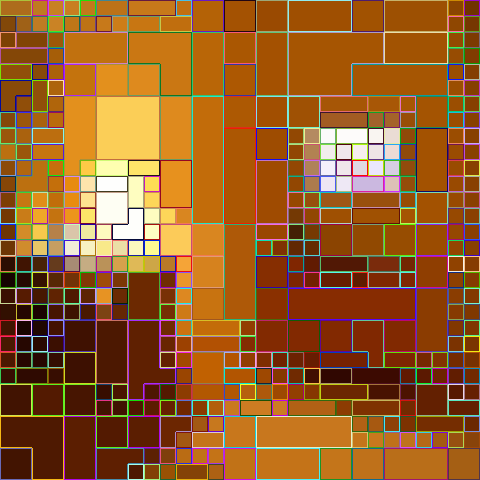}
\includegraphics[width=0.19\linewidth]{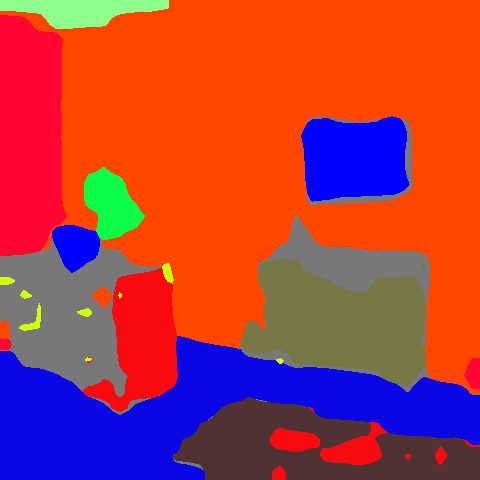}
\includegraphics[width=0.19\linewidth]{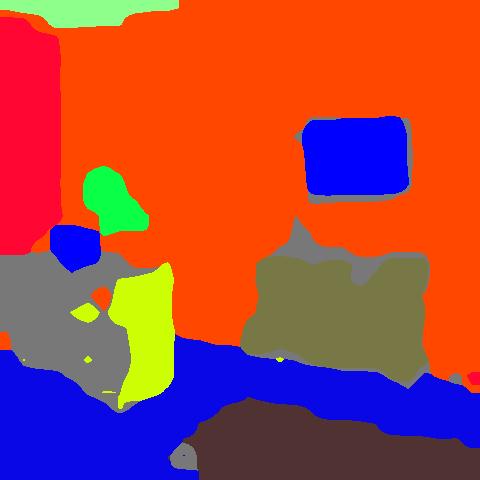}
\includegraphics[width=0.19\linewidth]{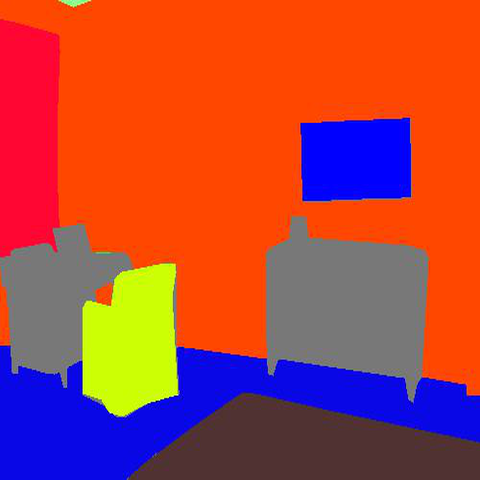}
\\
\includegraphics[width=0.19\linewidth]{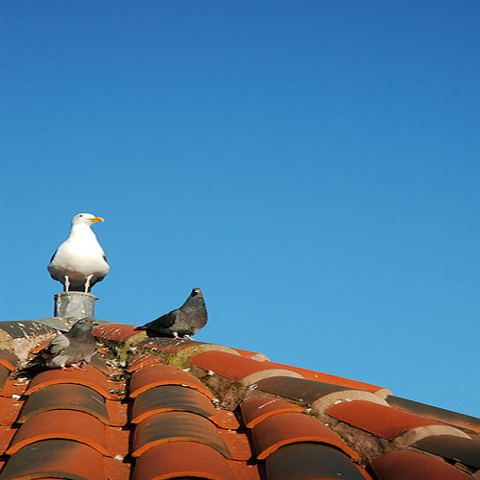}
\includegraphics[width=0.19\linewidth]{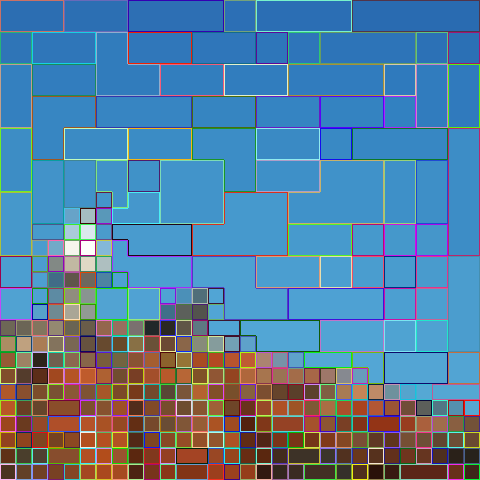}
\includegraphics[width=0.19\linewidth]{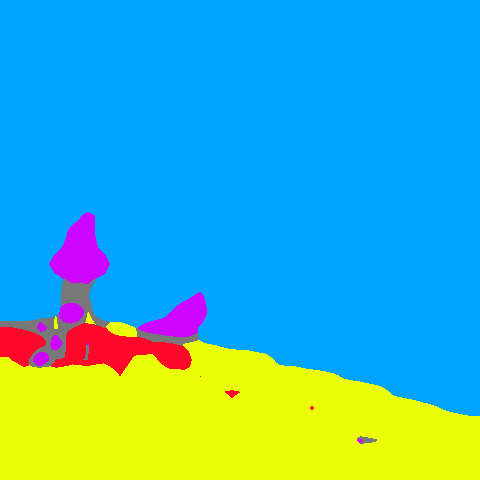}
\includegraphics[width=0.19\linewidth]{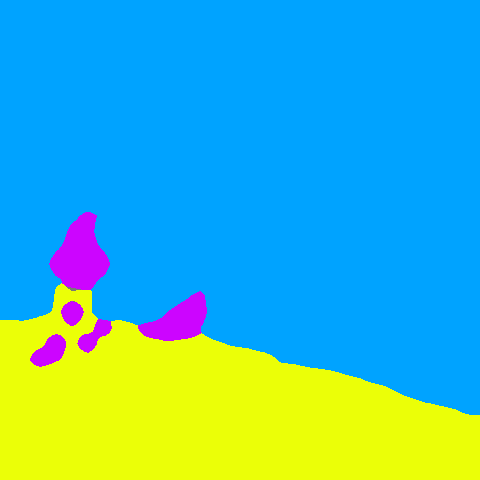}
\includegraphics[width=0.19\linewidth]{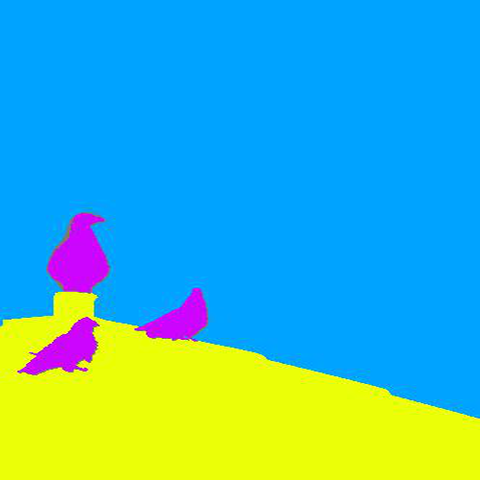}
\\
\includegraphics[width=0.19\linewidth]{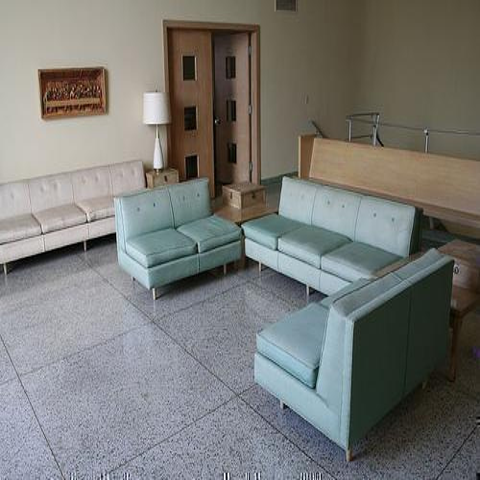}
\includegraphics[width=0.19\linewidth]{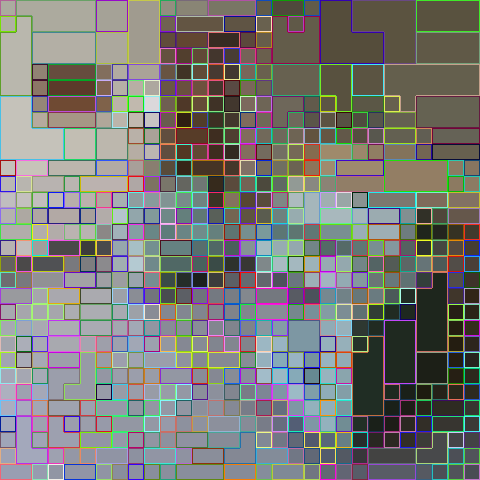}
\includegraphics[width=0.19\linewidth]{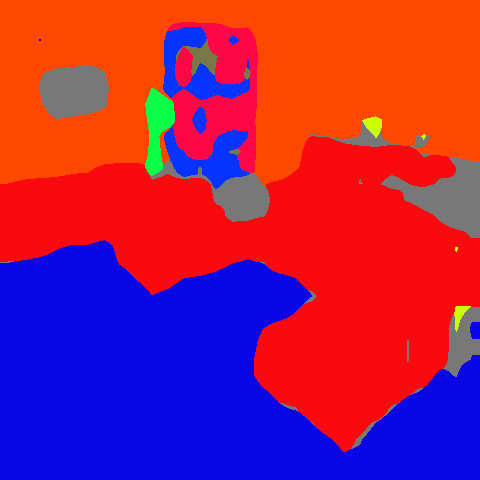}
\includegraphics[width=0.19\linewidth]{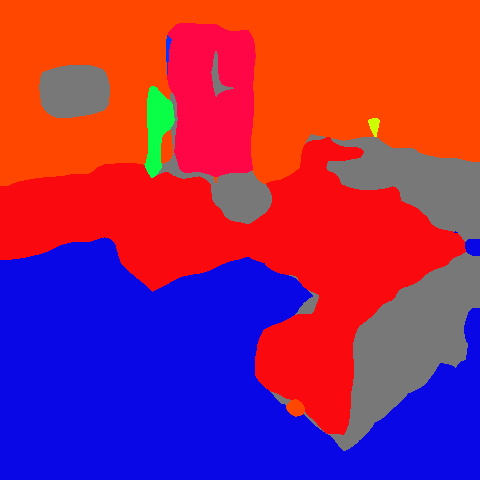}
\includegraphics[width=0.19\linewidth]{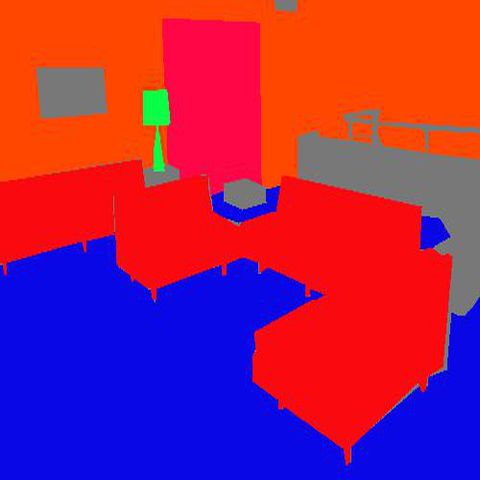}
\\

\includegraphics[width=0.19\linewidth]{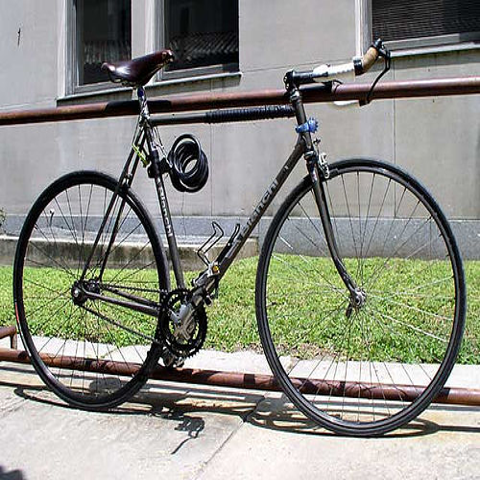}
\includegraphics[width=0.19\linewidth]{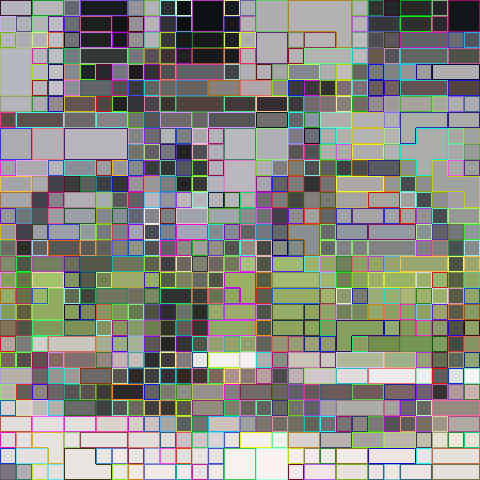}
\includegraphics[width=0.19\linewidth]{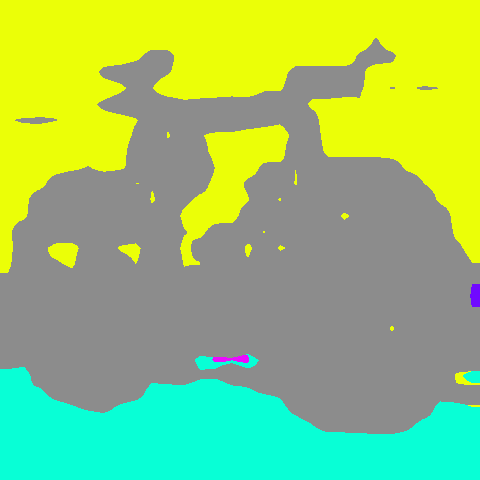}
\includegraphics[width=0.19\linewidth]{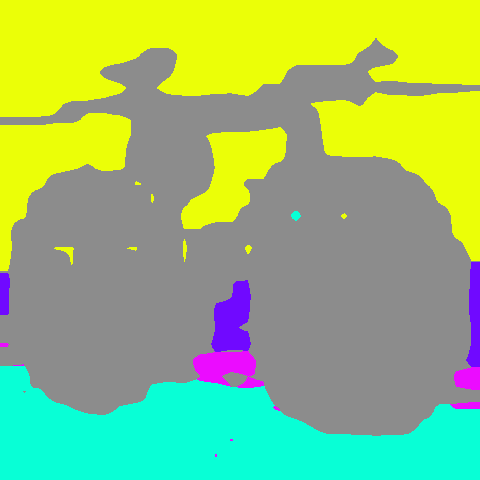}
\includegraphics[width=0.19\linewidth]{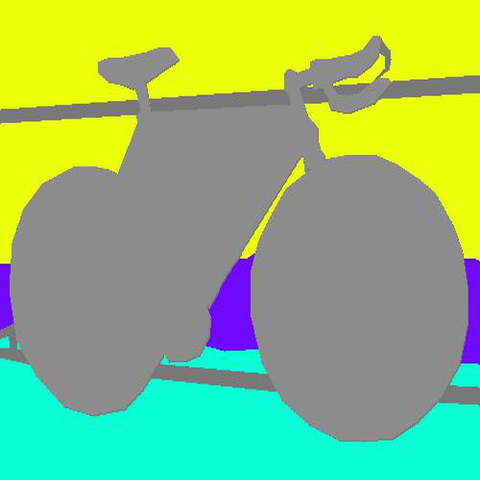}
\\

\includegraphics[width=0.19\linewidth]{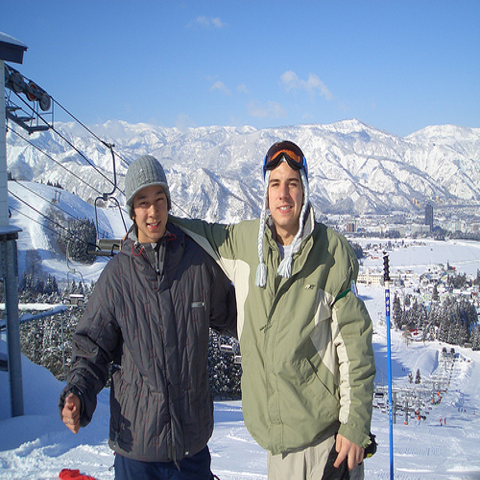}
\includegraphics[width=0.19\linewidth]{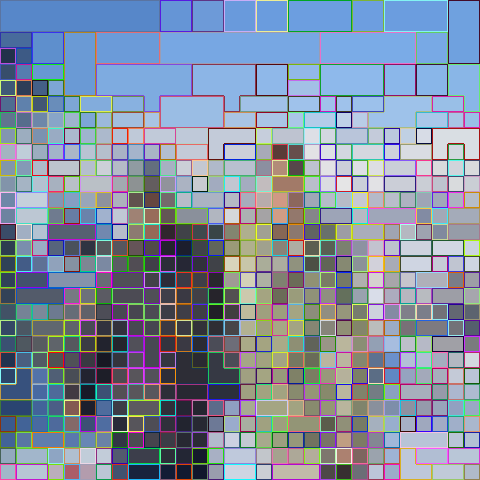}
\includegraphics[width=0.19\linewidth]{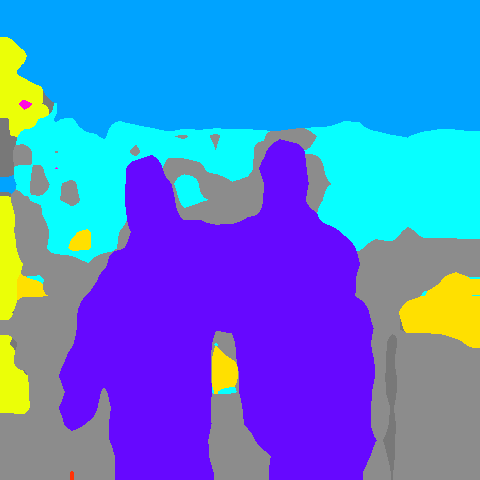}
\includegraphics[width=0.19\linewidth]{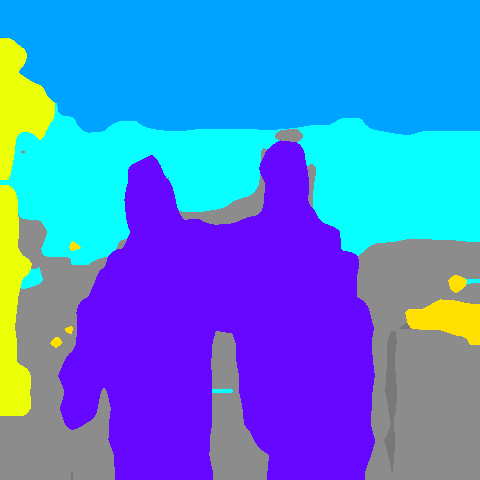}
\includegraphics[width=0.19\linewidth]{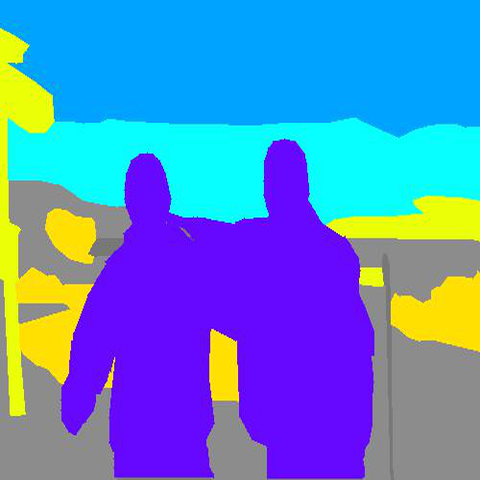}
\\

 \begin{subfigure}[b]{0.19\textwidth}
     \centering
     \caption{Input image}
 \end{subfigure}
  \begin{subfigure}[b]{0.19\textwidth}
     \centering
     \caption{Token merging}
 \end{subfigure}
 \begin{subfigure}[b]{0.19\textwidth}
     \centering
     \caption{Baseline}
 \end{subfigure}
 \begin{subfigure}[b]{0.19\textwidth}
     \centering
     \caption{With ALGM (ours)}
 \end{subfigure}
  \begin{subfigure}[b]{0.19\textwidth}
     \centering
     \caption{Ground truth}
 \end{subfigure}

\caption{\textbf{Examples of predictions by Segmenter with ViT-S and ALGM on Pascal-Context~\cite{mottaghi2014pascalcontext}.}} 
\label{fig:seg_results_pascalcontext}
\end{figure*}

In \cref{fig:seg_results_ade20k} and \cref{fig:seg_results_pascalcontext}, we show examples of semantic segmentation predictions by Segmenter~\cite{strudel2021segmenter} with ViT-S~\cite{dosovitskiy2021an}, both with and without our ALGM.

\end{document}